\documentclass[a4paper,fleqn]{cas-dc}

\usepackage[numbers]{natbib}
\usepackage{amsmath,amssymb,amsfonts}
\usepackage{algorithmic}
\usepackage{graphicx}
\usepackage{textcomp}
\usepackage{hyperref}
\usepackage{xcolor}
\usepackage{booktabs}
\usepackage{array}
\usepackage{subcaption}
\usepackage{pgfplots}
\usepackage{placeins}
\usepackage{float}
\usepackage{mathtools}
\usepackage[normalem]{ulem}
\newdefinition{definition}{Definition}
\newtheorem{proposition}{Proposition}
\newdefinition{remark}{Remark}
\newproof{proof}{Proof}

\newcommand{\R}{\mathbb{R}}

\newcommand{\E}{\mathbb{E}}

\newcommand{\Xspace}{\mathcal{X}}
\newcommand{\Yspace}{\mathcal{Y}}
\newcommand{\Grid}{\Omega}                 
\newcommand{\simplex}{\Delta}              
\newcommand{\attr}{\Phi}                   
\newcommand{\gaze}{\mathbf{G}}             
\newcommand{\sal}{\mathbf{S}}              
\newcommand{\norml}{\mathcal{N}}           
\newcommand{\proj}{\pi}                    
\newcommand{\resize}{\mathcal{R}}          
\newcommand{\KL}{\mathrm{D_{KL}}}

\newcommand{\JSS}{\mathrm{JSS}}
\newcommand{\PCC}{\mathrm{PCC}}
\newcommand{\CSS}{\mathrm{CSS}}
\newcommand{\relu}{\mathrm{ReLU}}
\newcommand{\pospart}[1]{\left(#1\right)^{+}}
\newcommand{\dset}{\mathcal{D}}
\newcommand{\Pspace}{\mathcal{P}}
\newcommand{\model}{f}
\newcommand{\params}{\theta}

\def\tsc#1{\csdef{#1}{\textsc{\lowercase{#1}}\xspace}}
\tsc{WGM}
\tsc{QE}

\begin{document}
\let\WriteBookmarks\relax
\def\floatpagepagefraction{1}
\def\textpagefraction{.001}

\shorttitle{Bridging Human and Model Attention}    

\shortauthors{M. Stefani et al.}  

\title [mode = title]{Bridging Human and Model Attention: Explainability Analysis of CNN, Mamba, and ViT Architectures with Gaze-Based Validation}  

\author[1]{Marco Stefani}[orcid=0009-0006-8807-7660]
\cormark[1]
\ead{marcostefani.research@gmail.com}

\author[1]{Edoardo Parpaiola}
\ead{edoardo.parpaiola@gmail.com}

\author[1]{Matteo Bino}
\ead{matteo.bino.2002@gmail.com}

\author[1]{Gianluca Caregnato}
\ead{gianluca.caregnato@gmail.com}

\author[1]{Francesco Roncolato}
\ead{roncolatofra@gmail.com}

\author[1]{Daniel Fusaro}
\ead{daniel.fusaro@studenti.unipd.it}

\author[1]{Loris Nanni}
\ead{loris.nanni@unipd.it}

\author[1]{Alberto Pretto}
\ead{alberto.pretto@unipd.it}

\affiliation[1]{organization={Department of Information Engineering (DEI), University of Padova},
            city={Padova},
            country={Italy}}

\cortext[1]{Corresponding author}

\begin{abstract}
The use of deep learning in high-risk application domains, under regulatory frameworks such as the EU AI Act, requires a precise characterization of the model's decision-making mechanisms. The existing literature provides limited quantitative and qualitative comparisons of explainability across widely adopted deep learning architectures.
We conduct a systematic evaluation of explainability in three representative model families: Convolutional Neural Networks (CNNs), Vision Transformers (ViTs), and Vision Mamba (ViM) state-space models.
Unlike existing studies, which are predominantly based on qualitative evaluations, we quantify the alignment between the model's attention and human visual perception using eye-tracking data treated as ground truth. We formalize this comparison by applying a standardized normalization pipeline that converts both post-hoc attribution maps and human eye-tracking heatmaps into valid spatial probability distributions. We then evaluate their alignment using a similarity metric whose range, optimal bounds, and invariance properties are rigorously characterized. We evaluate the models in three different domains: fine-grained classification (CUB-200-2011), medical imaging (MIMIC-CXR), and emotion recognition (KDEF). Using architecture-specific explainability (XAI) methods, including GradCAM, ScoreCAM, and AblationCAM for CNNs, ViTLRP for ViT, and MambaLRP for ViM, we measure the fidelity of model saliency maps to human gaze using Jensen-Shannon similarity, Chi-square similarity, and Pearson correlation, three estimators that we show to be sensitive to complementary aspects of a saliency map, namely support overlap, mass discrepancy, and structural co-variation.
Our results reveal distinct attention dynamics: CNNs tend to exhibit a more global focus, with spatially more diffuse saliency maps that correlate well with human perception in general contexts, while ViTs demonstrate fine-grained, patch-level attention, often focused on discriminative microfeatures. The Mamba architecture shows biased behavior: it achieves strong explainability in medical tasks but exhibits divergent intensity distributions in natural images; our analytical characterization of the metrics makes this dissociation explicit, since near-unit support overlap is attained simultaneously with a vanishing linear correlation.
We further analyze the effect of backbone freezing and ensemble strategies on model interpretability. The results highlight how architectural and training choices influence the alignment between model saliency and human visual attention.
\end{abstract}


\maketitle

\section{Introduction}
Recently, deep learning has emerged as the leading artificial intelligence paradigm for solving a wide range of tasks, spanning industrial applications, medical diagnosis, autonomous systems, and many others. In particular, deep learning models have achieved state-of-the-art performance in numerous computer vision tasks, including image classification, object detection, and semantic segmentation.

Regulations such as the European General Data Protection Regulation (GDPR) are making the use of black-box models more challenging across all business sectors, including healthcare, since the traceability of decisions is a fundamental requirement~\cite{DBLP:journals/corr/abs-1712-09923}. Furthermore, on August 1st, 2024, the Artificial Intelligence Act (AI Act) entered into force following its publication in the Official Journal of the European Union~\cite{aiact}, laying down harmonized rules for AI systems. The AI Act adopts a risk-based approach, whereby AI systems are subject to different requirements depending on their risk level, which is linked to their context of use and the potential impact on the health and safety or fundamental rights of individuals. In particular, high-risk AI systems must comply with a set of requirements that include, among others, transparency and human oversight~\cite{10.1145/3593013.3594069}.

A machine learning model is \emph{transparent} if a human can understand its entire logic at once. This typically requires that the model has a low computational complexity. A more relaxed definition of transparency requires that each individual part, including inputs, parameters, and calculations, has an intuitive, human-understandable meaning~\cite{lepri}.
In this context, \emph{interpretability} denotes the degree to which a human can understand a system’s internal mechanics~\cite{10.1145/3728637}. This may be achieved intrinsically through a model’s transparent design or through post-hoc methods (a post-hoc interpretation is generated after a model has already made a prediction). The latter refers to the ability to consistently predict a model's output based on its inputs,  an essential quality for testing complex ``black-box'' models~\cite{interpretability,10.1145/3233231}.
Building on these foundations, \emph{Explainable AI (XAI)}~\cite{BARREDOARRIETA202082} aims to formalize these interpretations using mathematically rigorous frameworks. In practice, \emph{explainability} refers to the suite of methods (often post-hoc, but not exclusively) that generate human-understandable justifications of a model’s behavior, leveraging internal parameters, outputs, or a combination of both.

While inherently interpretable architectures, such as linear regressors or shallow decision trees, offer direct decomposability~\cite{10.1145/3233231} and are easily visualized through their model parameters~\cite{DBLP:journals/corr/abs-2005-13799}, they are structurally limited by their representational capacity. In high-dimensional domains like computer vision, these models fail to capture the hierarchical, non-linear features necessary for competitive performance in tasks such as image classification and segmentation. This creates a fundamental accuracy-interpretability trade-off: as models scale toward the complexity of Deep Neural Networks (DNNs)~\cite{10.5555/2999134.2999257} or Transformers~\cite{vit} to minimize empirical risk, they systematically sacrifice transparency. Consequently, the adoption of black-box architectures becomes a functional requirement for complex tasks, necessitating the development of robust post-hoc XAI frameworks to recover human-understandable insights. Existing XAI approaches can be broadly categorized by the timing of the explanation (ante-hoc vs. post-hoc), the scope (global vs. local), and model agnosticism (agnostic vs. specific). Furthermore, they are distinguished by their output format (e.g., feature attribution, visualization, or natural language) and their underlying computational mechanism (e.g., surrogate-, perturbation-, or backpropagation-based). A concise taxonomy is provided in Section~\ref{sec:taxonomy}.

The goal of our work is to quantitatively evaluate the alignment between human visual attention and the saliency maps produced by architecture-specific XAI methods. Stated formally, given a trained classifier $\model_\params$, an attribution operator $\attr$ returning an attribution map $\sal=\attr(\model_\params,x,c)$ for a target class $c$, and an empirical human fixation-density map $\gaze$, representing the human eye-tracking heatmap recorded for the same visual stimulus $x$, we analyze the functional:
\begin{equation}
    \mathcal{A}_{\mathcal S}(\model_\params,\attr)\;=\;\E_{(x,\gaze)}\bigl[\,\mathcal{S}\bigl(\attr(\model_\params,x,\hat y(x)),\,\gaze\bigr)\bigr],
    \label{eq:intro_alignment}
\end{equation}
where $\mathcal{S}$ is a similarity functional between two spatial distributions and the expectation is taken over the data distribution. The central point is that $\gaze$ enters only through $\mathcal{S}$ and never through the training objective, so that $\mathcal{A}_{\mathcal S}$ probes an \emph{emergent} property of the trained model. The formal apparatus underlying Eq.~\eqref{eq:intro_alignment} is developed in Section~\ref{sec:problem}. 
Our study targets widely adopted CNN, ViT, and Vision Mamba architectures across three distinct domains: fine-grained natural image classification, medical imaging, and facial emotion recognition. By doing so, we provide insights into the internal behavior of each architectural family and highlight how specific design choices, such as backbone freezing and ensemble strategies, influence the fidelity of model explanations under diverse data regimes.

Closely related to our study are the works by Alqutayfi \emph{et al.} ~\cite{related_CNN_ViT} and Ruiu \emph{et al.} ~\cite{metrics}. While ~\cite{related_CNN_ViT} provides a comparative analysis of CNNs and ViTs in medical imaging, it relies primarily on qualitative expert assessment for explainability, without leveraging ground-truth gaze data to quantify the alignment. Conversely, ~\cite{metrics} establishes a rigorous framework for comparing model attention with human gaze using similarity metrics (such as Jensen-Shannon and Chi-square), but limits its scope to facial analysis, comparing ViTs against handcrafted features.
A similar work to ours is Kang \emph{et al.}~\cite{KANG2026102356}, which aims to decode human safety perceptions of urban environments by employing CNNs and various Class Activation Mapping~(CAM) techniques to generate visual attention heatmaps. To evaluate the alignment between XAI-generated heatmaps and human visual attention, the authors measured image similarity at two distinct levels. At the scene level, they employed L2 loss and Learned Perceptual Image Patch Similarity (LPIPS) scores to assess overall spatial and perceptual similarities, while at the element level, they utilized cosine similarity to verify whether both the human and the deep learning models focused on the same specific urban objects.

The need for such rigorous evaluation is underscored by the fact that explainability is no longer just a diagnostic tool but a driver of model integrity. Recent work by~\cite{muttenthaler2025aligning} demonstrates that aligning the internal representations of vision models with human conceptual hierarchies significantly enhances their generalization and out-of-distribution robustness. This principle is further supported by~\cite{11214385}, who argue that studying explainability is crucial not only for understanding which features the model relies on most, but also for enhancing overall performance. Indeed, the authors propose an intrinsic XAI framework for object detection with YOLOv8, where predictions are contingent upon the explicit verification of human-meaningful parts defined by the user. This transparency-by-design approach significantly improves model accuracy and provides enhanced robustness against adversarial attacks compared to traditional black-box models.

Our work bridges the gaps in the current literature by introducing a distinct evaluative perspective. While existing XAI research often relies on qualitative visual inspections of saliency maps, it lacks the objective metrics needed for rigorous comparison. Conversely, studies that do employ quantitative gaze data frequently use it as a ground-truth label during training to supervise the model’s attention.
For example, Govil \emph{et al.}~\cite{govil_2025} introduce DriverGaze360-Net, a transformer-based model explicitly trained on empirical human gaze data to predict omnidirectional driver attention. By utilizing these fixation distributions as ground truth, their network is designed to generate accurate gaze maps for unseen scenarios, essentially treating gaze prediction as the primary learning objective.
In contrast, we focus on the inherent explainability of standard image classifiers. Rather than using gaze data to guide the learning process, we train our models solely for classification and subsequently extract their post-hoc saliency maps using XAI methods. Consequently, the human fixation-density map $\gaze$ constitutes a genuine held-out ground truth reference rather than a supervision signal. By using human gaze as an independent benchmark, rather than a training target, we can quantitatively measure how naturally these different architectures align with human visual priority. This methodology allows us to provide a rigorous validation of CNN, ViT, and the novel Vision Mamba architectures across both fine-grained and medical imaging tasks.

Our contributions are:
\begin{enumerate}
    \item A review of recent explainability papers across a wide range of model architectures, including Convolutional Neural Networks (CNNs), Vision Mamba (ViM), and Vision Transformers (ViTs), presented within a unified notation that expresses Grad-CAM, Score-CAM, Ablation-CAM, ViTLRP, and MambaLRP as instances of a single attribution operator.
    \item A formal evaluation framework for gaze-based explainability assessment. We define a normalization pipeline that maps both attribution maps and human fixation densities onto the probability simplex (i.e., the space of non-negative spatial distributions summing to unity) and derive the range, extremal configurations, and invariance properties of the three similarity estimators employed. This analysis yields an operational reading of the metrics, demonstrating that support overlap, mass discrepancy, and structural co-variation are separately identifiable. In turn, this provides a theoretical explanation for the negative similarity values observed empirically for state-space models.
    \item An evaluation of XAI techniques across three radically distinct visual domains: fine-grained natural classification (CUB), specialized medical diagnostics (CXR), and human facial emotion recognition (KDEF-FGD). The analysis demonstrates how domain-specific semantic features deeply influence the alignment between machine saliency and human visual attention.
    \item A systematic comparison and quantitative evaluation of the explainability of different models, analyzing how these models attend to visual information using human gaze datasets as a reference, together with a pairwise model-to-model agreement analysis that decouples inter-architecture divergence from human-alignment.
    \item A qualitative analysis of model perception through the reproduction and comparison of the various heatmaps generated by different explainability methods.
    \item An evaluation of the performance benefits of model ensembles composed of networks with similar architectures, investigating whether consistent explainability patterns correlate with improved accuracy.
\end{enumerate}

The code is publicly available to ensure reproducibility at
\url{https://github.com/CaregnatoGianluca/VisionExplainability}  and \url{https://github.com/LorisNanni/VisionExplainability}. All the heatmaps produced by the various XAI techniques are provided at
\url{https://doi.org/10.5281/zenodo.21127362}.

\section{A Concise Taxonomy of XAI Methodologies}
\label{sec:taxonomy}
In this section, we provide a unified taxonomy and classification of explainable artificial intelligence (XAI) techniques, providing a formal mathematical framework that encompasses both ante-hoc and post-hoc paradigms across local and global scopes.

We first define the core notation, decision spaces, and predictive models underlying these methodologies. Let $\Grid=\{1,\dots,H\}\times\{1,\dots,W\}$ be the pixel lattice of the input images, $\Xspace\subseteq\R^{H\times W\times 3}$ the input space, $\Yspace=\{1,\dots,K\}$ a finite label set, and $\mathcal{F}=\{\model_\params:\Xspace\to\simplex^{K-1}\mid \params\in\R^{d}\}$ a parametric hypothesis class mapping images into the probability simplex $\simplex^{K-1}=\{p\in\R^{K}_{\ge0}:\sum_{c=1}^{K}p_c=1\}$. We denote by $\model_\params(x)_c$, $x \in \Xspace$, the predicted posterior of class $c \in \Yspace$, by $z_c(x)$ the corresponding pre-softmax logit, and by $\hat y(x)=\arg\max_{c\in\Yspace}\model_\params(x)_c$ the predicted label. All the local, feature-attribution techniques discussed below are then instances of a single object.

\begin{definition}[Attribution operator]
\label{def:attribution}
An \emph{attribution operator} is a map
\begin{equation}
    \attr:\ \mathcal{F}\times\Xspace\times\Yspace\longrightarrow\R^{\Grid},\qquad \sal=\attr(\model_\params,x,c),
    \label{eq:attr_operator}
\end{equation}
assigning to every location $u\in\Grid$ a signed score $\sal(u)$ that quantifies the contribution of $u$ to the evidence supporting class $c$ for the input $x$ under the model $\model_\params$. Unless stated otherwise we always set $c=\hat y(x)$, i.e.\ explanations are computed for the class actually predicted by the model.
\end{definition}

Definition~\ref{def:attribution} is deliberately agnostic: the taxonomy below organizes the admissible choices of $\attr$ according to \emph{when} it is applied, the \emph{scope} it describes, the \emph{model access} it requires, \emph{what} it returns, and \emph{by which mechanism} it is computed.

The most fundamental split in explainable artificial intelligence involves the timing of explanation, with a distinction between ante-hoc and post-hoc methodologies. Ante-hoc approaches~\cite{Rudin2019-RUDSEB}, often referred to as intrinsic interpretability, involve the use of models that are transparent by design, where the internal logic and decision-making pathways are inherently understandable to a human observer. These methods contrast with post-hoc techniques, which are applied to a model after the training phase has been completed to extract explanations from an otherwise opaque ``black-box'' architecture~\cite{10.1145/3233231}. While the former relies on the structural simplicity of the model to ensure clarity, the latter employs external mechanisms, such as perturbation or gradient analysis, to provide meaningful insights into the learned representations of complex systems without altering the underlying model itself.

A second distinction is between Global and Local Interpretable Methods ~\cite{DBLP:journals/corr/abs-2005-13799}. \textit{Global Methods} aim to describe the overall behavior of a model, capturing its general decision logic across the dataset. They provide insights into how input features influence predictions on average, for example, through feature importance measures or partial dependence analysis~\cite{DBLP:journals/corr/abs-2005-13799}. In contrast, \textit{local methods} focus on explaining individual predictions. These approaches aim to clarify why a model produced a specific output for a given input instance, often by identifying the most influential features or by generating counterfactual examples. The distinction is captured by the domain of the explanation map: a local method returns $\attr(\model_\params,x,c)$ for a fixed $x$, whereas a global method returns a single summary $\attr^{\mathrm{glob}}(\model_\params)=\E_{x\sim p(x)}\bigl[\,\attr(\model_\params,x,\hat y(x))\bigr]$, or an analogous aggregate over the data distribution, in which the dependence on the individual instance has been marginalized out. Local explainability methods are widely used and constitute the main focus of this paper.

Model agnosticism refers to whether an explanation method can be applied independently of the underlying model. \textit{Model-agnostic methods} treat the model as a black-box and can be applied to any model, regardless of its internal structure, e.g., Local
Interpretable Model-agnostic Explanations (LIME)~\cite{lime}. Conversely, \textit{model-specific methods} are designed for particular model classes and typically exploit internal information such as gradients, activations, or weights, e.g., Layer-wise Relevance Propagation (LRP) for neural networks~\cite{lrp,xai_book_overview}.

The output format or modality of an explanation represents a critical distinction in XAI taxonomies, determining how the interpretation is communicated to the end-user. These formats are generally categorized based on the type of information they provide:
\begin{itemize}
    \item \textit{Feature attribution}: These methods identify and rank the input features that primarily influenced a specific decision. In computer vision, this is typically achieved through spatial visualizations, such as heatmaps or saliency maps, which highlight the regions or pixels contributing most to the prediction (e.g., via Grad-CAM~\cite{gradcam} or LRP~\cite{xai_book_overview}).
    \item \textit{Textual explanations}: With the rise of Large Language Models (LLMs), some approaches generate natural language justifications. These methods produce human-readable descriptions of the reasoning behind a prediction, often utilizing chain-of-thought (CoT) style prompting.
    \item \textit{Example-based explanations}: These approaches explain predictions by selecting representative samples from the training set. They typically identify prototypes (examples structurally similar to the input) or criticisms (examples that contrast with it) to explain a decision through analogy.
\end{itemize}

Furthermore, feature attribution methods can be classified into two main categories based on their underlying computational mechanism: \textit{perturbation-based} and \textit{propagation-based} approaches.

\textit{Perturbation/Occlusion-based methods} represent the most intuitive approach for probing a black-box model. These techniques function by systematically masking, removing, or modifying specific input features, executing a forward pass to compute the new output, and measuring the resulting deviation from the original prediction. Let $m\in\{0,1\}^{\Grid}$ be a binary mask and let $x\odot m+\bar{x}\odot(1-m)$ denote the input in which the locations selected by $m$ are replaced by a reference (baseline) value $\bar{x}$, with $\odot$ the Hadamard product. The occlusion attribution of a region $\mathcal{U}\subseteq\Grid$ is
\begin{equation}
    \attr^{\mathrm{occ}}(\model_\params,x,c)\big|_{\mathcal U}=z_c(x)-z_c\bigl(x\odot(1-m_{\mathcal U})+\bar{x}\odot m_{\mathcal U}\bigr),
    \label{eq:occlusion}
\end{equation}
where $m_{\mathcal U}$ is the indicator of $\mathcal U$. However, they are highly computationally expensive; running a forward pass after perturbing every individual feature or pixel one-by-one requires $\mathcal{O}(|\Grid|)$ evaluations of $\model_\params$ and can take several hours per image~\cite{DBLP:journals/corr/abs-1711-06104}. To mitigate this overhead, it is common to apply perturbation masks over localized patches (e.g., configurations of $5\times5$ or $10\times10$ pixels) depending on the spatial resolution required and the available computational budget, which reduces the cost by the patch area at the price of a coarser spatial support. A prominent example within this domain is Shapley value sampling, which approximates true Shapley values by repeatedly sampling permutations of the input space~\cite{mitchell2022samplingpermutationsshapleyvalue}. Interpreting the logit as the value function $v(\mathcal U)=z_c(x_{\mathcal U})$ of a cooperative game played by the $|\Grid|$ input features, the exact attribution of feature $u$ is the Shapley value
\begin{equation}
\begin{split}
    \phi_u(v)=\ &\sum_{\mathcal{U}\subseteq\Grid\setminus\{u\}}\frac{|\mathcal{U}|!\,\bigl(|\Grid|-|\mathcal{U}|-1\bigr)!}{|\Grid|!}\\[-2pt]
    &\quad\times\Bigl[v\bigl(\mathcal{U}\cup\{u\}\bigr)-v(\mathcal{U})\Bigr],
\end{split}
    \label{eq:shapley}
\end{equation}
whose exact evaluation is exponential in $|\Grid|$, which is precisely what motivates Monte-Carlo estimation over sampled permutations.

Conversely, \textit{propagation-based methods} evaluate feature importance by utilizing the internal structure of the network, typically extracting attribution scores via a backward pass, at a cost of a single backward sweep, i.e.\ $\mathcal{O}(1)$ evaluations of $\model_\params$ irrespective of $|\Grid|$. The most straightforward implementation computes the partial derivatives of the output neuron with respect to the input features,
\begin{equation}
    \attr^{\mathrm{grad}}(\model_\params,x,c)(u)=\frac{\partial z_c(x)}{\partial x_u},\qquad u\in\Grid,
    \label{eq:vanilla_gradient}
\end{equation}
which is the first-order term of the Taylor expansion of $z_c$ around $x$. To improve upon the noisiness of raw gradients, Guided Backpropagation (GBP)~\cite{springenberg2015strivingsimplicityconvolutionalnet} restricts the backward flow of gradients to only positive contributions by applying the ReLU function during the backward pass. Denoting by $a^{(l)}$ the activations and by $R^{(l)}$ the back-propagated signal at layer $l$, GBP replaces the standard ReLU backward rule by
\begin{equation}
    R^{(l)}_i=\mathbf{1}\bigl[a^{(l)}_i>0\bigr]\cdot\mathbf{1}\bigl[R^{(l+1)}_i>0\bigr]\cdot R^{(l+1)}_i ,
    \label{eq:gbp}
\end{equation}
so that both negative activations and negative incoming signals are discarded.
Writing $R^{(l)}_j$ for the relevance assigned to neuron $j$ of layer $l$, these methods are designed to satisfy a \emph{conservation} (or completeness) axiom
\begin{equation}
    \sum_{j\in l}R^{(l)}_j=\sum_{k\in l+1}R^{(l+1)}_k=\dots=z_c(x)\quad\forall\, l,
    \label{eq:conservation}
\end{equation}
which states that the evidence produced at the output is redistributed, and neither created nor destroyed, along the backward pass. 
Other advanced propagation techniques mathematically decompose the output score layer by layer. Prominent examples include Deep Taylor Decomposition~\cite{DBLP:journals/corr/MontavonBBSM15}, Deep Learning Important FeaTures (DeepLIFT)~\cite{DBLP:journals/corr/ShrikumarGK17}, DeepSHAP~\cite{DBLP:journals/corr/LundbergL17}, and Layer-wise Relevance Propagation (LRP)~\cite{lrp}.

Most of the XAI frameworks evaluated in the remainder of this paper fall precisely within the taxonomic intersection of post-hoc, local methods that utilize a feature attribution output modality derived through propagation-based mechanisms.

\section{Related works}
\label{sec:related}

Our work introduces a metric-driven framework to quantitatively evaluate Explainable AI (XAI) methods against human visual attention. To contextualize this contribution, this section reviews the existing literature on explainability across CNN, ViT, and Vision Mamba architectures. As will be shown, while these prior studies explore model transparency, they do not incorporate empirical gaze-based validation, relying instead on qualitative visual inspections.

\subsection{Explainable AI for CNNs} \label{sec:XAI CNN}
Given the ``black-box'' nature of deep neural networks, several studies in the literature have addressed this issue, seeking to make the decisions made by different models more transparent and reliable.
Among these, some studies propose new explainability techniques or apply existing ones to validate different models.

One of the most popular Explainable AI (XAI) approaches is \textbf{Gradient-weighted Class Activation Mapping} (Grad-CAM)~\cite{gradcam}.
This approach uses the gradients flowing through the final convolutional layer to produce a gradient-weighted class activation map highlighting the regions of the image that most influence the target prediction.
Formally, let $A^{k} \in \mathbb{R}^{h \times w}$, $k = 1, \dots, D$, denote the $D$ feature maps of the final convolutional layer, and let $Z = h \times w$. Grad-CAM computes the neuron importance weights $\alpha^{c}_{k}$ via global average pooling of the pre-softmax logit gradients:
\begin{equation}
    \alpha^{c}_{k} = \frac{1}{Z} \sum_{i=1}^{h} \sum_{j=1}^{w} \frac{\partial z_{c}(x)}{\partial A^{k}_{ij}}.
    \label{eq:gradcam_alpha}
\end{equation}
with $z_c(x)$ the class logit. The attribution operator combines these feature maps linearly, discarding negative contributions via a ReLU activation:
\begin{equation}\attr^{\mathrm{GC}}(\model_\params,x,c)=\resize_{\Grid}\!\left(\relu\!\left(\sum_{k=1}^{D}\alpha^{c}_{k}A^{k}\right)\right)\!,
    \label{eq:gradcam}
\end{equation}
where $\resize_{\Grid}$ denotes bilinear upsampling from the feature-map lattice to the input lattice $\Grid$. Two structural limitations discussed in the remainder of this section follow directly from Eq.~\eqref{eq:gradcam}: the $\relu$ discards negative evidence, and the intrinsic resolution of the explanation is bounded above by $h\times w\ll|\Grid|$, so that the spatial granularity of the map is dictated by the architecture rather than by the attribution method. While initially popularized in medical imaging to make decisions more transparent, it is increasingly applied to diverse computer vision tasks. For instance, a recent study by Goyal \emph{et al.}~\cite{GOYAL2026104624} successfully leveraged Grad-CAM to evaluate \textbf{$G_{image}$}, a hyperparameter-optimized CNN for eye gaze classification. By analyzing Grad-CAM heatmaps, the authors were able to diagnose why baseline Transformers and CNNs failed: while $G_{image}$ accurately focused on essential features like the iris and eye corners, competing models missed these critical areas.
Our work builds upon this premise by formalizing the comparison: rather than relying on qualitative visual inspection, we introduce a rigorous quantitative framework that uses human gaze data as a ground truth to systematically measure the fidelity of these architecture-specific saliency maps.

For instance, the study in~\cite{related_CNN_1} demonstrates how, in the context of action recognition, Grad-CAM allows the validation of a multimodal approach based on InceptionV3, ResNetV2, and DenseNet-121. Using Grad-CAM, the authors report that it was possible to understand how the multimodal model focused on regions strictly related to actions and movement, even under variable environmental conditions. Grad-CAM has proven to be an explainability technique capable of distinguishing itself from other techniques for its ability to produce more detailed and robust heat maps.

In~\cite{related_CNN_ViT}, the limitations of Grad-CAM when used in non-CNN architectures are highlighted.
This study compares the results obtained by different ViT architectures (such as ViT-l-16, ViT-l-32, ViT-h-14) and CNNs (such as ResNet-50, EfficientNet-B5, VGG-16, Inception-V3) in the field of Disease Classification. The study highlights that ViTs, compared to CNNs, are not suitable for use with Grad-CAM, despite being able to obtain more precise results than CNNs thanks to their ability to capture long-range dependencies.
In particular, following an evaluation by experts in the domain in which the experiments were conducted, the visualizations generated by CNNs are more coherent and consistent with human perceptions.
Even for CNNs, the heatmaps generated by Grad-CAM fail to capture small regions of interest, underscoring this limitation.
The study thus demonstrates the non-universality of Grad-CAM, emphasizing the importance of adapting XAI techniques to non-CNN architectures and the need to adapt Grad-CAM to detect small relevant regions.

In addition to validating the reliability of the models, Grad-CAM was also used in~\cite{related_CNN_2} to highlight the structural weaknesses of some CNNs (such as Inception-V3, XceptionNet, and VGG-19). Experiments have shown that each structure has specific weaknesses, for example, Inception-V3 tends to overfit.
In the same study, they highlight the need for different approaches, agreeing with the discussion in~\cite{related_CNN_ViT}. The use of model-agnostic methods such as \textbf{Local Interpretable Model-agnostic Explanations} (LIME)~\cite{lime} and \textbf{SHapley Additive exPlanations} (SHAP)~\cite{shap-lundberg} can also provide explanations at the global level, capabilities that methods based solely on backpropagation cannot provide.
To overcome the limitations of Grad-CAM, which only analyzes the last convolutional layer and is therefore not robust to fine details, they propose the use of \textbf{Layer-wise Relevance Propagation} (LRP)~\cite{lrp}.
LRP is another XAI technique that, unlike coarser methods, provides a detailed pixel-wise visualization by redistributing the prediction score layer-by-layer via a backward pass. Using specific rules such as the $\epsilon$-rule to ensure numerical stability, LRP effectively propagates the prediction relevance down to the single raw pixels of the input. Denoting by $a_j^{(l)}$ the activation of neuron $j$ in layer $l$, by $w_{jk}$ the weight connecting it to neuron $k$ in layer $l+1$, and setting $z_{jk} = a_j^{(l)} w_{jk}$, the $\epsilon$-rule redistributes relevance $R_k^{(l+1)}$ according to:
\begin{equation}R^{(l)}_{j} = \sum_{k} \frac{z_{jk}}{\sum_{j'} z_{j'k} + \epsilon \operatorname{sign}\left( \sum_{j'} z_{j'k} \right)} \; R^{(l+1)}_{k},
    \label{eq:lrp_eps}
\end{equation}
where the stabilizer $\epsilon>0$ prevents the quotient from diverging when the denominator vanishes. The price is that the conservation law of Eq.~\eqref{eq:conservation} holds only approximately, since the relevance carried by weakly activated paths is absorbed by $\epsilon$; the choice of $\epsilon$ therefore governs an explicit trade-off between numerical stability and completeness of the attribution.
An analysis in~\cite{xai_book_overview} shows that there is no single ideal rule for propagating information with LRP. For this reason, it requires in-depth knowledge of the model in which it is applied and, consequently, the results obtained are subjective with respect to the models and data used.
Because of this limitation, LRP cannot be directly applied to any model but requires a manually designed composite strategy.

\begin{equation}
    \Delta^{c}_{k}=\model_\params\bigl(x\odot H^{k}\bigr)_{c}-\model_\params(x_{b})_{c},
    \label{eq:scorecam_cic}
\end{equation}
The channel importance weights $\alpha^{c}_{k}$ are then obtained by applying a softmax transformation across all $D$ channels, $\alpha^{c}_{k}=\exp(\Delta^{c}_{k})/\sum_{k'}\exp(\Delta^{c}_{k'})$, which are then combined exactly as in Eq.~\eqref{eq:gradcam}.
This methodology effectively avoids gradient saturation and noise issues, often producing more faithful visual explanations in deeper networks, since $\alpha^{c}_{k}$ depends on $\model_\params$ only through forward evaluations; the cost is $D$ additional forward passes per explanation.

Another approach in the literature that is gradient-free is \textbf{Ablation-based Class Activation Mapping} (Ablation-CAM)~\cite{ablationcam}. It determines the importance of each feature map by systematically ``switching it off'' to understand how the target class's prediction score drops.
In other words, through ablation analysis, this approach resets each feature map and measures the subsequent drop in the target class's score. Writing $z^{(\setminus k)}_{c}(x)$ for the class logit obtained after zeroing the $k$-th feature map, the weights of Eq.~\eqref{eq:gradcam_alpha} are replaced by the normalized ablation slope
\begin{equation}
    \alpha^{c}_{k}=\frac{z_{c}(x)-z^{(\setminus k)}_{c}(x)}{z_{c}(x)},
    \label{eq:ablationcam}
\end{equation}
which is a finite-difference surrogate of $\partial z_{c}/\partial A^{k}$ evaluated over the full ablation interval rather than in an infinitesimal neighborhood. By quantifying this direct impact, Ablation-CAM produces localization maps that are robust to gradient instability and, in most cases, more intuitive for identifying discriminative regions.

\subsection{Explainable AI for ViTs}
While the XAI techniques highlighted previously have proven effective for CNN architectures, their application to newer architectures is not straightforward.
As discussed in the previous subsection and highlighted in~\cite{related_CNN_ViT}, methods designed for CNNs struggle to adapt to models based on different prediction mechanisms, such as Vision Transformers (ViTs).

The authors of~\cite{vit_exp} highlight the lack of research in the literature on the explainability of ViTs. Indeed, one critical aspect is that, unlike CNNs, these types of models lack a clear method for visualizing their attention maps, making it difficult to find a method that makes their predictions as transparent as possible. 
However, it has been pointed out that the dimension used by the transformers' attention mechanism corresponds to that of the input sequence, and thus the attention mechanism can be easily visualized. It is precisely based on this insight that one of the most popular explainability approaches, called \textbf{Attention Rollout}~\cite{rollout}, was introduced to to accurately estimate the attention of ViTs. Alongside this, another widely used method, as in CNNs, is LIME~\cite{lime}, which generates local and interpretable models to explain each prediction. More recently, \textbf{Layerwise Relevance Propagation for Transformers} (TransLRP)~\cite{translrp} has been proposed, which allows aggregation of backward gradients and LRP across all layers and heads of attention modules to derive explanatory relevance.
The authors of~\cite{vit_exp} compare these three explainability approaches for ViTs in a medical imaging task by analyzing the sensitivity and faithfulness of ViTs.
Their results show that TransLRP outperforms the other two approaches in terms of faithfulness and sensitivity. This indicates that this approach produces features that are highly correlated with the model's predicted outcome. The results obtained by LIME are consistent across the different images analyzed, but they are not always accurate. Meanwhile, the attention rollout results appear to be excessively complex and sparse.

To address these issues, authors in ~\cite{transformer_interpretability} argue that methods like Attention Rollout and standard LRP are insufficient for Transformers. They point out that Rollout relies on simplistic assumptions of linearity that fail to distinguish between positive and negative contributions, often highlighting irrelevant tokens. Similarly, they note that standard LRP suffers from numerical instabilities due to the non-positive values inherent in GELU activations and skip connections. Consequently, they propose a novel solution based on the \textbf{Deep Taylor Decomposition} (DTD) principle, which facilitates class-specific relevance propagation that effectively handles these non-linearities and preserves total relevancy across layers. Concretely, let $B$ be the number of encoder blocks and let $A^{(b)}\in\R^{(N+1)\times(N+1)}$ be the attention matrix of block $b$ over the $N$ patch tokens plus the \texttt{[CLS]} token. Denoting by $\nabla A^{(b)}=\partial z_{c}(x)/\partial A^{(b)}$ the gradient of the target logit and by $R^{(b)}$ the DTD relevances of the same matrix, the method builds a per-block, gradient-weighted relevance map and composes the blocks multiplicatively,
\begin{equation}
\begin{split}
    \bar{A}^{(b)}&{}=I+\E_{h}\Bigl[\pospart{\nabla A^{(b)}\odot R^{(b)}}\Bigr],\\
    \mathbf{C}&{}=\bar{A}^{(1)}\bar{A}^{(2)}\cdots\bar{A}^{(B)},
\end{split}
    \label{eq:vitlrp}
\end{equation}
where $(\cdot)^{+}=\max(\cdot,0)$ removes the negative contributions that make plain rollout ambiguous and $\E_{h}$ averages over the attention heads. The explanation is the \texttt{[CLS]} row of $\mathbf{C}$ restricted to the patch tokens, $\mathbf{C}_{0,1:N}$, reshaped onto the $\sqrt{N}\times\sqrt{N}$ token grid and upsampled to $\Grid$. The identity term $I$ in Eq.~\eqref{eq:vitlrp} accounts for the residual connections, and its omission is exactly what makes naive attention visualization unfaithful. We refer to this operator as ViTLRP in the remainder of the paper.

\subsection{Explainable AI for Vision Mamba}

Moving beyond ViTs, a more recent architecture is represented by Mamba models, which are based on State Space Models (SSM)~\cite{gu2024mamba}, aiming to be an alternative to Transformers for long-context modeling. 
    \label{eq:ssm_continuous}
At its core, a continuous-time SSM maps a 1D input signal sequence $x_t \in \mathbb{R}$ to an output sequence $y_t \in \mathbb{R}$ through a latent state $h(t) \in \mathbb{R}^{n}$ governed by the linear system:
\begin{equation}
    h'(t) = \mathbf{A} h(t) + \mathbf{B} x_t, \qquad y(t) = \mathbf{C}^{\top} h(t),
    \label{eq:ssm_continuous}
\end{equation}
where $\mathbf{A} \in \mathbb{R}^{n \times n}$, $\mathbf{B} \in \mathbb{R}^{n \times 1}$, and $\mathbf{C} \in \mathbb{R}^{n \times 1}$. Discretizing Eq.~\eqref{eq:ssm_continuous} via a Zero-Order Hold (ZOH) with a step size parameter $\Delta > 0$ yields the discretized system matrices:
\begin{equation}
    \bar{\mathbf{A}} = \exp(\Delta \mathbf{A}), \qquad \bar{\mathbf{B}} = (\Delta \mathbf{A})^{-1} \bigl( \exp(\Delta \mathbf{A}) - I \bigr) \Delta \mathbf{B},
    \label{eq:ssm_zoh}
\end{equation}
which defines the linear recurrence evaluated during the forward pass:
\begin{equation}
    h_t = \bar{\mathbf{A}}_t h_{t-1} + \bar{\mathbf{B}}_t x_t, \qquad y_t = \mathbf{C}_t^{\top} h_t.
    \label{eq:ssm_discrete}
\end{equation}
The defining innovation of Mamba is \emph{selective context processing}: the parameters $(\Delta_t, \mathbf{B}_t, \mathbf{C}_t)$ are input-dependent, computed as learned linear projections of the current token input $x_t$, i.e., $\mathbf{B}_t = W_B x_t$, $\mathbf{C}_t = W_C x_t$, and $\Delta_t = \operatorname{softplus}(W_\Delta x_t)$.
Two key theoretical properties follow from this formulation. First, unrolling the recurrence in Eq.~\eqref{eq:ssm_discrete} expresses the output as $y_t = \sum_{s \le t} \mathbf{C}_t^\top \left( \bar{\mathbf{A}}_t \bar{\mathbf{A}}_{t-1} \cdots \bar{\mathbf{A}}_{s+1} \right) \bar{\mathbf{B}}_s x_s$, establishing an implicit causal attention operator driven by a data-dependent kernel. Second, because $(\Delta_t, \mathbf{B}_t, \mathbf{C}_t)$ explicitly depend on $x_t$, the recurrence becomes non-linear (bilinear) in $x_t$. This input dependency directly violates the linear conservation axiom in Eq.~\eqref{eq:conservation}, necessitating dedicated relevance propagation rules for state-space vision models.
Despite their growing adoption, to our knowledge, the literature contains a limited number of studies that thoroughly examine XAI for Mamba models.
Among these, we find~\cite{mamba_vit}, in which the authors introduce an in-depth comparison of the structural differences between the Transformers and the Mamba SSM. Specifically, the authors illustrate that the Mamba layer can be reformulated as an implicit form of a causal self-attention mechanism. While Mamba does not have an explicit mechanism to reveal where it is ``paying attentio'', this reformulation allows for the application of interpretability techniques similar to those traditionally used for Transformer-based models. They therefore propose an approach called Mamba-Attention, which provides greater transparency into the model's internal reasoning process.
In~\cite{mambalrp}, the authors analyze the use of LRP for the Mamba architecture. They highlight critical theoretical limitations arising from Mamba's distinct architectural components, specifically the Smooth Sigmoid Linear Unit (SiLU) activation function, the selective State Space Model, and the multiplicative gating applied to the SSM output stream.
The SiLU activation function is defined as $\mathrm{SiLU}(x) = x \cdot \sigma(x)$, where $\sigma(x) = (1 + e^{-x})^{-1}$, providing a smooth, non-monotonic nonlinearity.
The origin of these limitations can be made precise. A selective SSM block maps an input sequence $\{\mathbf{x}_{t}\}_{t=1}^{T}$ to an output sequence $\{\mathbf{y}_{t}\}_{t=1}^{T}$ via the recurrence in Eq.~\eqref{eq:ssm_discrete}, followed by a multiplicative gating operation $\mathbf{o}_{t} = \mathbf{y}_{t} \odot \operatorname{SiLU}(\mathbf{g}_{t})$, where $\mathbf{g}_t$ denotes the gate projection and $\odot$ represents the channel-wise Hadamard product. Relevance propagation through the recurrence itself is unproblematic, because it is linear once the discretized operators $(\bar{\mathbf{A}}_t, \bar{\mathbf{B}}_t, \mathbf{C}_t)$ are treated as fixed forward parameters. Conservation is instead broken at every \emph{bilinear product node} $\mathbf{o} = \mathbf{u} \odot \mathbf{v}$, since a symmetric decomposition assigns $R_{\mathbf{u}} + R_{\mathbf{v}} = 2 R_{\mathbf{o}}$ and therefore doubles the total relevance at each such node, and analogously at the SiLU, whose factorization $x\cdot\sigma(x)$ is itself bilinear. MambaLRP restores Eq.~\eqref{eq:conservation} by treating the gating factor as a constant multiplier, i.e.\ by detaching it from the backward graph and enforcing:
\begin{equation}
    R_{\mathbf{u}} = R_{\mathbf{o}}, \qquad R_{\mathbf{v}} = \mathbf{0},
    \label{eq:mambalrp_gate}
\end{equation}
so that relevance flows exclusively along the value path and $\sum_{t=1}^T R_{\mathbf{x}_t} = z_{c}(x)$ is preserved layer by layer. The same treatment applied to $\sigma(x)$ inside the SiLU makes the activation effectively linear in the backward pass.
The authors propose modifications to the LRP framework to accommodate these unique architectural elements of Mamba.

In~\cite{latim}, the authors attempt to address a fine-grained interpretability problem in Mamba models, going some way towards addressing the limitations previously exposed in~\cite{mambalrp}. Specifically, they seek to understand how these models selectively process token-level sequences across network layers. The approach presented is called LaTIM, a novel token-level decomposition method that reformulates the SSM computation to enable token-by-token analysis, allowing it to adapt attention-based interpretability techniques such as Aggregation of Layer-wise Token-to-Token Interactions (ALTI)~\cite{ferrando2022measuring}. Based on experiments conducted on tasks such as machine translation and retrieval-based generation, a direct comparison with the previously cited~\cite{mamba_vit} and~\cite{mambalrp} studies is also proposed.
The results obtained highlight the validity of the presented approach, demonstrating that the resulting heatmaps are more informative than those generated by MambaLRP and Mamba-Attention (which capture the general structure but are not precise in fine-grained analysis). The results suggest that Mamba's memory limitations in long contexts arise from a sparse and weak focus on relevant tokens.
The main limitation of LaTIM is that, due to the nonlinearity introduced by SiLU, token decomposition requires an approximation strategy. While this is empirically valid, an exact decomposition would require model modifications and retraining. Consequently, improving the robustness and reliability of Mambas remains an open problem.

A recent comprehensive survey by \cite{medmamba} analyzes the paradigm shift across CNNs, ViTs, and Mamba models. The authors highlight the limitations of CNNs in modeling long-range dependencies and context, a limitation addressed by ViTs, which, however, require large computational resources.
Since Mamba's SSM is capable of modeling long-range dependencies while maintaining linear complexity, the authors define Mamba as a combination of RNNs and CNNs. In \cite{medmamba}, the first Mamba model equipped with explainability approaches for medical tasks is proposed. Regarding XAI problems, they use t-SNE (t-distributed Stochastic Neighbor Embedding) and Grad-CAM to improve the model's reliability. The results obtained confirm Mamba's superior ability to extract more representative and discriminative features compared to CNN and ViT.

\subsection{Gaze-based Explainable AI}

A crucial frontier in XAI is the quantitative validation of saliency maps using human visual attention as a ground-truth benchmark. While many studies, such as Alqutayfi \emph{et al.}~\cite{related_CNN_ViT}, rely on qualitative expert assessments to validate model focus, there is a growing shift toward objective metrics. Ruiu \emph{et al.}~\cite{metrics} formalized this by using similarity measures (e.g., Jensen-Shannon) to compare ViT attention with human fixations, though their scope was limited to facial analysis.
The application of gaze data varies significantly in the literature. Some works, like Kang \emph{et al.}~\cite{KANG2026102356}, use gaze-alignment metrics (LPIPS, cosine similarity) to validate how CNNs perceive urban environments. Others, such as Govil \emph{et al.}~\cite{govil_2025}, integrate human attention directly into the training process to supervise the model's gaze.

A similar work by Hwu \emph{et al.}~\cite{hwu2021matching} explores the correlation between human eye gaze and visual explanations generated via LRP in the domain of autonomous driving. By formally treating human gaze heatmaps as a quantitative ground truth, they measure the alignment of model attention across different training stages using correlation metrics such as Cosine Similarity and Spearman's rank. Furthermore, they observe that despite producing heatmaps that are spatially more diffuse and noisy than human gaze, LRP has the potential to identify task-relevant stimuli in an emergent manner, emphasizing both low-level and object-based features without relying on explicit object segmentation or bounding box labels. However, while their study quantitatively demonstrates that highly task-specific models produce saliency maps more aligned with human attention, their analysis remains strictly limited to a single CNN architecture (VGG16).

In~\cite{kumar2025seeing}, Kumar \emph{et al.} extended gaze-based validation to the temporal domain by comparing human fixations with the attention maps of a CNN+Transformer model trained on video memorability. Similarly to Hwu \emph{et al.}~\cite{hwu2021matching}, they treat human gaze as a quantitative benchmark, utilizing metrics such as AUC-Judd and Normalized Scanpath Saliency to show how models naturally mirror human spatial attention. Their analysis goes further by employing panoptic segmentation to reveal a shared semantic hierarchy: both humans and models prioritize ``things'' (discrete objects) over ``stuff'' (amorphous background). Their work shows how, despite no architectural bias, the model's emergent behavior aligns with the human tendency to assign greater importance to the initial frames of a visual event.

In the next sections, Grad-CAM, LRP, Score-CAM, Ablation-CAM, and other XAI methods are applied to different architectures, including CNN, ViT, and Mamba, to compare and validate their results. The goal is to identify which XAI approach is best suited for each architecture and, if necessary, determine whether there is an overall superior approach.

\section{Datasets}
\label{sec:datasets}

In this study, we selected datasets from completely different domains to evaluate different models. In particular, the datasets also include, for each sample, gaze heatmaps recorded for each image via eye-tracking. 
This choice is not accidental, since the study of the human gaze is a key element in understanding the extent to which models can replicate human perceptual processes.

The importance of having data relating to the human gaze lies in the large amount of information they contain.
Given the overwhelming volume of data and objects around us, it is necessary to ensure that the available resources are dedicated to what is most important to understand the context in which we find ourselves, since any model, whether computational, biological or digital, has limited capabilities.

For this reason, in recent years the demand for data of this type has grown significantly in the fields of computer vision, natural language processing, decision-making learning, and robotics.
In the field of computer vision, in particular, it is essential to extract the most ``salient'' visual characteristics from images~\cite{humangaze}.

However, since eye-tracking devices and software are expensive, datasets providing this type of freely accessible data are scarce in the literature. Therefore, in the following, we describe in more detail the datasets used in this study, which include annotations of human gazes in support of our analysis.

\subsection{CUB dataset}
\label{cub}
This dataset consists of 11,788 images ($448\times448$) belonging to 200 species of birds (about 60 images for each species)~\cite{welinder2010caltech}.
It was chosen because it has been widely used in several computer vision and explainable AI fine-grained classification tasks~\cite{finegrained, visualexplain, interpretable}
The peculiarities of this dataset lie in the local differences between each class, consequently making human attention more unified than when observing significantly different classes that have scattered differences.

Regarding the human gaze data, those provided by~\cite{cubgaze} were used. These were collected and analyzed using the Tobii Spectrum Eye Tracker (1200~Hz) with chin support to ensure greater precision during eye recordings. Twenty-five subjects participated, all with no specific knowledge of ornithology, to collect the gaze data. In particular, each participant is shown two images belonging to different classes but with similar characteristics to each other, and must observe the distinctive differences between them. In a second phase, a single photo belonging to one of the two previous classes is shown, and the participant must classify the image. This second phase lasts only 3 seconds, so that the participant focuses only on the salient areas of the image. 
The raw gaze data is preprocessed to extract fixation locations using the Velocity-Threshold Identification (I-VT) algorithm~\cite{olsen2012tobii}. The resulting fixation points offered in the dataset include coordinates and duration information. Based on this information, saliency maps for human gaze are generated. More information can be found in the original paper~\cite{cubgaze}. It is worth noticing that relying on non-expert participants introduces a potential bias related to the specific image pairs shown during the sessions. We hypothesize that, while an ornithologist would possess a priori knowledge of which anatomical details to observe, novices may rely entirely on the immediate visual comparison. Consequently, the precision of their attention is highly dependent on the similarity of the compared birds; comparing highly similar species might guide their gaze toward precise, discriminative features, whereas comparing more visually distinct species could result in a more scattered attention.

\subsection{CXR dataset}
This dataset consists of 1083 chest x-ray images ($3000\times3000$) from the MIMIC-CXR dataset~\cite{ karargyris2021creation, DBLP:journals/corr/abs-1901-07042}. The images are classified into three distinct categories: Normal (indicating a healthy condition), Pneumonia, and Congestive Heart Failure (CHF). This dataset was selected because it differs significantly from the one described in Subsection \ref{cub}, allowing us to evaluate the same models and explainability methods within a medical domain. Many studies~\cite{lungcancer, DBLP:journals/corr/abs-2005-13799} have used explainability techniques to validate classification results produced by neural network architectures, without having access to ground-truth human gaze data to assess the findings. Instead, for the creation of this dataset, the fixation data were acquired using a Gazepoint GP3 eye-tracking system (60~Hz sampling rate) mounted on a monitor with a resolution of $1920\times1080$. The data collection involved a single American Board of Radiology certified radiologist with over five years of experience. To ensure an objective visual assessment, the expert performed a ``cold reading'' of 1,083 frontal chest X-rays, blinded to the patients' specific clinical history aside from the emergency department setting. The recording process was divided into sessions of 30 cases to minimize fatigue, with a 9-point calibration procedure performed at the beginning of each session to ensure tracking accuracy. More information can be found in the original paper~\cite{karargyris2021creation}.

\subsection{Uniss-FGD dataset}
\label{uniss}
This dataset introduces a completely different domain focusing on human faces and expressions. It consists of 120 high-quality facial images selected from the Karolinska Directed Emotional Faces (KDEF) database~\cite{lundqvist1998karolinska, metrics}. In the following, we use the term KDEF-FGD to refer to the dataset comprising both the images and their corresponding human gaze annotations. The images depict 40 different individuals (20 female and 20 male), each displaying three distinct expressions: happy, sad, and neutral. The inclusion of this dataset serves to evaluate how models and explainability methods behave when confronted with human faces, a highly salient stimulus that naturally drives human visual attention.
The human gaze data were collected using a Tobii Pro TX300 Eye Tracker, which ensures high precision with a sampling frequency of 300~Hz. A total of 20 observers (10 female and 10 male), recruited among university students and academic staff, participated in the recording sessions. During the experiment, participants were asked to freely view the sequence of faces, with each image displayed on the screen for exactly 3~seconds. This free-viewing task, without specific recognition constraints, allows capturing natural human attention patterns as they vary across different facial expressions~\cite{metrics}.
Notably, since recognizing basic human emotions does not require specific prior training or domain expertise, relying on a general audience reliably reflects innate human visual processing. However, it is essential to consider the demographic constraints of the dataset, as the original study highlights that both culture and age significantly influence visual attention patterns on faces. Since the data were predominantly gathered from a cohort of young Western adults (mostly aged 19–24), the resulting gaze patterns might not generalize to non-Western populations or to different age groups, such as children and the elderly~\cite{metrics}. The raw gaze data were then processed to extract precise fixations and saccades.

\begin{table*}[t]
\caption{Dataset details.}\label{tab:datasets}
\begin{tabular*}{\tblwidth}{@{} LLLLLL@{} }
\toprule
\textbf{Dataset} & Task & \textbf{Classes} & \textbf{Training/Test Nums} & \textbf{Reference} & \textbf{Gaze Download Link} \\
\midrule
CUB-200-2011 & Birds Classification & 200 & 5995/5794 & \cite{wah2011cub} & \cite{cubgazedataset} \\ 
CXR & Chest X-ray Classification & 3 & 976/107 & \cite{karargyris2021creation} & \cite{gazecxrdataset}\\
KDEF & Facial Emotion Classification & 7 & 3920/980 & \cite{lundqvist1998karolinska} & \cite{metrics}\\
\bottomrule
\end{tabular*}
\end{table*}

\begin{figure*}[t]
    \centering
    \captionsetup[subfigure]{labelformat=parens, labelsep=space, font=scriptsize}
    
    \newcommand{\commonheight}{2.2cm}
    
    \newcommand{\fixedgap}{\hspace{0.15cm}}

    \subcaptionbox{}{\includegraphics[height=\commonheight]{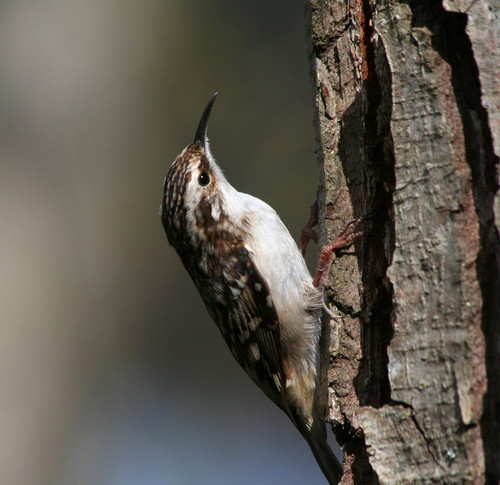}}%
    \fixedgap%
    \subcaptionbox{}{\includegraphics[height=\commonheight]{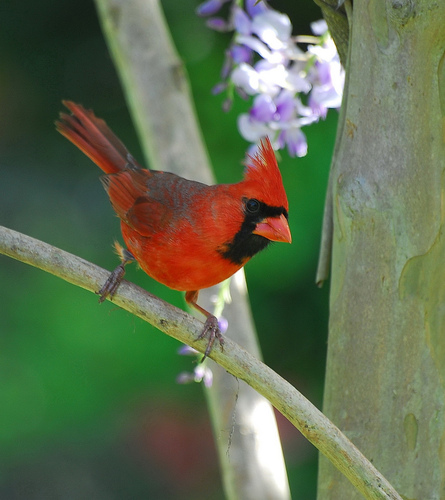}}%
    \fixedgap%
    \subcaptionbox{}{\includegraphics[height=\commonheight]{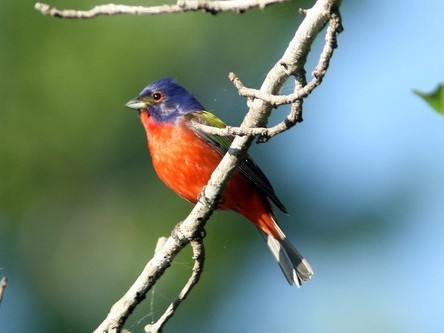}}%
    \fixedgap%
    \subcaptionbox{}{\includegraphics[height=\commonheight]{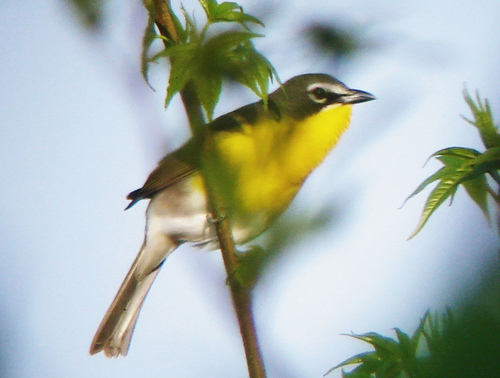}}%
    \fixedgap%
    \subcaptionbox{}{\includegraphics[height=\commonheight]{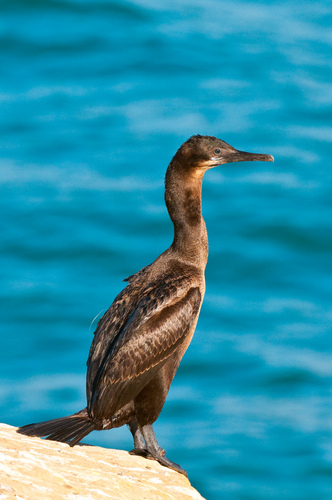}}%
    \fixedgap%
    \subcaptionbox{}{\includegraphics[height=\commonheight]{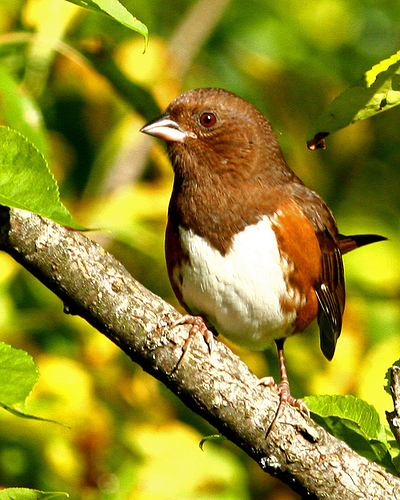}}%

    \vspace{0.2cm} 

    \subcaptionbox{}{\includegraphics[height=\commonheight]{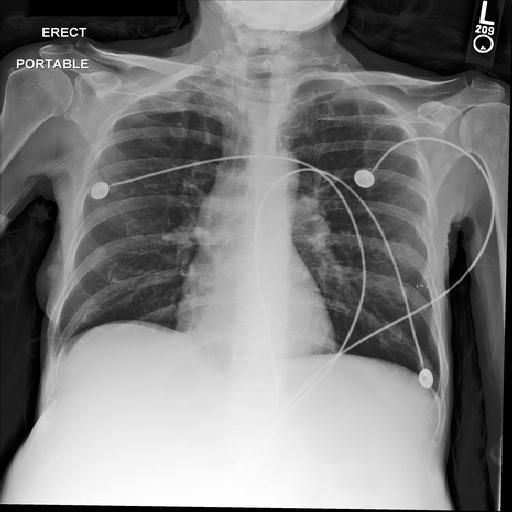}}%
    \fixedgap%
    \subcaptionbox{}{\includegraphics[height=\commonheight]{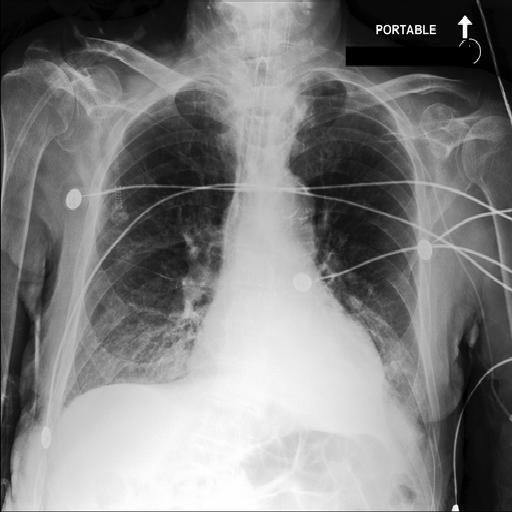}}%
    \fixedgap%
    \subcaptionbox{}{\includegraphics[height=\commonheight]{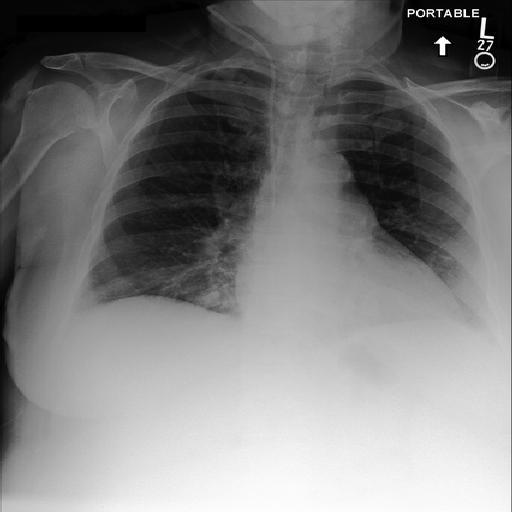}}%
    \fixedgap%
    \subcaptionbox{}{\includegraphics[height=\commonheight]{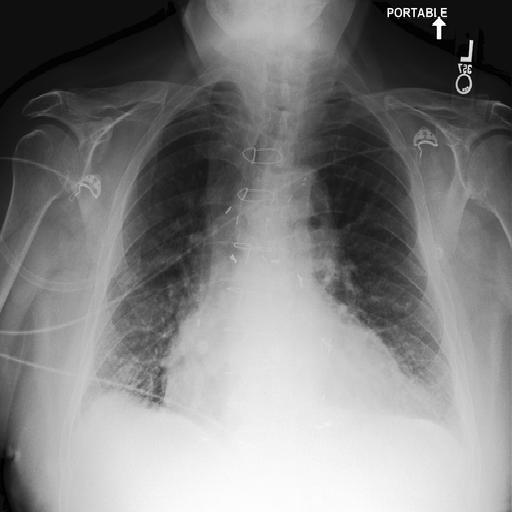}}%
    \fixedgap%
    \subcaptionbox{}{\includegraphics[height=\commonheight]{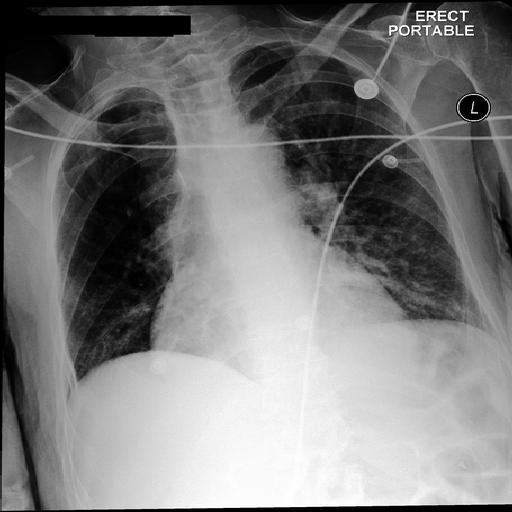}}%
    \fixedgap%
    \subcaptionbox{}{\includegraphics[height=\commonheight]{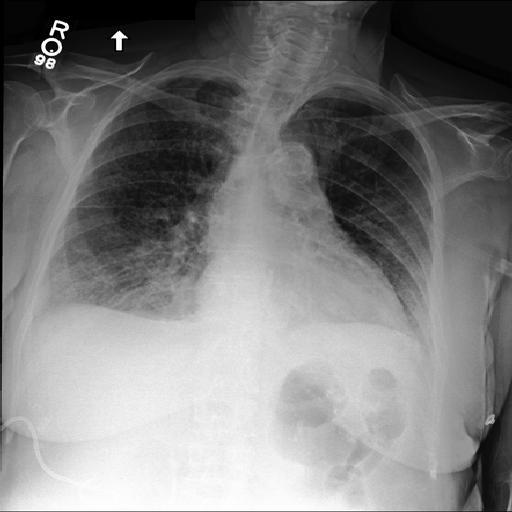}}%

    \vspace{0.2cm} 

    \subcaptionbox{}{\includegraphics[height=\commonheight]{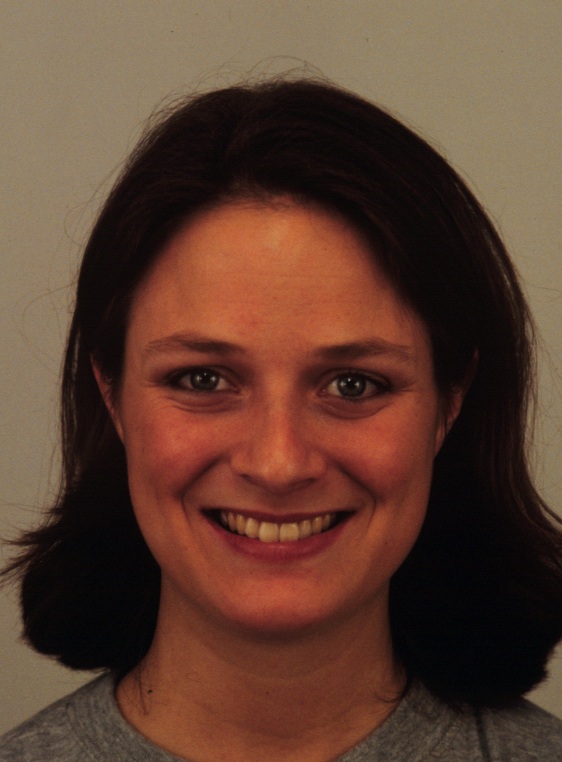}}%
    \fixedgap%
    \subcaptionbox{}{\includegraphics[height=\commonheight]{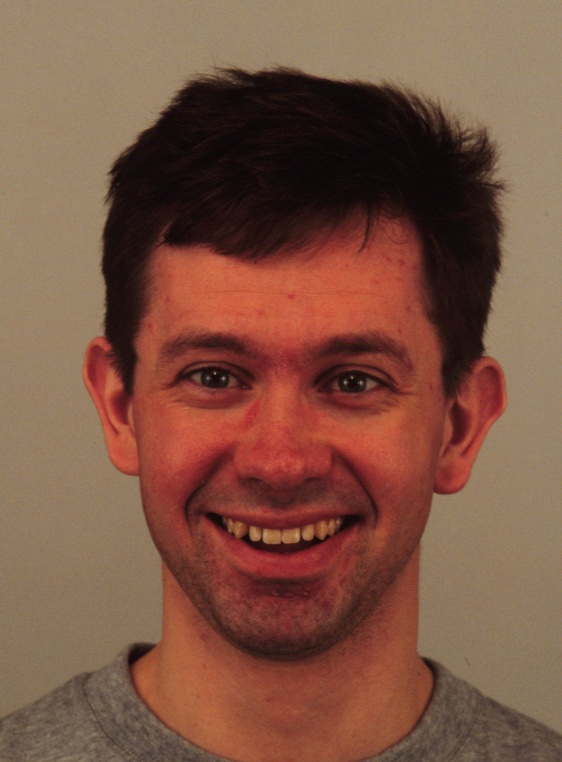}}%
    \fixedgap%
    \subcaptionbox{}{\includegraphics[height=\commonheight]{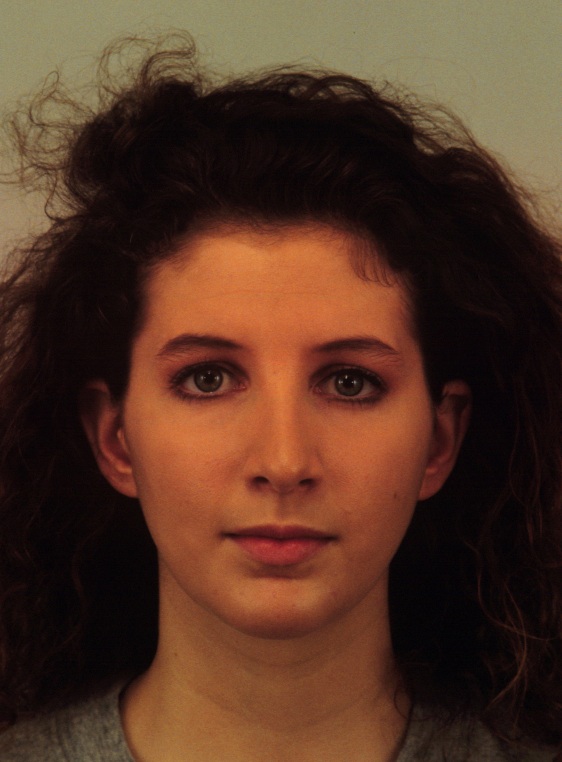}}%
    \fixedgap%
    \subcaptionbox{}{\includegraphics[height=\commonheight]{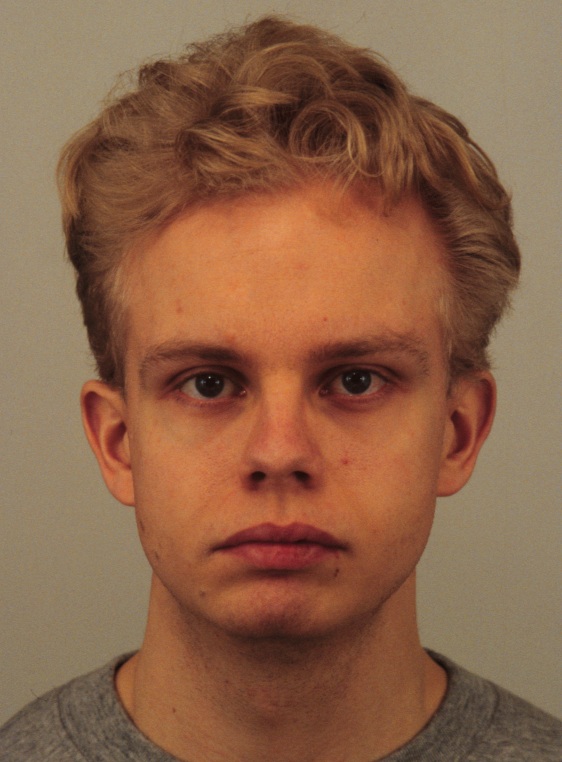}}%
    \fixedgap%
    \subcaptionbox{}{\includegraphics[height=\commonheight]{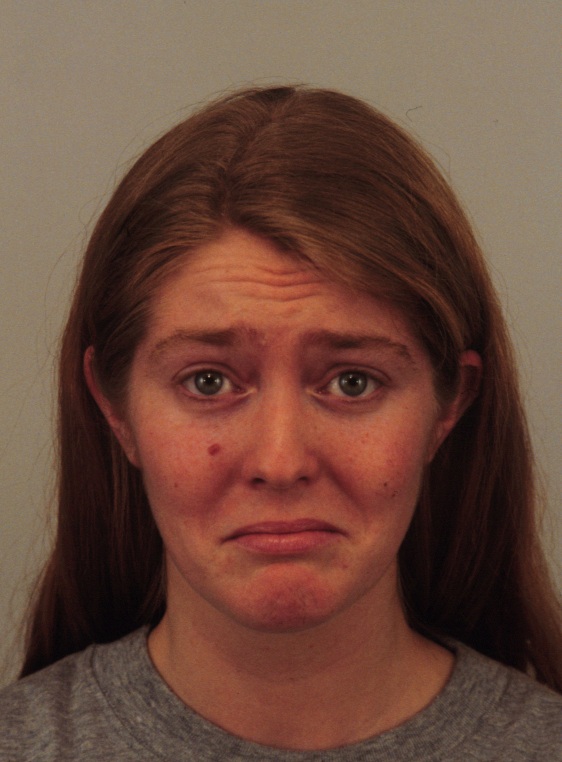}}%
    \fixedgap%
    \subcaptionbox{}{\includegraphics[height=\commonheight]{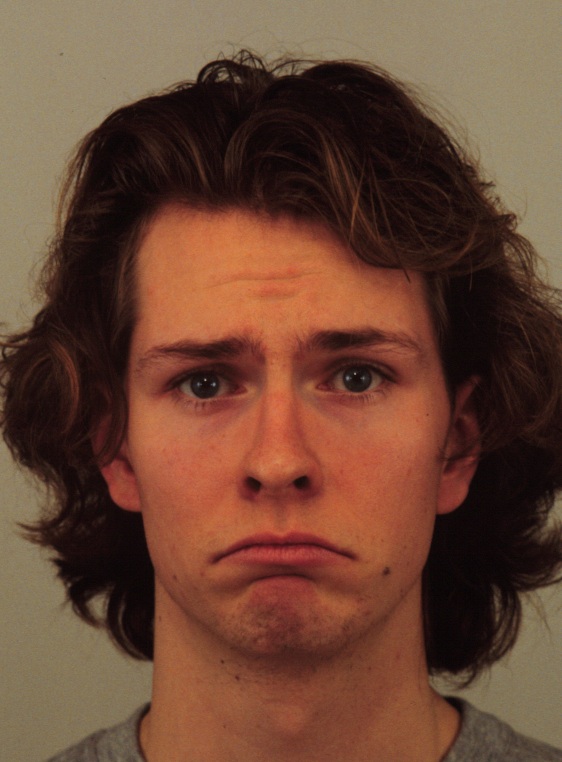}}

    \caption{Qualitative comparison of the datasets. \textbf{CUB}: (a) Brown Creeper, (b) Cardinal, (c) Painted Bunting, (d) Yellow Breasted Chat, (e) Pelagic Cormorant, (f) Eastern Towhee. \textbf{CXR}: (g) No acute cardiopulmonary process, (h) Stable mild pulmonary edema and moderate cardiomegaly, (i) Mild pulmonary edema, (j) Mild pulmonary vascular congestion and cardiomegaly, (k) No pneumonia. There is mild asymmetric pulmonary edema in the left lung, (l) Mild to moderate pulmonary edema. \textbf{KDEF}: (m) Happy Female, (n) Happy Man, (o) Neutral Female, (p) Neutral Male, (q) Sad Female, (r) Sad Male.}
    \label{fig:datasets_comparison}
\end{figure*}

\section{Models}
\label{sec:models}
Regarding the CNN used for the experiments, a \textbf{ResNet-50} pre-trained on ImageNet~\cite{resnet} was used.
This choice was dictated by the fact that through skip connections, the gradient flows more easily during backpropagation. This solves the vanishing gradient problem that can affect deeper networks, such as VGG.
To correctly use the network with the different datasets, the last layer (adapted to the 1000 ImageNet classes) was removed and replaced with a new linear layer to classify the required classes. The new layer is initialized with Kaiming (He Initialization) to maintain stable variance at the beginning of training~\cite{initialization}. This method accounts for the non-linearity of ReLU activation functions by scaling the weights according to the number of input units, thereby preventing the signal magnitude from diminishing as it propagates through the final stages of the network.
Images are resized to $448\times448$ pixels and augmentation is performed only with random resized crop and random horizontal flip. To analyze and understand network decisions, the XAI techniques used are Grad-CAM~\cite{gradcam}, Score-CAM~\cite{scorecam} and Ablation-CAM~\cite{ablationcam}, i.e.\ the operators defined in Eqs.~\eqref{eq:gradcam}, \eqref{eq:scorecam_cic} and \eqref{eq:ablationcam}, all instantiated on the output of the last convolutional stage ($h=w=14$ at the input resolution used here).

Regarding Vision Mamba, we used the \textbf{ViM-Base} model pretrained on ImageNet-1K, whose layers implement the selective recurrence of Eq.~\eqref{eq:ssm_discrete} applied to the sequence of image patches scanned along both directions. We also experimented with smaller variants (e.g., ViM-Tiny and ViM-Small); however, their accuracy on the evaluated datasets was substantially lower, preventing a meaningful analysis of their explainability.
For Vision Mamba fine-tuning, different steps are performed:
\begin{enumerate}
    \item \textbf{Input}: To ensure compatibility between the pretrained model and the new architecture, the positional embeddings are interpolated to match the spatial resolution required by the current patch configuration. The extra tokens (such as class tokens) are preserved, while only the spatial positional tokens are reshaped and resized using bicubic interpolation. This procedure allows the model to reuse pretrained positional information even when the number of patches differs from that of the original checkpoint (in our experiments, from $224\times224$ to $448\times448$). 
    \item \textbf{Output}: The last layer was removed and replaced by a new linear layer to classify the required classes. Xavier uniform initialization was applied to initialize these new model parameters. This choice accounts for the linear transformations and gating mechanisms of the Mamba architecture. By preserving signal variance across the state-space sequences, Xavier initialization prevents parameter saturation and ensures stable gradient flow, which is critical for balancing the dynamics of the classification head with the pretrained SSM layers.
    \item \textbf{Inner layers}: All model parameters were frozen except for the head and the \texttt{[CLS]} token. In additional experiments, we explored partially unfreezing only the SSM layers, as well as applying LoRA (Low-Rank Adaptation)~\cite{lora}, a parameter-efficient fine-tuning technique, to the projection layer; however, these strategies did not yield significant performance improvements.
\end{enumerate}
We used the Adam optimizer with decoupled weight decay regularization~\cite{Loshchilov2017DecoupledWD}, rather than the classic stochastic gradient descent with momentum. To analyze and understand network decisions, the XAI technique used is MambaLRP~\cite{mambalrp}, with the conservation-preserving rule of Eq.~\eqref{eq:mambalrp_gate}.

Regarding Transformer-based models, we adopted the \textbf{ViT-Base} architecture~\cite{vit}, pretrained on the ImageNet 2012 dataset. For this model, we followed a fine-tuning strategy similar to the one used for ViM-Base. In particular, we employed bicubic interpolation to adapt the positional embeddings, allowing the use of higher-resolution input images during the fine-tuning phase without losing spatial consistency. Moreover, to reduce the risk of overfitting and to preserve the rich visual representations learned during pretraining, we froze all the backbone layers of the network. To adapt the architecture to the target task, we replaced the original MLP head with a new linear classification layer projecting the representations to the class logits. Only the classification head and the \texttt{[CLS]} token were left trainable, enabling the model to adapt to the target task while keeping the majority of the pretrained parameters fixed (in tables \ref{tab:explainability_evaluation_train}-\ref{tab:explainability_evaluation_test} with name \textbf{ViT-Base-F}, where F stands for frozen). The same architecture is trained without freezing any parameters. The Adam optimizer with cosine annealing learning rate schedule was used.
To analyze and understand the network's decisions, the XAI technique described in \cite{transformer_interpretability}, referred to in this paper as ViTLRP, i.e.\ the operator of Eq.~\eqref{eq:vitlrp}, is used.

The selection of vanilla architectures, namely ResNet-50, ViT-Base, and ViM-Base, rather than more complex, heavily optimized state-of-the-art variants, is a deliberate methodological choice driven by two primary factors. First, it ensures the isolation of the underlying computational paradigms (convolutional, attention-based, and state-space modeling, respectively) without introducing architectural biases or specialized modules that could compromise the comparative analysis. Second, the reliable application of XAI techniques remains an open challenge. Methods such as Grad-CAM and LRP rely on traceable gradient flows and standard structural assumptions to generate faithful attribution maps. Custom architectures risk violating these assumptions, making it difficult to discern whether an ambiguous saliency map stems from the model's opaque decision-making or a failure of the XAI algorithm itself. By utilizing these canonical baseline models, we guarantee a higher theoretical fidelity for the XAI methods, ensuring that our comparative evaluation reflects the true explainability of the core architectures.

\section{Experimental Analysis}
\label{sec:experiments}

This section presents the core empirical evaluation of our study, aimed at quantifying the alignment between machine-generated saliency maps and human visual attention. To move beyond purely qualitative assessments, we deploy an experimental framework across three representative architectural families—CNNs, ViTs, and Vision Mamba—and test them across three distinct visual domains: fine-grained natural classification, medical diagnostics, and facial emotion recognition. By testing architecture-specific XAI methods against empirical human eye-tracking ground truth, we systematically isolate how distinct network designs, backbone freezing, and ensemble strategies affect the fidelity of model explanations.

\subsection{Problem Formulation}
\label{sec:problem}

We now make precise the object that the remainder of this section estimates. The notation is collected in Table~\ref{tab:notation} and is used consistently throughout the experimental analysis.

\begin{definition}[Gaze dataset]
\label{def:dataset}
A \emph{gaze-annotated dataset} is a finite collection:
\begin{equation}
\begin{split}
    \dset=\bigl\{(x_{n},y_{n},\gaze_{n})\bigr\}_{n=1}^{N},&\\
    x_{n}\in\Xspace,\quad y_{n}\in\Yspace,\quad \gaze_{n}\in\R^{\Grid}_{\ge0},&
\end{split}
    \label{eq:dataset}
\end{equation}
in which $\gaze_{n}$ is the empirical fixation-density map recorded by an eye tracker on the stimulus $x_{n}$, obtained by kernel-smoothing the fixation points weighted by their Dwell\footnote{The total duration a user's gaze or cursor remains focused on a specific region.} time. The set is partitioned into disjoint subsets $\dset=\dset_{\mathrm{tr}}\,\dot\cup\,\dset_{\mathrm{te}}$.
\end{definition}

The three datasets of Section~\ref{sec:datasets} instantiate Definition~\ref{def:dataset} with different acquisition protocols, sampling frequencies, and observer populations; consequently, $\gaze$ is comparable \emph{within} a dataset but not across datasets, a caveat we return to in Section~\ref{sec:quantitative}.

Model parameters are obtained by empirical risk minimization of the categorical cross-entropy on the labels alone:
\begin{equation}
    \hat{\params}\in\arg\min_{\params\in\Theta}\ \frac{1}{|\dset_{\mathrm{tr}}|}\sum_{(x,y,\gaze)\in\dset_{\mathrm{tr}}}-\log \model_{\params}(x)_{y}\;,
    \label{eq:erm}
\end{equation}
We write $\model^{\mathrm{CNN}}$, $\model^{\mathrm{ViT}}$ and $\model^{\mathrm{ViM}}$ for the three architectural families of Section~\ref{sec:models}, and the feasible set $\Theta$ encodes the training intervention under study (full fine-tuning, or backbone freezing as defined in Section~\ref{sec:models}). Crucially, $\gaze$ does not appear in Eq.~\eqref{eq:erm}: the gaze annotations are never observed by the learner, so any agreement measured a posteriori is an emergent property of the representation rather than the result of supervision. This is the formal difference between our protocol and gaze-supervised approaches such as~\cite{govil_2025,zhu2022gazeguided}.

\begin{definition}[Attention alignment]
\label{def:alignment}
Write $\Pspace=\simplex^{|\Grid|-1}$ for the simplex over the pixel lattice, let $\mathcal{S}$ be a similarity functional on $\Pspace\times\Pspace$ (see Sec.~\ref{chp:measures}), and let $\proj$ denote the normalization pipeline of Eq.~\eqref{eq:pipeline} below. The \emph{alignment} of the pair $(\model_{\hat\params},\attr)$ over a subset $\dset'\subseteq\dset$ is
\begin{equation}
\begin{split}
    &\mathcal{A}_{\mathcal S}\bigl(\model_{\hat\params},\attr;\dset'\bigr)=\\[2pt]
    &\quad\frac{1}{|\dset'|}\sum_{(x,y,\gaze)\in\dset'}\mathcal{S}\Bigl(\proj\bigl[\sal_{x}\bigr],\,\proj[\gaze]\Bigr),
\end{split}
    \label{eq:alignment}
\end{equation}
where $\sal_{x}=\attr(\model_{\hat\params},x,\hat y(x))$ is the saliency map of the predicted class.
\end{definition}

Eq.~\eqref{eq:alignment} is the estimator reported, together with its empirical standard deviation, in Tables~\ref{tab:explainability_evaluation_train}--\ref{tab:explainability_evaluation_test}. With this notation, the experimental questions become unambiguous:
\begin{enumerate}
    \item[(Q1)] \emph{Architecture}: at a fixed domain, how does $\mathcal{A}_{\mathcal S}$ vary across $\model^{\mathrm{CNN}}$, $\model^{\mathrm{ViT}}$ and $\model^{\mathrm{ViM}}$?
    \item[(Q2)] \emph{Domain}: how does $\mathcal{A}_{\mathcal S}$ vary across the three datasets at fixed architecture?
    \item[(Q3)] \emph{Generalization}: is the gap between $\mathcal{A}_{\mathcal S}(\cdot;\dset_{\mathrm{tr}})$ and $\mathcal{A}_{\mathcal S}(\cdot;\dset_{\mathrm{te}})$ negligible, i.e.\ does the explanation generalize as the classifier does?
    \item[(Q4)] \emph{Interventions}: how do restrictions of $\Theta$ (backbone freezing) and aggregation rules (ensembling, Eq.~\eqref{eq:ensemble}) act on $\mathcal{A}_{\mathcal S}$?
\end{enumerate}
Finally, replacing $\proj[\gaze]$ in Eq.~\eqref{eq:alignment} by the normalized saliency map of a second model yields the model-to-model agreement reported in Table~\ref{tab:explainability_model_comparison}, which separates inter-architecture divergence from human alignment: two models may both be poorly aligned with $\gaze$ and yet agree with each other, or vice versa.

\begin{table}[t]
\centering
\caption{Summary of notation.}
\label{tab:notation}
\footnotesize
\begin{tabular}{@{}ll@{}}
\toprule
\textbf{Symbol} & \textbf{Meaning}\\
\midrule
$\Grid$, $|\Grid|$           & pixel lattice $\{1..H\}\times\{1..W\}$ and its size\\
$\Xspace$, $\Yspace$, $K$    & input space, label set, number of classes\\
$\simplex^{m-1}$              & probability simplex in $\R^{m}$\\
$\dset$, $\dset_{\mathrm{tr}}$, $\dset_{\mathrm{te}}$ & dataset and its train/test partition\\
$\model_{\params}$, $z_{c}$, $\hat y$ & classifier, class logit, predicted label\\
$\attr$                       & attribution operator (Def.~\ref{def:attribution})\\
$\sal=\attr(\model_\params,x,\hat y)$ & model saliency map\\
$\gaze$                        & human fixation-density map\\
$\resize$, $\norml$, $\proj$   & resampling, min--max scaling, simplex projection\\
$P$, $Q$                      & distributions induced by $\sal$ and $\gaze$\\
$\mathcal{S}$, $\mathcal{A}_{\mathcal S}$ & similarity functional and alignment (Def.~\ref{def:alignment})\\
$\odot$, $(\cdot)^{+}$         & Hadamard product, positive part $\max(\cdot,0)$\\
\bottomrule
\end{tabular}
\end{table}

\subsection{Similarity measures between saliency maps}
\label{chp:measures}
To compare two saliency maps (one from our models and one from human gazes), we convert each map into a probability distribution and then compute a similarity or distance measure between them. This conversion is not a formality: the three functionals below are defined on the simplex, so the pipeline that transports a raw attribution map onto it determines which properties of the explanation survive the comparison. We therefore state it explicitly and then characterize the resulting estimators.

\textbf{Normalization pipeline.} Let $\sal\in\R^{\Grid'}$ be a raw attribution map, possibly supported on a lattice $\Grid'$ coarser than the input lattice $\Grid$ (this is the case for the CAM operators of Eq.~\eqref{eq:gradcam}), and let $\gaze$ be the fixation-density map, provided as an $8$-bit grayscale image with values in $\{0,\dots,255\}$. We denote by $\proj$ the composition of the four steps
\begin{equation}
\begin{split}
    \text{(i)}\ \ &\hat{\sal}=\resize_{\Grid}(\sal),\\
    \text{(ii)}\ \ &\sal^{+}=\pospart{\hat{\sal}=\max\bigl(\hat{\sal},0\bigr),}\\
    \text{(iii)}\ \ &\bar{\sal}=\norml(\sal^{+})=\frac{\sal^{+}-\min_{u}\sal^{+}(u)}{\max_{u}\sal^{+}(u)-\min_{u}\sal^{+}(u)},\\
    \text{(iv)}\ \ &P(u)=\proj[\sal](u)=\frac{\bar{\sal}(u)}{\sum_{v\in\Grid}\bar{\sal}(v)},
\end{split}
\label{eq:pipeline}
\end{equation}
where $\resize_{\Grid}$ is bilinear resampling onto $\Grid$ and $u,v$ index pixels. The distribution associated with the human reference, $Q=\proj[\gaze]$, is obtained by the same pipeline, in which step (ii) acts as the identity because $\gaze\ge0$ by construction. Step (ii) is likewise inactive for every operator whose output is non-negative by design, i.e.\ for all CAM variants (Eq.~\eqref{eq:gradcam}) and for ViTLRP (Eq.~\eqref{eq:vitlrp}); it is effective only for MambaLRP, whose relevances are signed, and its role is discussed in Remark~\ref{rem:positivepart}. By construction $P,Q\in\Pspace$, so that $\sum_{u}P(u)=\sum_{u}Q(u)=1$ and $P,Q\ge0$.

\begin{remark}[Positive part]
\label{rem:positivepart}
MambaLRP returns signed relevances, positive values denoting evidence for the predicted class and negative values denoting counter-evidence. Since human fixation densities are non-negative, no meaningful comparison of counter-evidence is possible, and step (ii) restricts the comparison to the class-supporting component of the explanation. This is a deliberate restriction of the hypothesis being tested, not a numerical artifact: what is measured for ViM is the alignment of $\pospart{\sal}$ with $\gaze$, and any evidence carried by $\pospart{-\sal}$ is discarded for all architectures alike.
\end{remark}


Throughout, we adopt the standard continuous extensions $0 \log 0 \equiv 0$ and $0 / 0 \equiv 0$, ensuring that all information-theoretic and ratio-based functionals defined below remain well-defined across the entire probability product space $\Pspace\times\Pspace$.

Following~\cite{metrics}, we adopt the following metrics:

\begin{enumerate}
    \item {Jensen-Shannon Similarity (JSS)}:

    The \emph{Jensen-Shannon Divergence} (JSD) measures the dissimilarity between two probability distributions and is defined as:
    \begin{equation}
        \mathrm{JSD}(P,Q)=\tfrac{1}{2}\KL(P\,\|\,M)+\tfrac{1}{2}\KL(Q\,\|\,M),
        \label{eq:jsd}
    \end{equation}
    where $M=\frac{1}{2}(P+Q)$~\cite{JSD}\ and $\KL$ is the Kullback--Leibler divergence, which we evaluate in base two:
    \begin{equation}
        \KL(P\,\|\,M)=\sum_{u\in\Grid}P(u)\log_{2}\frac{P(u)}{M(u)} .
        \label{eq:kld}
    \end{equation}

    Given this, the \emph{Jensen-Shannon Similarity} (JSS) is defined as:
    \begin{equation}
        \JSS(P,Q)=1-\mathrm{JSD}(P,Q).
        \label{eq:jss}
    \end{equation}

    A JSS value close to 1 indicates high similarity between the two saliency maps.
    In summary, this metric aims to quantify the similarity between the entire probability distributions, indicating whether the global high-attention areas (saliency peaks) are shared between the model and humans.

    \item {Chi-square Similarity}~(CSS): It can be seen as a normalized variant of Chi-square distance, which measures the divergence between two distributions. It is computed as:
    \begin{equation}
        \CSS(P,Q)=1-\sum_{u\in\Grid}\frac{\bigl(P(u)-Q(u)\bigr)^{2}}{P(u)+Q(u)} .
        \label{eq:css}
    \end{equation}

    A value close to 1 indicates high similarity. This metric heavily penalizes discrepancies between $P(u)$ and $Q(u)$ when they occur in areas where the total saliency activity ($P(u)+Q(u)$) is low, thus emphasizing accuracy even in less globally salient details.

    \item{Pearson Correlation Coefficient (PCC)}:
    The Pearson correlation coefficient measures the linear correlation between two distributions, treated here as two real-valued signals indexed by the pixel lattice:
    \begin{equation}
        \PCC(P,Q)=\frac{\sum_{u\in\Grid}\bigl(P(u)-\mu_{P}\bigr)\bigl(Q(u)-\mu_{Q}\bigr)}{\bigl[\sum_{u}(P(u)-\mu_{P})^{2}\,\sum_{u}(Q(u)-\mu_{Q})^{2}\bigr]^{1/2}},
        \label{eq:pcc}
    \end{equation}
    where $\mu_{P}=|\Grid|^{-1}\sum_{u}P(u)$ and $\mu_{Q}$ are the spatial means. A useful consequence of step (iv) of Eq.~\eqref{eq:pipeline} is that $\mu_{P}=\mu_{Q}=|\Grid|^{-1}$ exactly, so that Eq.~\eqref{eq:pcc} compares two centered signals of identical mean and $\PCC$ is entirely determined by the spatial structure of the two maps.
    A value of $\PCC$ close to 1 indicates a strong positive correlation between the two saliency maps. In general, a high value indicates that salient and non-salient points occur in the same relative order and position, evaluating the structural alignment of the maps.

\end{enumerate}

The three functionals are not redundant, and the following elementary observation makes the division of labor between them explicit.

\begin{proposition}[Complementarity]
\label{prop:complementarity}
Let $P,Q\in\Pspace$. Then: (a) $\JSS$ and $\CSS$ are separable, being sums of pointwise terms, and are invariant under any permutation of $\Grid$ applied jointly to both arguments, but not under a monotone reparameterization of the intensity of a single argument; (b) $\PCC$ is invariant under the affine maps $P\mapsto aP+b$ with $a>0$ applied to either argument, hence it is blind to intensity scale and offset and measures only structural co-variation; (c) consequently, two maps supported on the same region but with unrelated intensity profiles may yield $\JSS,\CSS\to1$ together with $\PCC\to0$, whereas two maps with proportional profiles on disjoint supports yield $\CSS=-1$ with $\PCC$ near zero.
\end{proposition}

Statements (a) and (b) are immediate from Eqs.~\eqref{eq:jss}, \eqref{eq:css} and~\eqref{eq:pcc}, and (c) follows by combining them with the known facts $\JSS(P,Q)\in[0,1]$ and $\CSS(P,Q)\in[-1,1]$. In operational terms, $\JSS$ measures the agreement of the two distributions as a whole, $\CSS$ isolates the discrepancy of mass, and $\PCC$ isolates the agreement of spatial structure up to intensity. Reporting the three jointly is therefore what allows the dissociation observed on the medical dataset, where a model attains near-maximal $\JSS$ and $\CSS$ together with a vanishing $\PCC$, to be interpreted rather than merely recorded (Section~\ref{sec:quantitative}).

\textbf{Aggregation.} For a fixed pair $(\model_{\hat\params},\attr)$, a fixed subset $\dset'$ and a fixed functional $\mathcal{S}\in\{\JSS,\CSS,\PCC\}$, let $m_{n}=\mathcal{S}(P_{n},Q_{n})$ be the per-image score. Tables~\ref{tab:explainability_evaluation_train}--\ref{tab:explainability_model_comparison} report:
\begin{equation}
\begin{split}
    \hat\mu&{}=\frac{1}{|\dset'|}\sum_{n=1}^{|\dset'|}m_{n},\\
    \hat\sigma&{}=\Bigl(\tfrac{1}{|\dset'|-1}\textstyle\sum_{n=1}^{|\dset'|}\bigl(m_{n}-\hat\mu\bigr)^{2}\Bigr)^{1/2},
\end{split}
    \label{eq:aggregation}
\end{equation}
in the form $\hat\mu\pm\hat\sigma$, where $\hat\mu$ is exactly the empirical alignment $\mathcal{A}_{\mathcal S}$ of Eq.~\eqref{eq:alignment} and $\hat\sigma$ quantifies its dispersion over the images, i.e.\ the stability of the explanation across the domain. Agreement among the three functionals is assessed through the rank correlation:
\begin{equation}
    \rho_{\mathcal S,\mathcal S'}=\PCC\bigl(\mathrm{rk}(m^{\mathcal S}),\,\mathrm{rk}(m^{\mathcal S'})\bigr),
    \label{eq:spearman}
\end{equation}
i.e.,\ Spearman's coefficient, computed by applying Eq.~\eqref{eq:pcc} to the rank vectors $\mathrm{rk}(\cdot)$ of the per-image scores, which detects monotone rather than merely linear agreement and is insensitive to the different ranges established by the range of values of JSS and CSS.


\subsection{Model validation}
\label{sec:model_validation}
Before presenting the quantitative evaluation of the explainability methods, we first provide an overview of the classification capabilities of the trained models. The primary objective of this initial validation is to demonstrate that the baseline architectures, despite lacking the extreme optimizations of highly specialized state-of-the-art variants, achieve solid and competitive classification performance. Establishing this baseline is crucial, as it legitimizes the subsequent XAI benchmark: interpreting model attention against human gaze is only meaningful if the underlying model has learned robust and accurate representations of the data.

\subsubsection{Classification Metrics}
\label{sec:classification_metrics}

The \textit{AUC (Multi-Class OVR)} metric is computed by applying the One-vs-Rest (OVR) strategy to the multi-class setting. For each class $c \in \{1, \dots, K\}$, a binary classifier is formed by considering class $c$ as the positive class and all other classes as the negative class. A ROC curve is then generated for each class, and its corresponding AUC value, denoted as $\mathrm{AUC}_c$, is computed. Denoting by $\mathcal{I}^{+}_{c}=\{n:y_{n}=c\}$ and $\mathcal{I}^{-}_{c}=\{n:y_{n}\neq c\}$ the positive and negative index sets and by $s_{n,c}=\model_{\hat\params}(x_{n})_{c}$ the score assigned to class $c$, the class-wise AUC admits the Mann--Whitney form:
\begin{equation}
\begin{split}
    \mathrm{AUC}_{c}&{}=\frac{1}{|\mathcal{I}^{+}_{c}||\mathcal{I}^{-}_{c}|}\sum_{i\in\mathcal{I}^{+}_{c}}\sum_{j\in\mathcal{I}^{-}_{c}}\Bigl(\mathbf{1}\bigl[s_{i,c}>s_{j,c}\bigr]\\
    &\qquad\qquad\qquad\quad{}+\tfrac{1}{2}\,\mathbf{1}\bigl[s_{i,c}=s_{j,c}\bigr]\Bigr),
\end{split}
    \label{eq:auc_mw}
\end{equation}
i.e.\ it estimates the probability that a randomly drawn positive sample receives a higher score than a randomly drawn negative one, which is exactly the area under the corresponding ROC curve and makes the metric invariant to any strictly increasing transformation of the scores. The final multi-class AUC score is the average over all classes:
\begin{equation}
    \mathrm{AUC}_{\text{OVR}} = \frac{1}{K} \sum_{c=1}^{K} \mathrm{AUC}_c
\end{equation}

Consistent with the AUC computation, Precision, Recall, and F1-score are extended to the multi-class setting using the same One-vs-Rest strategy. For each class $c$, we compute the class-specific True Positives ($TP_c$), False Positives ($FP_c$), and False Negatives ($FN_c$). Here, $TP_c$ represents samples correctly classified as $c$, $FP_c$ represents samples incorrectly assigned to class $c$, and $FN_c$ represents samples belonging to class $c$ but misclassified as another class; in set notation:
\begin{equation}
\begin{split}
    TP_{c}&{}=\bigl|\{n:\hat y(x_{n})=c,\ y_{n}=c\}\bigr|,\\
    FP_{c}&{}=\bigl|\{n:\hat y(x_{n})=c,\ y_{n}\neq c\}\bigr|,\\
    FN_{c}&{}=\bigl|\{n:\hat y(x_{n})\neq c,\ y_{n}=c\}\bigr| .
\end{split}
    \label{eq:confusion}
\end{equation}
The metrics for each class are defined as:
\begin{equation}
    \mathrm{Precision}_c = \frac{TP_c}{TP_c + FP_c}, \quad \mathrm{Recall}_c = \frac{TP_c}{TP_c + FN_c}
\end{equation}

The F1-score for class $c$ is the harmonic mean of Precision and Recall:
\begin{equation}
    \mathrm{F1}_c = 2 \cdot \frac{\mathrm{Precision}_c \cdot \mathrm{Recall}_c}{\mathrm{Precision}_c + \mathrm{Recall}_c} 
\end{equation}

The final global performance metrics reported in Tables~\ref{tab:comparison-cub}-\ref{tab:comparison-kdef} are computed as the macro-average over all $K$ classes (e.g., $\mathrm{F1}_{\text{macro}} = \frac{1}{K} \sum_{c=1}^{K} \mathrm{F1}_c$). The macro-average weights every class equally regardless of its cardinality, which is the appropriate choice for CUB-200-2011, whose $K=200$ classes are nearly balanced, and a deliberately conservative one for CXR, where the class frequencies are not. KDEF's $K=7$ emotion classes are balanced by design (70 subjects $\times$ 7 expressions), so the choice of averaging scheme is immaterial there.

\subsubsection{Classification Results}
\label{sec:classification_results}

The classification performance of the trained models, as evaluated on the test set, is reported in Tables~\ref{tab:comparison-cub}-\ref{tab:comparison-kdef}. 

When comparing our baseline classification results on the KDEF dataset (Table~\ref{tab:comparison-kdef}) to recent literature, it is important to contextualize state-of-the-art performance claims. For instance, the multi-domain fusion transformer proposed by Shu and Wang~\cite{sym18010015} achieves a remarkable accuracy of $99.7\%$ on KDEF. This impressive performance is obtained through a 70/30 train-validation split tailored for model selection and hyperparameter tuning, which effectively optimizes the architecture for that specific data distribution. Additionally, their framework introduces a sophisticated multi-branch design incorporating spatial, channel, and frequency domain fusions. While highly effective for maximizing nominal accuracy, such architectural complexity can pose inherent challenges for certain XAI techniques when tracing gradient flows, as discussed in Section~\ref{sec:models}. Therefore, maintaining standard canonical baselines in our experiments represents a deliberate methodological choice to ensure the theoretical fidelity and interpretability of the generated saliency maps, prioritizing architectural transparency for explanatory purposes.


\paragraph{Ensemble Aggregation (Q4).} We also investigated ensembles composed of pairs of neural networks, aiming to determine whether architectures with similar inductive biases lead to improved performance when combined, as well as investigating whether models exhibiting different explainability gazes can achieve improved classification performance when combined. We adopted a \emph{sum rule} strategy for ensemble inference: given the softmax output vectors of each model, these were summed element-wise, and the final prediction was obtained by applying the $\arg\max$ operator to the resulting vector. Formally, given $M$ models with softmax outputs $\mathbf{p}_k$, the ensemble prediction $\hat{y}$ is defined as:
\begin{equation}
   \hat{y}_{\mathrm{ens}}(x)=\arg\max_{c\in\Yspace}\ \frac{1}{M}\sum_{k=1}^{M}\model_{\hat\params_{k}}(x)_{c}\,,
   \label{eq:ensemble}
\end{equation}
where the normalization by $M$ is immaterial for the $\arg\max$ but makes the aggregate a member of $\simplex^{K-1}$, so that the ensemble is itself a classifier in $\mathcal{F}$ and Eq.~\eqref{eq:alignment} applies to it verbatim. We use $M$ rather than $K$ for the number of models to avoid a collision with the number of classes. Averaging posteriors attenuates the component of the estimation error that is zero-mean and uncorrelated across members, while leaving the shared bias untouched; the sum rule is therefore expected to help precisely when the errors of the members are weakly correlated, which is the hypothesis empirically tested below.

Experimental results show that, on the CUB dataset, the ensemble composed of a fine-tuned Transformer with frozen and unfrozen backbone models achieved higher performance compared to the corresponding individual models. This indicates that combining models with the same architecture but different training strategies can effectively improve generalization. Additionally, the ensemble formed by ViT and Mamba architectures also yielded competitive results, suggesting that these two models provide complementary representations despite their architectural differences. In contrast, the ensemble combining CNN-based models and ViTs did not lead to performance improvements. This outcome suggests that the representations learned by these architectures may be less complementary in this setting, limiting the effectiveness of simple ensemble aggregation strategies. However, experiments on the KDEF-FGD dataset reveal a significantly different behavior regarding cross-architecture ensembles. Unlike in CUB, combining a CNN-based model with either ViTs or Mamba architectures led to substantial performance gains. Notably, the ResNet50 + ViM-base ensemble achieved the highest accuracy among our configurations at 99.29\%, effectively outperforming the individual ResNet50 (96.32\%) and ViM-base (97.24\%) models. Similarly, the ResNet50 + ViT-base ensemble reached 99.08\%. This indicates that for the KDEF-FGD domain, the local feature extraction of CNNs and the global context capture of ViT or ViM provide highly complementary representations, making simple ensemble strategies highly effective.

\begin{table*}[t]
\centering
\caption{Comparison with existing works on CUB-200-2011 dataset.}
\begin{tabular*}{\tblwidth}{@{} LLLLLLL@{} }
\toprule
\textbf{Paper} & \textbf{Model} & \textbf{Accuracy (\%)} & \textbf{Precision (\%)} & \textbf{Recall (\%)} & \textbf{F1 (\%)} & \textbf{AUC (\%)} \\
\midrule
\cite{sakthi2025}           & InceptionV3   &         $65.56$  &         $45-93$  &         $57-95$  &         $40-91$  &             $-$  \\
\cite{sakthi2025}           & MobileNet     &         $74.60$  &         $65-81$  &         $37-96$  &         $71-94$  &             $-$  \\
\cite{yongshun2019finetune} & ResNet50      &         $84.05$  &             $-$  &             $-$  &             $-$  &             $-$  \\ 
\cite{transFG}              & TransFG       & $\mathbf{91.70}$ &             $-$  &             $-$  &             $-$  &             $-$  \\

\textit{Ours}               & Resnet50      &         $84.48$  &         $85.00$  &         $84.57$  &         $84.45$  &         $99.70$  \\
\textit{Ours}               & ViT-base      &         $87.59$  & $\mathbf{88.19}$ & $\mathbf{87.71}$ & $\mathbf{87.46}$ & $\mathbf{99.76}$ \\
\textit{Ours}               & ViT-base-F    &         $83.71$  &         $84.09$  &         $83.68$  &         $83.58$  & $\mathbf{99.76}$ \\
\textit{Ours}               & ViM-base      &         $84.31$  &         $85.22$  &         $84.34$  &         $84.18$  &         $99.62$  \\

\textit{Ours}                & Resnet50 + ViT-base                                       &         $87.57$ & $88.18$ & $87.68$ & $87.44$ & $99.45$ \\
\textit{Ours}                & Resnet50 + ViM-base                                        &        $82.19$ & $83.13$ & $82.23$ & $82.05$ & $99.30$ \\
\textit{Ours}                & ViT-base + ViM-base                                        &         $88.51$ & $89.10$ & $88.57$ & $88.43$ & $99.47$ \\
\textit{Ours}                & ViT-base + ViT-base-F                                        &         $88.68$ & $89.21$ & $88.77$ & $88.57$ & $99.73$ \\
\bottomrule
\end{tabular*}
\label{tab:comparison-cub}
\end{table*}

\begin{table*}[t]
\centering
\caption{Comparison with existing works on CXR dataset.}
\begin{tabular*}{\tblwidth}{@{} LLLLLLL@{} }
\toprule
\textbf{Paper} & \textbf{Model} & \textbf{Accuracy (\%)} & \textbf{Precision (\%)} & \textbf{Recall (\%)} & \textbf{F1 (\%)} & \textbf{AUC (\%)} \\
\midrule

\cite{saab2021observational} & GazeMTL trained by~\cite{wang2023gazegnn}       &         $78.50$  &          $78.60$ &          $78.10$ &          $77.90$ &          $88.70$ \\
\cite{gao2022IAA}            & IAA trained by~\cite{wang2023gazegnn}           &         $78.50$  &          $78.00$ &          $77.40$ &          $77.60$ &          $90.00$ \\
\cite{zhu2022gazeguided}     & EffNet+GG-CAM trained by~\cite{wang2023gazegnn} &         $77.57$  &          $77.00$ &          $77.20$ &          $77.00$ &          $88.80$ \\
\cite{wang2023gazegnn}       & GazeGNN                                         & $\mathbf{83.18}$ & $\mathbf{83.90}$ & $\mathbf{82.10}$ & $\mathbf{82.30}$ & $\mathbf{92.30}$ \\

\textit{Ours}                & Resnet50                                        &         $78.50$ & $78.04$ & $76.77$ & $77.03$ & $89.75$ \\
\textit{Ours}                & ViT-base                                        &         $81.31$ & $81.16$ & $82.09$ & $81.27$ & $89.89$ \\
\textit{Ours}                & ViT-base-F                                      &         $    73.83 $ & $ 72.76    $ & $ 71.68    $ & $  71.08   $ & $  83.31   $ \\
\textit{Ours}                & ViM-base                                        &         $63.55$ & $63.52$ & $63.32$ & $61.88$ & $81.64$ \\

\textit{Ours}                & Resnet50 + ViT-base                                       &         $81.31$ & $81.16$ & $82.09$ & $81.27$ & $88.91$ \\
\textit{Ours}                & Resnet50 + ViM-base                                        &        $64.49$ & $64.36$ & $64.46$ & $63.66$ & $79.03$ \\
\textit{Ours}                & ViT-base + ViM-base                                        &         $77.57$ & $77.75$ & $77.79$ & $77.02$ & $89.72$ \\
\textit{Ours}                & ViT-base + ViT-base-F                                        &         $80.37$ & $79.80$ & $80.41$ & $79.90$ & $88.67$ \\
\bottomrule
\end{tabular*}
\label{tab:comparison-cxr}
\end{table*}

\begin{table*}[t]
\centering
\caption{Comparison with existing works on KDEF dataset.}
\begin{tabular*}{\tblwidth}{@{} LLLLLLL@{} }
\toprule
\textbf{Paper} & \textbf{Model} & \textbf{Accuracy (\%)} & \textbf{Precision (\%)} & \textbf{Recall (\%)} & \textbf{F1 (\%)} & \textbf{AUC (\%)} \\
\midrule

\cite{sari2021simple} & CNN        & $82.38$ & $-$ & $-$ & $-$ & $-$\\
\cite{Melaugh8868630} & Shallow CNN & $89.4$ & $-$ & $-$ & $-$ & $-$\\

\cite{sym18010015} & AlexNet           & $94.0$ & $93.2$ & $92.6$ & $92.5$ & $-$\\
\cite{sym18010015} & VGG16             & $96.2$ & $95.7$ & $95.1$ & $95.1$ & $-$\\
\cite{sym18010015} & DenseNet121       & $98.5$ & $98.2$ & $97.8$ & $97.8$ & $-$\\
\cite{sym18010015} & EfficientNet-B0   & $98.7$ & $98.4$ & $97.9$ & $98.0$ & $-$\\
\cite{sym18010015} & Resnet50          & $98.2$ & $97.8$ & $97.4$ & $97.4$ & $-$\\
\cite{sym18010015} & ViT-B             & $98.9$ & $98.6$ & $98.2$ & $98.3$ & $-$\\
\cite{sym18010015} & ViT-L             & $99.1$ & $98.8$ & $98.4$ & $98.5$ & $-$\\
\cite{sym18010015} & MDFEFT (Proposed in \cite{sym18010015}) & $\mathbf{99.7}$ & $\mathbf{99.6}$ & $\mathbf{99.2}$ & $\mathbf{99.3}$ & $-$\\

\textit{Ours} &  Resnet50   & $96.32$ & $96.25$ & $96.22$ & $96.22$ & $99.83$ \\
\textit{Ours} &  ViT-base   & $97.85$ & $97.76$ & $97.78$ & $97.76$ & $\mathbf{99.90}$ \\
\textit{Ours} &  ViT-base-F & $70.17$ & $70.24$ & $69.55$ & $68.90$ & $94.31$ \\
\textit{Ours} &  ViM-base   & $97.24$ & $97.24$ & $97.15$ & $97.18$ & $99.85$ \\

\cite{sym18010015} & Gimefive & $99.2$ & $98.9$ & $98.5$ & $98.6$ & $-$\\
\cite{sym18010015} & Attention-Enhanced Ensemble & $99.4$ & $99.1$ & $98.7$ & $98.8$ & $-$\\
\textit{Ours} & Resnet50+ViT-base   &$99.08$&$99.09$&$99.08$&$99.08$&$99.81$\\
\textit{Ours} & Resnet50+ViM-base   &$99.29$&$99.30$&$99.29$&$99.28$&$99.66$\\
\textit{Ours} & ViT-base+ViM-base   &$99.18$&$99.19$&$99.18$&$99.18$&$99.99$\\
\textit{Ours} & ViT-base+ViT-base-F &$99.08$&$99.09$&$99.08$&$99.08$&$99.83$\\
\bottomrule
\end{tabular*}
\label{tab:comparison-kdef}
\end{table*}

\subsection{XAI Quantitative Results and Discussion}
\label{sec:quantitative}

\subsubsection{Metrics}
Tables \ref{tab:explainability_evaluation_train}-\ref{tab:explainability_evaluation_test} provide evaluation metric results for each XAI method and its associated architecture, where the specific metrics are described in detail in Section \ref{chp:measures}. Unlike standard performance reporting, which typically focuses on the reserved test set, all reported metrics were computed using the entire dataset (including training, validation, and test sets), but each subset is shown separately. For KDEF-FGD, the gaze-annotated subset (120 images) was by design allocated entirely to the test partition (see Section~\ref{uniss}); consequently, no gaze ground truth is available for the training split and Table~\ref{tab:explainability_evaluation_train} does not include KDEF-FGD rows.

The tables report the mean and standard deviation of each metric, i.e.\ the estimators $\hat\mu$ and $\hat\sigma$ of Eq.~\eqref{eq:aggregation}. Additionally, a correlation matrix using the Spearman coefficient, Eq.~\eqref{eq:spearman}, was computed for the three metrics, revealing a very strong relationship. This confirms that the results are consistent and not ambiguous. We stress, however, that a strong rank agreement is not a contradiction of Proposition~\ref{prop:complementarity}: monotone concordance in aggregate is compatible with the pointwise dissociations characterized there, and it is exactly those dissociations that carry the architectural signal discussed below.

The saliency maps produced by the considered explainability methods are normalized and projected onto the simplex through the pipeline of Eq.~\eqref{eq:pipeline} and subsequently compared with the human gaze ground truth, which is represented as grayscale images with intensity values in the range $[0,255]$ and scaled to $[0,1]$ prior to the evaluation. This normalization step is required to ensure a consistent scale across methods and to allow a meaningful quantitative comparison. Unlike other explainability methods and architectures, the MambaLRP approach produces saliency maps whose values lie in the interval $[-1,1]$. Positive values close to $1$ indicate pixels that positively influence the predicted class, whereas negative values close to $-1$ correspond to pixels that provide counter-evidence or introduce confusion with respect to the selected class. However, in order to compare these saliency maps with the ground truth human gaze maps using the same evaluation criterion adopted for the other methods, a thresholding operation is applied: all negative values are set to $0$, i.e.\ step (ii) of Eq.~\eqref{eq:pipeline}. In this way, only positively contributing regions are retained, ensuring consistency with the comparison protocol; the interpretive consequences of this choice are those stated in Remark~\ref{rem:positivepart}.

It is important to clarify that this analysis does not aim to assess whether a specific architecture and technique combination is better simply because it resembles human behavior. Instead, the goal is to quantitatively measure the degree of difference between model attention and human gaze, and thus to evaluate how closely the model’s visual focus aligns with that of humans. This comparison is conducted while acknowledging that gaze data across the different datasets are obtained using heterogeneous acquisition methodologies.

\subsubsection{Analysis and discussion}
\begin{figure*}[t]
    \centering
    
    \begin{subfigure}{0.23\linewidth}
        \centering
        \includegraphics[width=\linewidth]{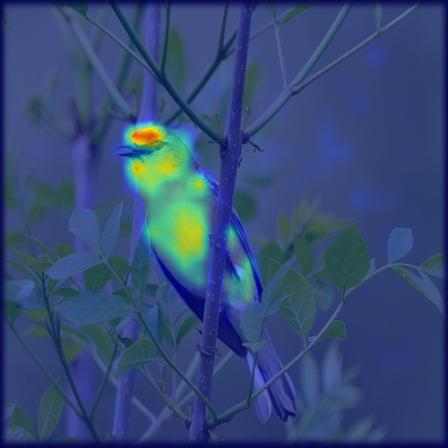}
        \caption{Vit-base-F CUB Image}
    \end{subfigure}
    \hfill
    \begin{subfigure}{0.23\linewidth}
        \centering
        \includegraphics[width=\linewidth]{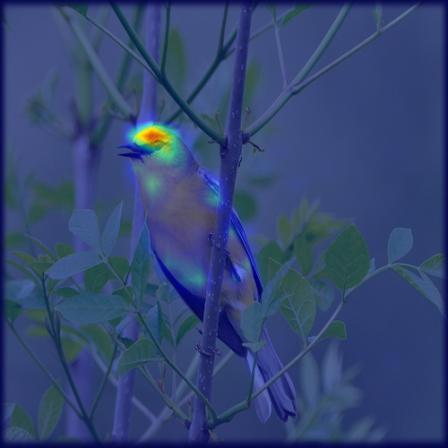}
        \caption{Vit-base CUB Image}
    \end{subfigure}
    \hfill
    \begin{subfigure}{0.23\linewidth}
        \centering
        \includegraphics[width=\linewidth]{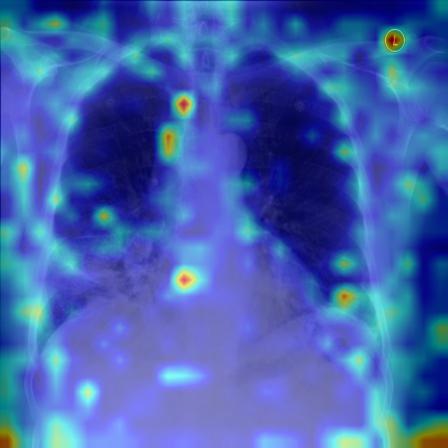}
        \caption{Vit-base-F CXR Image}
    \end{subfigure}
    \hfill
    \begin{subfigure}{0.23\linewidth}
        \centering
        \includegraphics[width=\linewidth]{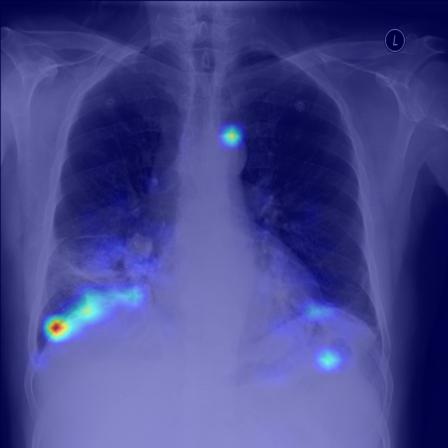}
        \caption{Vit-base CXR Image}
    \end{subfigure}
    
    \caption{Qualitative comparison of the explainability outputs of a Vision Transformer with a frozen backbone during training (a) and (c) versus a non-frozen backbone (b) and (d).}
    \label{fig:differenzaTraTrasformerFrozenENon}
\end{figure*}

Based on the quantitative evaluation reported in Tables~\ref{tab:explainability_evaluation_train}-\ref{tab:explainability_evaluation_test}, several key observations can be derived regarding the stability of explainability methods, the impact of model architectures, and the domain-specific challenges.
\begin{itemize}
    \item \textbf{Generalization and Robustness (Q3):}
    Comparing the results between the train and test sets, we observe a high degree of consistency in terms of mean values and standard deviations across almost all metrics. Quantitatively, the discrepancy $|\mathcal{A}_{\mathcal S}(\cdot;\dset_{\mathrm{tr}})-\mathcal{A}_{\mathcal S}(\cdot;\dset_{\mathrm{te}})|$ never exceeds $0.027$ on CUB-200-2011 and $0.069$ on CXR for any convolutional or Transformer configuration and any $\mathcal{S}$, i.e.\ at most one fifth of the corresponding within-split dispersion $\hat\sigma$, so that the two estimates are statistically indistinguishable at the resolution afforded by the sample size. This suggests that the generated saliency maps are robust and do not suffer from overfitting, and the explainability methods capture intrinsic feature attributions that generalize well to unseen data. This behavior is expected, since a well-generalized model learns stable and meaningful patterns from the data, which are consistently reflected across both subsets.

    One configuration departs from this pattern and we report it explicitly: ViM-base with MambaLRP on CXR shifts by $0.083$, $0.140$ and $0.126$ in $\JSS$, $\CSS$ and $\PCC$ respectively between the two splits, the first two gaps exceeding the corresponding $\hat\sigma$ on either split. Since the remaining eleven model--method pairs are stable and the deviation is confined to a single attribution operator, the effect is more plausibly ascribed to the numerical conditioning of the relevance-preserving rule of Eq.~\eqref{eq:mambalrp_gate} on the selective recurrence than to a generic overfitting of the classifier. We return to this point in Section~\ref{sec:conclusion}.

    \item \textbf{Impact of Backbone Freezing (ViT-base-F) (Q4):}
     As depicted in Figure~\ref{fig:differenzaTraTrasformerFrozenENon}, the ViT models with and without frozen internal parameters exhibit very different learning behaviors and focus on different regions of the input. From Tables~\ref{tab:comparison-cub}-\ref{tab:comparison-kdef}, it can be observed that during fine-tuning, the Transformer with a non-frozen backbone achieves better performance on the test set. Allowing the attention layers to learn and adapt to the new task leads to improvements both in classification performance and in the quality of the explainability results. Nevertheless, a notable improvement according to explainability metrics is observed with the ViT-base-F variant compared to the standard ViT-base, particularly on the medical dataset (CXR). For instance, on the CXR test set, ViT-base-F achieves a Jensen–Shannon (JS) score of $0.6528$ compared to $0.4822$ for ViT-base. Since the two variants differ only in the feasible set of Eq.~\eqref{eq:erm}, namely $\Theta_{\mathrm{F}}\subset\Theta$, the comparison isolates backbone freezing as a controlled intervention in the sense of (Q4), and the $+0.17$ gain in $\JSS$ is obtained together with a loss in classification accuracy ($73.83\%$ against $81.31\%$). Alignment with human attention and predictive accuracy are thus not monotonically related, which is itself an argument for reporting the two separately. This indicates that preserving pre-trained feature representations via backbone freezing could yield more stable and semantically meaningful explanations in medical imaging contexts.

    \item \textbf{Architecture- and Domain-Specific Behavior (Q1--Q2):}
    \begin{itemize}
        \item \textbf{CNNs (ResNet50):} Demonstrate the most versatile performance, maintaining solid scores across both natural (CUB/KDEF-FGD) and medical (CXR) domains. However, on the medical dataset, gradient-based methods like GradCAM exhibit high variance ($\sigma \approx 0.36$ for Chi-square), whereas activation-based methods like ScoreCAM offer superior stability ($\sigma \approx 0.07$).
        \item \textbf{SSMs (ViM-base):} Exhibit a polarized behavior. While MambaLRP struggles on the natural image datasets, it achieves state-of-the-art localization scores on CXR (JS $\approx 0.82$, Chi-square $\approx 0.87$).
    \end{itemize}

    \item \textbf{Metric Divergence in Medical Imaging:}
    On the CXR dataset, we observe a significant discrepancy between distribution-based metrics (JS, Chi-square) and correlation metrics (PCC). Specifically, MambaLRP achieves the highest JS scores but near-zero or negative PCC. This implies that while the model correctly identifies the \textit{region} of interest (high overlap), the intensity distribution within the saliency map does not linearly correlate with human annotations. This is precisely the regime isolated by Proposition~\ref{prop:complementarity}(c): with $\JSS=0.8214$ and $\CSS=0.8742$ the two distributions share their support almost entirely, whereas $\PCC=-0.0108$ certifies that, once the common mean $|\Grid|^{-1}$ is removed, the residual spatial profiles are essentially uncorrelated. The observation is therefore not an inconsistency between metrics but the expected signature of a saliency map that is correct in \emph{where} it looks and uninformative in \emph{how strongly} it looks there. Reporting $\JSS$ alone, as is common practice, would have concealed the distinction. Conversely, ResNet50 with ScoreCAM offers the best trade-off, providing both accurate localization and reasonable correlation with human perception ($PCC \approx 0.35$).
\end{itemize}

Furthermore, we performed a pairwise comparison of the generated heatmaps to quantitatively evaluate how explanations differ across architectures on the test sets. Table \ref{tab:explainability_model_comparison} contains all the results obtained using GradCAM as an explainability technique for CNNs. Our analysis reveals three key insights:

\begin{itemize}
    \item \textbf{Domain Sensitivity:} Explainability agreement is highly dependent on the visual domain. Models tend to converge on similar features for everyday objects (CUB) and consistently localized facial markers (KDEF-FGD), demonstrating that emotion recognition aligns closely with natural image processing. However, this agreement drops significantly when shifting to highly specialized medical diagnostics (CXR), where models diverge on critical diagnostic features.
    
    \item \textbf{Architectural Alignment and Fine-Tuning Sensitivity:} In natural domains, CNNs and standard Transformers focus on similar regions, with ResNet50 and ViT-base exhibiting high consistency on CUB ($PCC \approx 0.74$). On human faces (KDEF-FGD), ResNet50 maintains this strong agreement, particularly with ScoreCAM ($PCC \approx 0.71$). Conversely, fine-tuning Transformers on specific human-centric tasks (ViT-base-F) fragments their attention maps, causing a notable drop in correlation on KDEF-FGD ($PCC \approx 0.28$) and shifting focus away from the traditional morphological features prioritized by CNNs.
    
    \item \textbf{Divergence of ViM:} The ViM-base architecture produces fundamentally different saliency maps compared to ResNet and ViT. It shows low or negligible correlation with other models across all datasets, indicating a unique spatial attention mechanism. In particular, ViM-base yields completely divergent saliency maps on facial datasets, dropping to a negative correlation ($PCC \approx -0.11$). This reinforces the hypothesis that Mamba's state-space spatial attention operates on a fundamentally different visual logic compared to both CNNs and standard self-attention mechanisms, regardless of the visual domain. It is worth noting that this divergence is measured against the other \emph{models}, not against the human reference, and that the two are logically independent: on KDEF-FGD, ResNet50 and ViT-base agree with each other ($\PCC\approx0.70$) and both align with $\gaze$, whereas on CXR they disagree with each other ($\PCC\approx0.22$) while both remain weakly aligned with $\gaze$. Model-to-model agreement is therefore not a proxy for human alignment, and the two quantities have to be estimated separately, as done in Table~\ref{tab:explainability_model_comparison}.
    
\end{itemize}

We acknowledge that our findings are based on three datasets, a scope directly constrained by the significant computational overhead of benchmarking multiple complex architectures and XAI pipelines simultaneously. To maximize robustness within this constraint, we intentionally selected three visual domains with highly divergent structural and semantic characteristics. While bounded in volume, these contrasting data regimes provide diverse testing environments that make the observed attention dynamics and architectural insights highly generalizable and, in our opinion, of strong interest to the XAI community.

\FloatBarrier

\begin{table*}[Ht]
\centering
\caption{Explainability evaluation on train sets. KDEF-FGD is absent because its gaze annotations (120 images) were allocated entirely to the test set by construction; no gaze ground truth is available for the training split.}
\label{tab:explainability_evaluation_train}
\begin{tabular*}{\tblwidth}{@{} LLLLLL@{} }        
\toprule
\textbf{Model} & \textbf{Dataset} &  \textbf{Method} & \textbf{Jensen–Shannon} & \textbf{Chi-square} & \textbf{PCC} \\
\midrule
ResNet50   & CUB-200-2011 & GradCAM     & $\mathbf{0.6067 \pm 0.0810}$ & $\mathbf{0.4617 \pm 0.2050}$ &         $0.7386 \pm 0.1390$  \\
ResNet50   & CUB-200-2011 & ScoreCAM    &         $0.5620 \pm 0.0900$  &         $0.3413 \pm 0.2471$  &         $0.6822 \pm 0.1450$  \\
ResNet50   & CUB-200-2011 & AblationCAM &         $0.6005 \pm 0.0790$  &         $0.4477 \pm 0.2050$  & $\mathbf{0.7413 \pm 0.1360}$ \\
ViT-base   & CUB-200-2011 & ViTLRP      &         $0.5342 \pm 0.0670$  &         $0.2724 \pm 0.2000$  &         $0.5959 \pm 0.1570$  \\
ViT-base-F & CUB-200-2011 & ViTLRP      &         $0.5224 \pm 0.0650$  &         $0.2386 \pm 0.1960$  &         $0.5625 \pm 0.1560$  \\
ViM-base   & CUB-200-2011 & MambaLRP    &         $0.3423 \pm 0.0430$  &         $-0.3579 \pm 0.1583$  &         $0.1350 \pm 0.2641$  \\
\midrule
ResNet50   & CXR & GradCAM     &           $0.5695 \pm 0.1310$  &          $0.3340 \pm 0.3445$  &          $0.1976 \pm 0.2640$  \\ 
ResNet50   & CXR & ScoreCAM    & $ \mathbf{0.7432 \pm 0.0378}$  & $\mathbf{0.7473 \pm 0.0720}$  & $ \mathbf{0.3704 \pm 0.1887}$  \\
ResNet50   & CXR & AblationCAM &         $0.6499 \pm 0.1114$  &         $0.5419 \pm 0.2471$  &         $0.2516 \pm 0.2792$  \\   
ViT-base   & CXR & ViTLRP      & $0.4821 \pm 0.0580$ & $0.0842 \pm 0.1830$ & $0.1615 \pm 0.1440$ \\
ViT-base-F & CXR & ViTLRP      & $0.6579 \pm 0.0370$ & $0.5702 \pm 0.0873$ & $-0.0166 \pm 0.1272$ \\
ViM-base   & CXR & MambaLRP    & $0.7384 \pm 0.0405$ & $0.7339 \pm 0.0768$ & $-0.1371 \pm 0.1561$ \\
\bottomrule
\end{tabular*}
\end{table*}

\begin{table*}[Ht]
\centering
\caption{Explainability evaluation on test sets.}
\label{tab:explainability_evaluation_test}
\begin{tabular*}{\tblwidth}{@{} LLLLLL@{} }        
\toprule
\textbf{Model} & \textbf{Dataset} &  \textbf{Method} & \textbf{Jensen–Shannon} & \textbf{Chi-square} & \textbf{PCC} \\
\midrule
ResNet50   & CUB-200-2011 & GradCAM     & $\mathbf{0.6123 \pm 0.0840}$ & $\mathbf{0.4732 \pm 0.2150}$ &         $0.7362 \pm 0.1470$  \\
ResNet50   & CUB-200-2011 & ScoreCAM    &         $0.5712 \pm 0.0870$  &         $0.3675 \pm 0.2380$  &         $0.6935 \pm 0.1460$  \\
ResNet50   & CUB-200-2011 & AblationCAM &         $0.6005 \pm 0.0800$  &         $0.4578 \pm 0.2090$  & $\mathbf{0.7402 \pm 0.1400}$ \\
ViT-base   & CUB-200-2011 & ViTLRP      &         $0.5391 \pm 0.0670$  &         $0.2867 \pm 0.1960$  &         $0.6047 \pm 0.1530$  \\
ViT-base-F & CUB-200-2011 & ViTLRP      &         $0.5239 \pm 0.0670$  &         $0.2434 \pm 0.1910$  &         $0.5669 \pm 0.1540$  \\
ViM-base   & CUB-200-2011 & MambaLRP    &         $0.3410 \pm 0.0430$  &         $-0.3626 \pm 0.1779$  &         $0.1315 \pm 0.2719$  \\
\midrule
ResNet50   & CXR & GradCAM     &         $0.5450 \pm 0.1340$ &          $0.2651 \pm 0.3610$  & $0.1524 \pm 0.2809$ \\     
ResNet50   & CXR & ScoreCAM    &         $0.7365 \pm 0.0385$ &          $0.7344 \pm 0.0751$  & $\mathbf{0.3522 \pm  0.1929}$ \\         
ResNet50   & CXR & AblationCAM &         $0.6383 \pm 0.1110$ &          $0.5141 \pm 0.2430$  & $0.2143 \pm 0.3070$ \\
ViT-base   & CXR & ViTLRP      &         $0.4822 \pm 0.0560$ &          $0.0846 \pm 0.1770$  & $0.1615 \pm 0.1350$ \\
ViT-base-F & CXR & ViTLRP      &         $0.6528 \pm 0.0367$ &          $0.5583 \pm 0.0878$  & $-0.0433 \pm 0.1174$ \\
ViM-base   & CXR & MambaLRP    & $\mathbf{0.8214 \pm 0.0171}$ & $\mathbf{0.8742 \pm 0.0234}$ & $-0.0108 \pm 0.0791$ \\
\midrule
ResNet50   & KDEF-FGD & GradCAM     & $0.6245\pm0.0525$ &  $0.5090\pm0.1338$ &  $0.6799\pm0.1118$ \\     
ResNet50   & KDEF-FGD & ScoreCAM    & $\mathbf{0.6345\pm0.0386}$ &  $\mathbf{0.5373\pm0.1338}$ & $\mathbf{0.7129\pm0.0714}$ \\         
ResNet50   & KDEF-FGD & AblationCAM & $0.6233\pm0.0452$ &  $0.5074\pm0.1133$ &  $0.6797\pm0.0984$ \\
ViT-base   & KDEF-FGD & ViTLRP      & $0.5402\pm0.0377$ &  $0.2808\pm0.1128$ &  $0.4639\pm0.1064$ \\
ViT-base-F & KDEF-FGD & ViTLRP      & $0.4388\pm0.0475$ & $-0.0193\pm0.1567$ &  $0.2788\pm0.1525$ \\
ViM-base   & KDEF-FGD & MambaLRP    & $0.3644\pm0.0232$ & $-0.2750\pm0.0850$ & $-0.1076\pm0.1930$ \\
\bottomrule
\end{tabular*}
\end{table*}

\begin{table*}[Ht]
\centering
\caption{Explainability Comparison between Models on test sets.}
\label{tab:explainability_model_comparison}
\begin{tabular*}{\tblwidth}{@{} LLLLLLL@{} }        
\toprule
\textbf{Model} & \textbf{Model} & \textbf{Dataset} & \textbf{Jensen–Shannon} & \textbf{Chi-square} & \textbf{PCC} \\
\midrule
ResNet50 & ViT-base & CUB-200-2011 & $0.6411 \pm 0.0528$ & $0.5544 \pm 0.1270$ &  $0.7362 \pm 0.0986$ \\
ResNet50 & ViM-base & CUB-200-2011 & $0.4703 \pm 0.0655$ & $0.0492 \pm 0.2018$ &  $0.1558 \pm 0.2659$ \\
ViT-base & ViM-base & CUB-200-2011 & $0.4447 \pm 0.0801$ & $-0.0344 \pm 0.2498$ &  $0.1625 \pm 0.2785$ \\
\midrule
ResNet50 & ViT-base & CXR  & $0.3868 \pm 0.0716$ & $-0.1782 \pm 0.2446$ & $0.2192 \pm 0.1709$ \\
ResNet50 & ViM-base & CXR  & $0.5404 \pm 0.1153$ &  $0.2652 \pm 0.3083$ & $ -0.0138 \pm 0.083$ \\
ViT-base & ViM-base & CXR  & $0.3941 \pm 0.0688$ & $-0.1691 \pm 0.2286$ & $-0.0035 \pm 0.1091$ \\
\midrule
ResNet50 & ViT-base & KDEF-FGD & $0.6731 \pm 0.0449$ & $ 0.6211 \pm 0.1052$ & $ 0.7015 \pm 0.0941$ \\
ResNet50 & ViM-base & KDEF-FGD & $0.3957 \pm 0.0164$ & $-0.1635 \pm 0.0587$ & $-0.0420 \pm 0.2002$ \\
ViT-base & ViM-base & KDEF-FGD & $0.4416 \pm 0.0501$ & $-0.0217 \pm 0.1563$ & $-0.0308 \pm 0.1858$ \\
\bottomrule
\end{tabular*}
\end{table*}

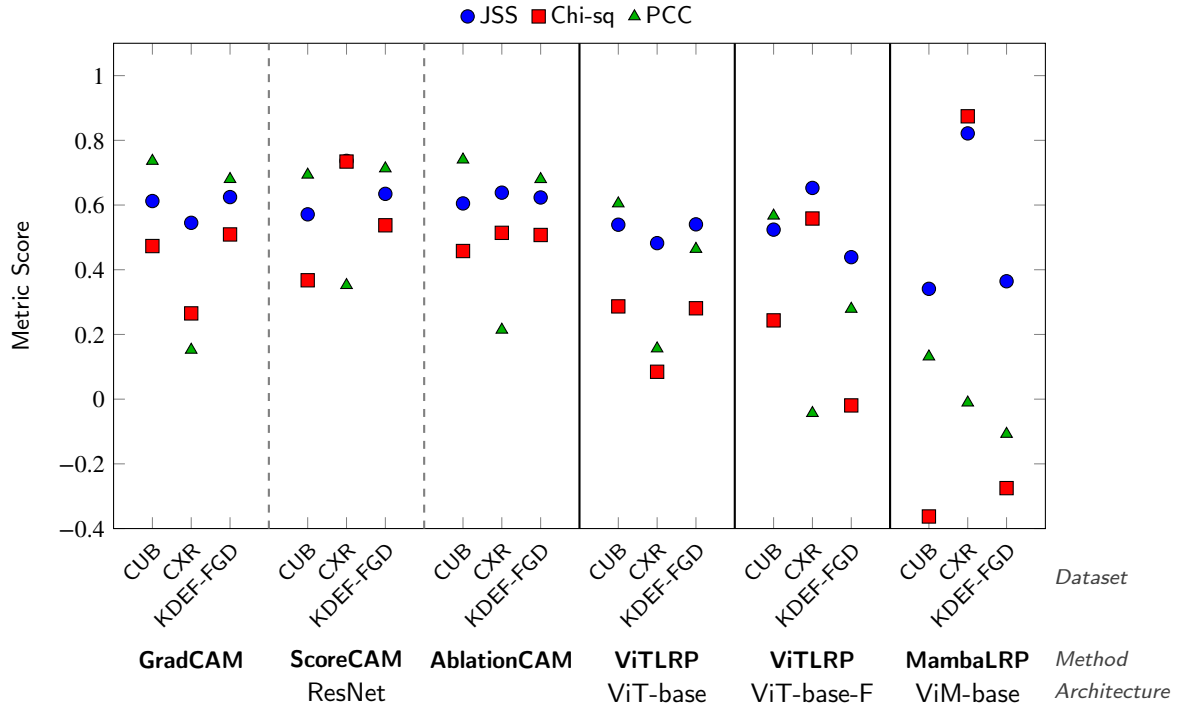
\begin{figure*}[t]
\centering
\begin{tikzpicture}
    \begin{axis}[
        width=0.8\textwidth, 
        height=8cm,
        ymin=-0.4, ymax=1.1,
        ytick={-0.4, -0.2, 0, 0.2, 0.4, 0.6, 0.8, 1.0},
        ylabel={Metric Score},
        xmin=0, xmax=24, 
        xtick={1,2,3, 5,6,7, 9,10,11, 13,14,15, 17,18,19, 21,22,23},
        xticklabels={
            CUB, CXR, KDEF-FGD, 
            CUB, CXR, KDEF-FGD, 
            CUB, CXR, KDEF-FGD, 
            CUB, CXR, KDEF-FGD, 
            CUB, CXR, KDEF-FGD, 
            CUB, CXR, KDEF-FGD
        },
        xticklabel style={font=\footnotesize, rotate=45, anchor=north east},
        legend style={at={(0.5, 1.1)}, anchor=north, legend columns=-1, draw=none, fill=none, font=\small},
        only marks,
        mark size=2.5pt,
        clip=false 
        ]
        
        \addplot[mark=*, fill=blue, draw=black] coordinates {
            ( 1, 0.612339)  ( 2, 0.545037)  ( 3, 0.624579)      
            ( 5, 0.571234)  ( 6, 0.736467)  ( 7, 0.634572)      
            ( 9, 0.604966)  (10, 0.638251)  (11, 0.623351)      
            (13, 0.539171)  (14, 0.482235)  (15, 0.540298)      
            (17, 0.523938)  (18, 0.652867)  (19, 0.438862)      
            (21, 0.341008)  (22, 0.821414)  (23, 0.364407)      
        };
        \addlegendentry{JSS \hspace{10pt}} 
        
        \addplot[mark=square*, fill=red, draw=black] coordinates {
            ( 1, 0.473211)  ( 2, 0.265068)  ( 3,  0.509099)     
            ( 5, 0.367452)  ( 6, 0.734409)  ( 7,  0.537399)     
            ( 9, 0.457849)  (10, 0.514056)  (11,  0.507424)     
            (13, 0.286727)  (14, 0.084622)  (15,  0.280884)     
            (17, 0.243472)  (18, 0.558339)  (19, -0.019371)     
            (21,-0.362662)  (22, 0.874262)  (23, -0.275004)     
        };
        \addlegendentry{Chi-sq \hspace{10pt}}
        
        \addplot[mark=triangle*, fill=green!70!black, draw=black] coordinates {
            ( 1, 0.736168)  ( 2, 0.152263)  ( 3,  0.679944)     
            ( 5, 0.693527)  ( 6, 0.352235)  ( 7,  0.712952)     
            ( 9, 0.740222)  (10, 0.214263)  (11,  0.679723)     
            (13, 0.604711)  (14, 0.156432)  (15,  0.463985)     
            (17, 0.566925)  (18,-0.043306)  (19,  0.278823)     
            (21, 0.131520)  (22,-0.010853)  (23, -0.107684)     
        };
        \addlegendentry{PCC}
        
        \draw[dashed, gray, thick] (axis cs:4, -0.4) -- (axis cs:4, 1.1);
        \draw[dashed, gray, thick] (axis cs:8, -0.4) -- (axis cs:8, 1.1);
        
        \draw[solid, black, thick] (axis cs:12, -0.4) -- (axis cs:12, 1.1);
        \draw[solid, black, thick] (axis cs:16, -0.4) -- (axis cs:16, 1.1);
        \draw[solid, black, thick] (axis cs:20, -0.4) -- (axis cs:20, 1.1);
        
        \node[anchor=north, font=\bfseries\color{black}\small] at (axis cs:  2, -0.75) {GradCAM};
        \node[anchor=north, font=\bfseries\color{black}\small] at (axis cs:  6, -0.75) {ScoreCAM};
        \node[anchor=north, font=\bfseries\color{black}\small] at (axis cs: 10, -0.75) {AblationCAM};
        \node[anchor=north, font=\bfseries\color{black}\small] at (axis cs: 14, -0.75) {ViTLRP};
        \node[anchor=north, font=\bfseries\color{black}\small] at (axis cs: 18, -0.75) {ViTLRP};
        \node[anchor=north, font=\bfseries\color{black}\small] at (axis cs: 22, -0.75) {MambaLRP};
        
        \node[anchor=north, font=\normalsize] at (axis cs:  6, -0.85) {ResNet};
        \node[anchor=north, font=\normalsize] at (axis cs: 14, -0.85) {ViT-base};
        \node[anchor=north, font=\normalsize] at (axis cs: 18, -0.85) {ViT-base-F};
        \node[anchor=north, font=\normalsize] at (axis cs: 22, -0.85) {ViM-base};
        
        \node[anchor=north west, font=\itshape\color{darkgray}\footnotesize] at (axis cs: 24, -0.5) {Dataset};
        \node[anchor=north west, font=\itshape\color{darkgray}\footnotesize] at (axis cs: 24, -0.75) {Method};
        \node[anchor=north west, font=\itshape\color{darkgray}\footnotesize] at (axis cs: 24, -0.85) {Architecture};
        
    \end{axis}
\end{tikzpicture}
\caption{Quantitative comparison of XAI metrics. Visual representation of the fidelity scores across different attribution methods and architectures. Raw values in Table~\ref{tab:explainability_evaluation_test}.}
\label{fig:metrics_plot}
\end{figure*}

\subsection{Qualitative Analysis}
\label{sec:qualitative}

Table \ref{tab:explainability_comparison} shows the explainability results for selected samples across the three datasets. 
As already discussed in Section~\ref{sec:related}, the Grad-CAM method is not suitable for the analysis of attention-based models such as Transformers, and therefore alternative techniques based on Layer-wise Relevance Propagation (LRP) have been adopted. The qualitative results clearly highlight substantial differences between these approaches. Grad-CAM typically operates on the last convolutional layer of a network. Since the feature maps at this stage have a much lower spatial resolution than the original input image, the resulting explanations are inherently coarse and require up-sampling, leading to blurred and low-resolution saliency maps. In contrast, LRP-based methods propagate relevance backward through all layers of the model down to the input space. This process enables the attribution of relevance directly at the pixel-level, resulting in high-resolution and more precise explanations. This difference explains why the visualizations obtained from Transformer-based models and Mamba architectures tend to focus on well-defined and localized regions of the image, as opposed to the more diffuse and coarse regions typically highlighted by Grad-CAM. For instance, on the CUB dataset, CNNs generally tend to focus on the entire bird body or on the head region. In contrast, Transformer-based architectures, when identifying specific bird classes, mainly rely on distinctive features such as the beak or particular wing color patterns, as shown in Table~\ref{tab:explainability_comparison}.
This granular focus is intrinsically linked to the ViT architecture, which processes the input image as a sequence of fixed-size patches (specifically $16 \times 16$ pixels). Unlike CNNs, where pooling layers progressively reduce spatial resolution, effectively aggregating features into coarser regions, the token-based representation of ViTs allows the self-attention mechanism to assign high relevance scores to individual patches. This capability enables the model to isolate and attend to minute details within specific $16 \times 16$ regions without the blurring effect typical of upsampled convolutional feature maps. The effect is quantifiable rather than merely visual: the intrinsic support of a ViTLRP explanation is the $28\times28$ token grid induced by $P=16$ at the input resolution used here, against the $14\times14$ feature-map lattice of Eq.~\eqref{eq:gradcam} for the CAM family. The two operators therefore differ by a factor of four in the number of independently addressable cells, which upper-bounds the spatial detail an explanation can express irrespective of the faithfulness of the attribution rule.
However, this behavior is highly image-dependent.

In the CXR dataset, we observe a distinct behavioral divergence between architectures regarding the scope of attention. CNN models tend to focus on broader anatomical regions, as evidenced in the \textit{Congestive Heart Failure} (CHF) sample, where the heatmap covers the entire cardiac and pulmonary area. Conversely, Transformers and Mamba architectures display an extreme, highly localized focus. While this pixel-level specificity could be theoretically precise, it may prove less intuitive for radiologists diagnosing diffuse pathologies (e.g., lobar pneumonia or edema), where the disease manifests as a regional texture change rather than a focal point. Thus, the CNN's approach of highlighting the ``general area'', despite lacking edge precision, often appears more natural and aligned with the holistic review process typical of clinical practice.

Finally, the analysis of the KDEF-FGD dataset reveals further significant discrepancies between machine attention and human visual processing. As observed in Table~\ref{tab:explainability_comparison}, the \textit{Human Gaze} reference concentrates more on the central region of the face. Across different emotional states, such as happy, sad and neutral, human subjects predominantly anchor their visual focus on the nose and the periocular area. This central fixation likely serves as an optimal pivot point for humans to rapidly process holistic facial cues. In contrast, the CNN-based saliency methods emphasize the lower facial regions, exhibiting strong coarse activations around the mouth, cheeks and cheekbones. This suggests that these convolutional models rely disproportionately on variations in mouth morphology to classify facial expressions, rather than the eye region preferred by humans. Meanwhile, the attention-based models exhibit behaviors similar to those observed in the previous datasets. ViTLRP continues to demonstrate its characteristic patch-level granularity, showing attention across multiple discrete facial landmarks simultaneously. MambaLRP, on the other hand, produces highly sparse point-like activations, suggesting that the model is focusing on isolated minute details and thus failing to capture the contiguous regional focus demonstrated both by human gaze and CNN models, further highlighting the gap between human visual strategies and current XAI outputs.

\section{Conclusion}
\label{sec:conclusion}
In this work, we presented a comprehensive comparative analysis of explainability across three distinct deep learning paradigms: CNNs, ViTs, and ViM, evaluating their alignment with human visual attention. By leveraging ground-truth gaze data from fine-grained natural image classification (CUB), medical diagnosis (CXR), and facial emotion classification (KDEF-FGD) tasks, we quantitatively evaluated model attention beyond qualitative visual inspection. Methodologically, we formulated the comparison as the estimation of the alignment functional of Definition~\ref{def:alignment} under an explicitly specified normalization pipeline (Eq.~\eqref{eq:pipeline}), and we characterized the invariances of the three similarity estimators (Proposition~\ref{prop:complementarity}). This apparatus is what turns several of the observations below from qualitative impressions into statements with a determinate meaning. Our results show that architectural inductive biases strongly affect explainability: CNNs exhibit more spatially distributed attention patterns, whereas ViTs display a more localized and fine-grained focus, a difference whose upper bound is fixed by the $14\times14$ and $28\times28$ intrinsic supports of the respective attribution operators. In addition, backbone freezing improves interpretability in medical imaging scenarios by preserving pretrained feature representations that yield more semantically meaningful explanations, although our results show that this gain in alignment is not accompanied by a gain in accuracy, so that the two objectives have to be traded off explicitly rather than assumed to coincide.

\begin{table*}[pos=t]
\caption{Qualitative Comparison of Explainability Methods.}
\label{tab:explainability_comparison}
\renewcommand{\arraystretch}{1.2}
\setlength{\tabcolsep}{3pt} 
\begin{tabular*}{\tblwidth}{@{\extracolsep{\fill}} >{\centering\arraybackslash}m{2.8cm} *{6}{>{\centering\arraybackslash}m{1.9cm}} @{}}
\toprule
\textbf{Description} 
& \textbf{Human Gaze} 
& \textbf{GradCAM} 
& \textbf{ScoreCAM} 
& \textbf{AblationCAM} 
& \textbf{ViTLRP} 
& \textbf{MambaLRP} \\
\midrule

CUB - Red Wing BlackBird
& \includegraphics[width=\linewidth]{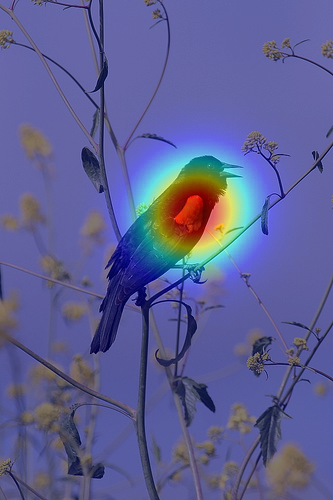}
& \includegraphics[width=\linewidth]{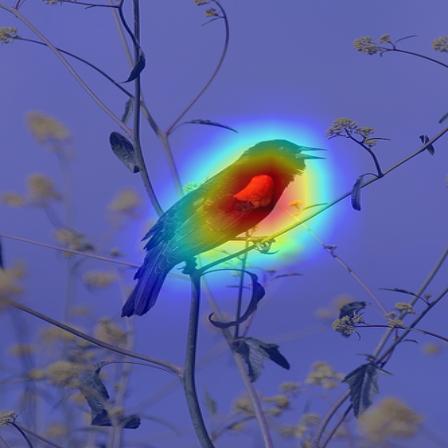}
& \includegraphics[width=\linewidth]{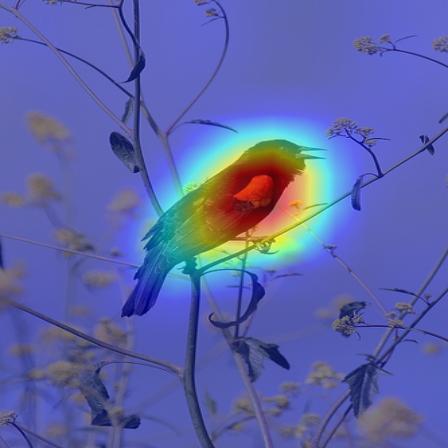}
& \includegraphics[width=\linewidth]{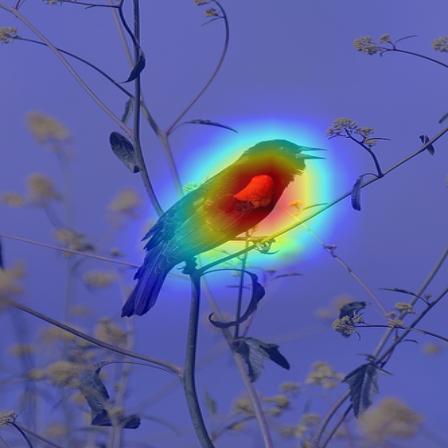}
& \includegraphics[width=\linewidth]{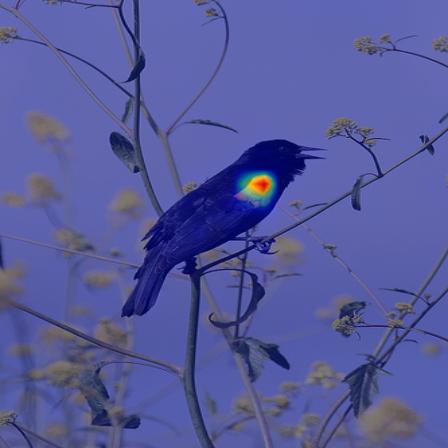}
& \includegraphics[width=\linewidth]{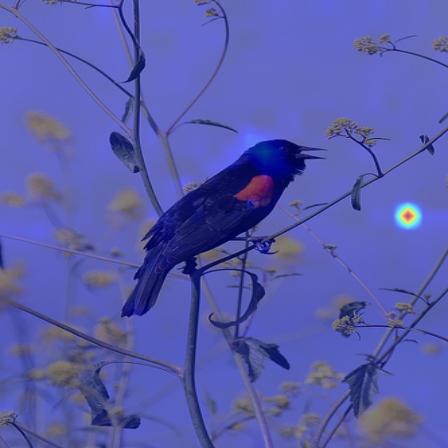} \\

CUB - Least Auklet
& \includegraphics[width=\linewidth]{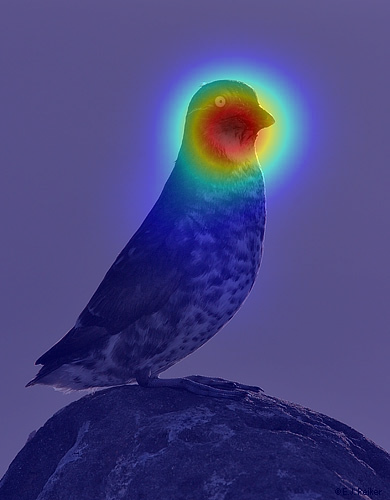}
& \includegraphics[width=\linewidth]{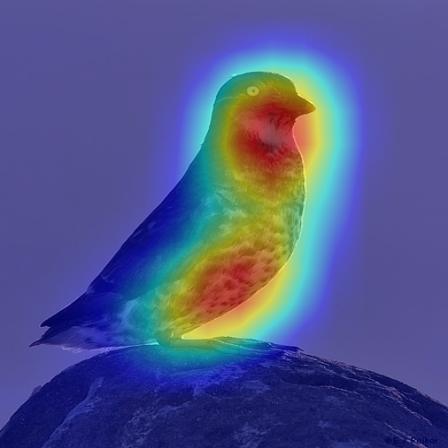}
& \includegraphics[width=\linewidth]{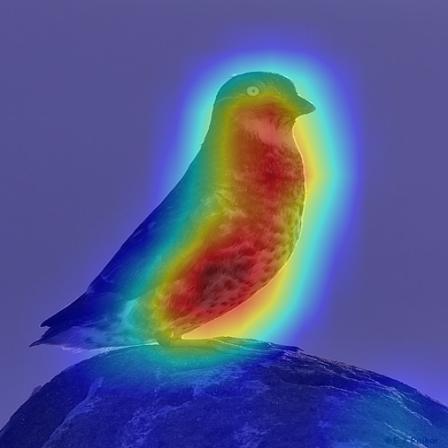}
& \includegraphics[width=\linewidth]{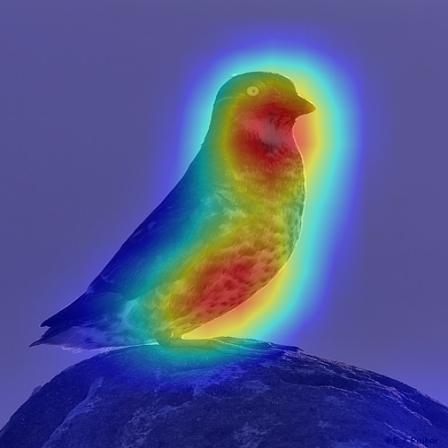}
& \includegraphics[width=\linewidth]{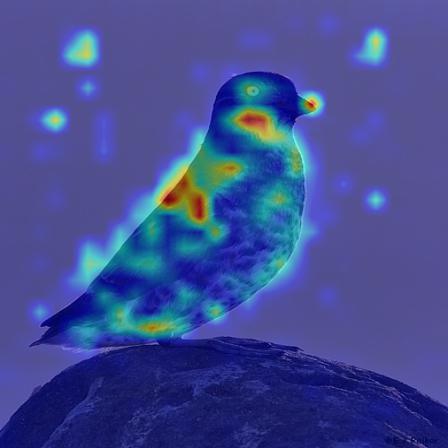}
& \includegraphics[width=\linewidth]{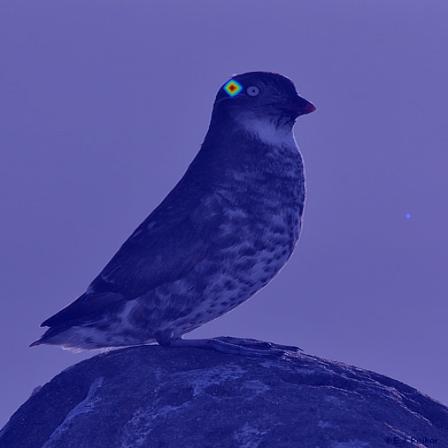} \\

CUB - Prothonotary Warbler
& \includegraphics[width=\linewidth]{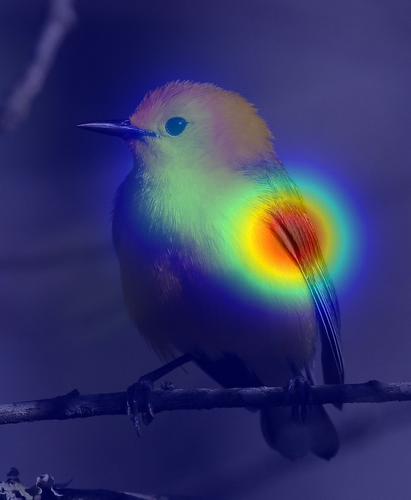}
& \includegraphics[width=\linewidth]{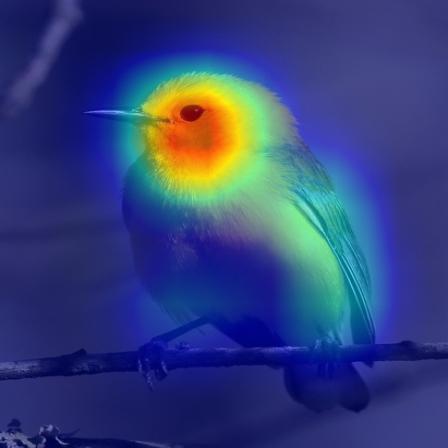}
& \includegraphics[width=\linewidth]{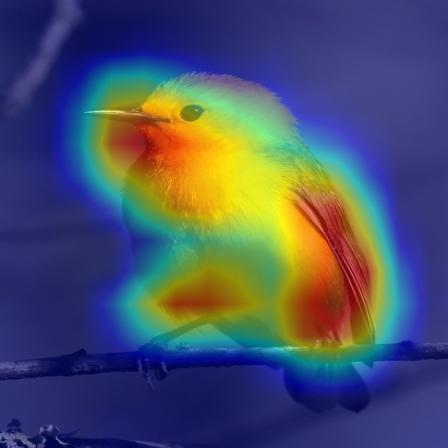}
& \includegraphics[width=\linewidth]{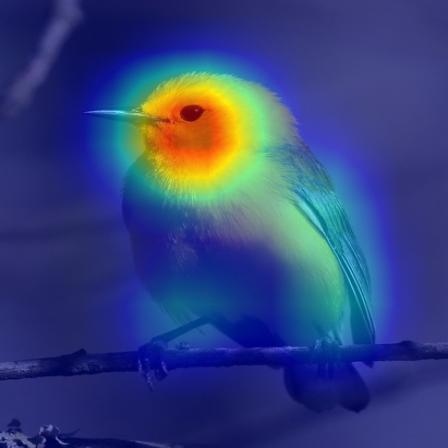}
& \includegraphics[width=\linewidth]{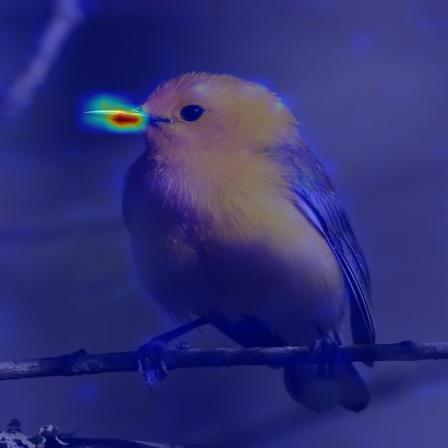}
& \includegraphics[width=\linewidth]{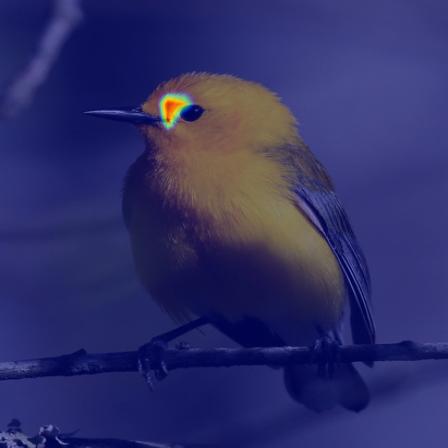} \\

CXR - Pneumonia
& \includegraphics[width=\linewidth]{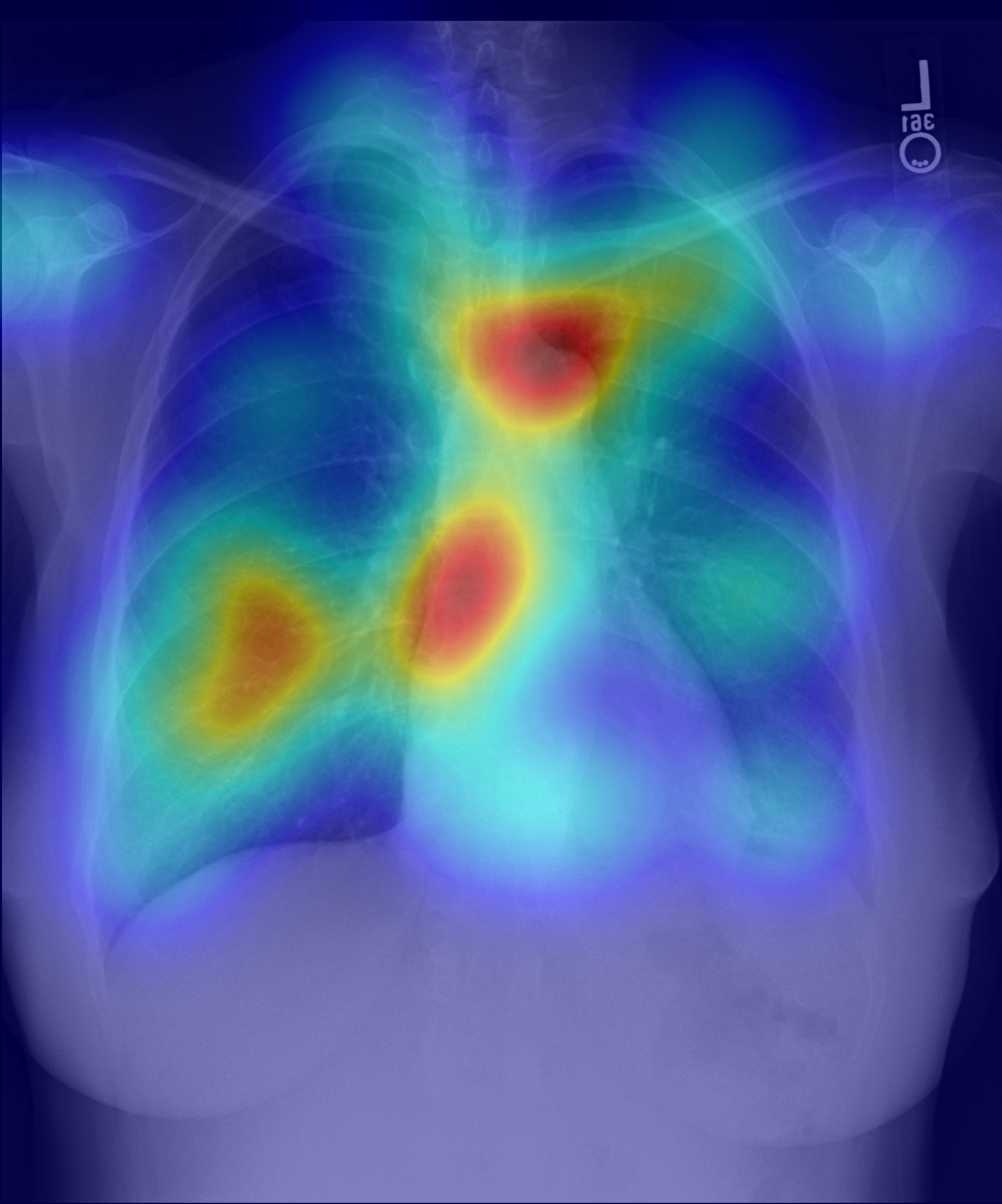}
& \includegraphics[width=\linewidth]{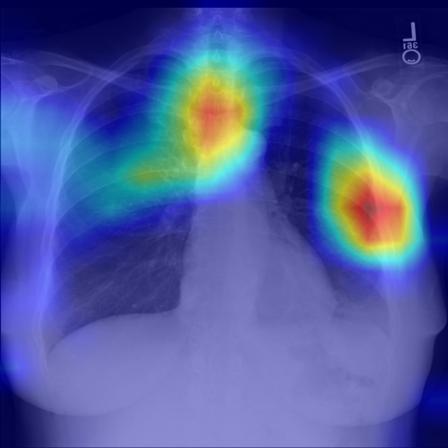}
& \includegraphics[width=\linewidth]{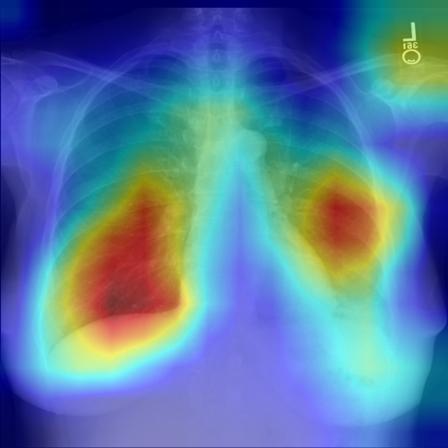}
& \includegraphics[width=\linewidth]{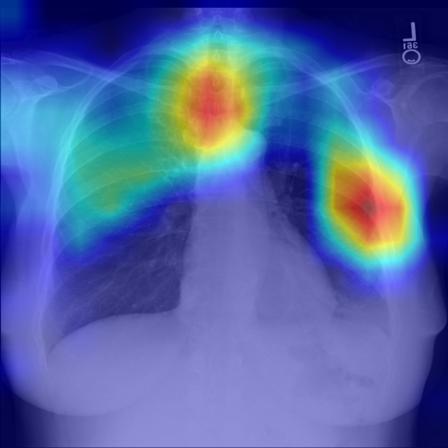}
& \includegraphics[width=\linewidth]{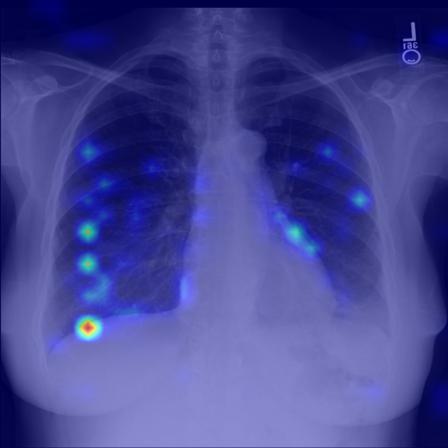}
& \includegraphics[width=\linewidth]{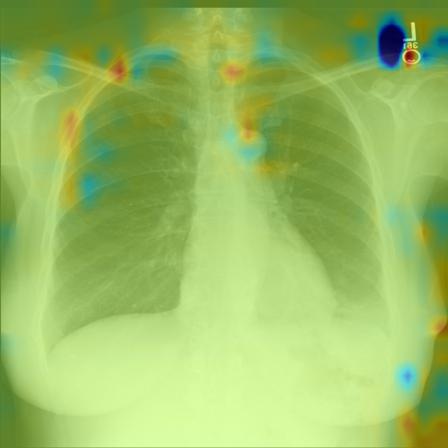} \\

CXR - CHF
& \includegraphics[width=\linewidth]{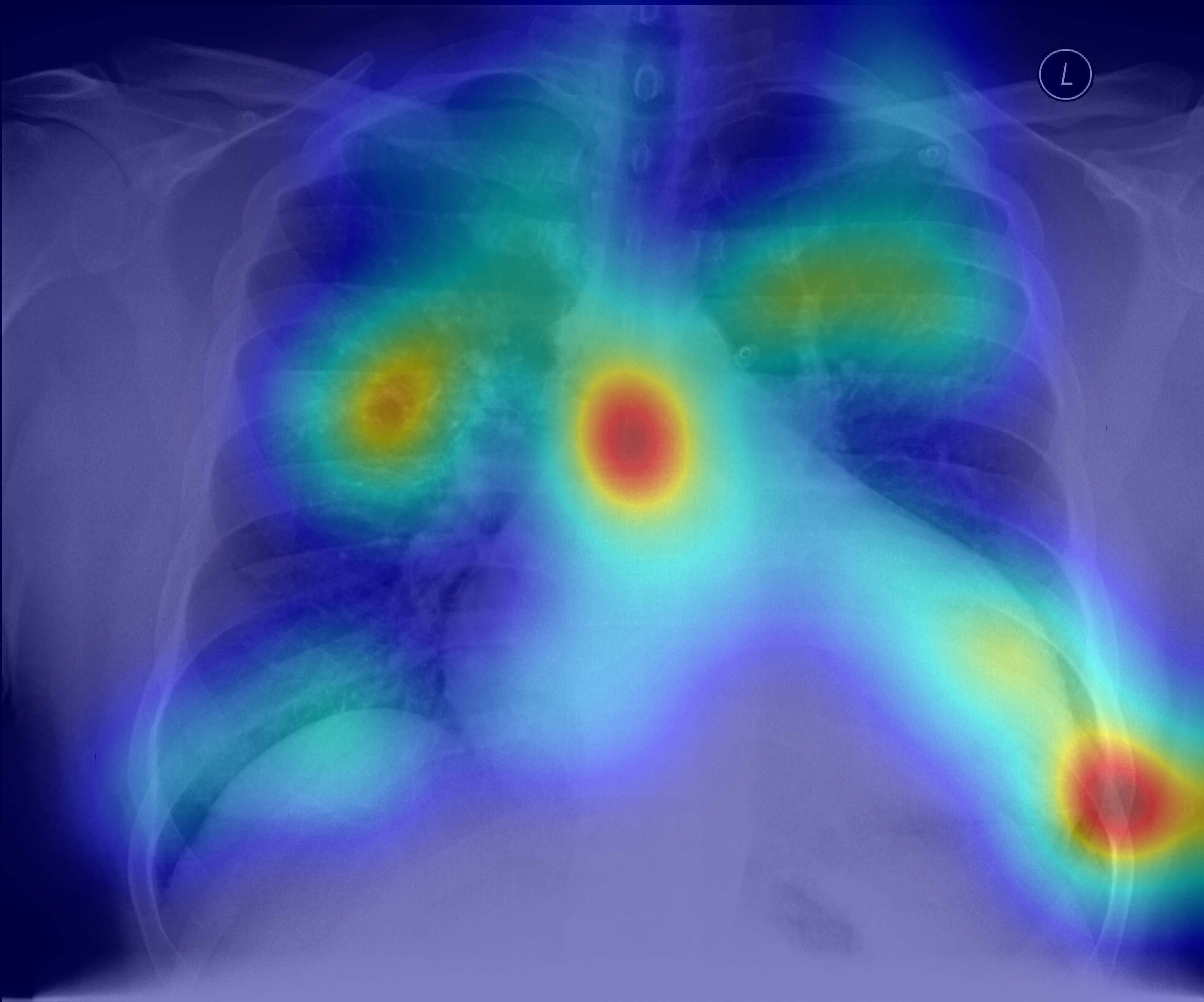}
& \includegraphics[width=\linewidth]{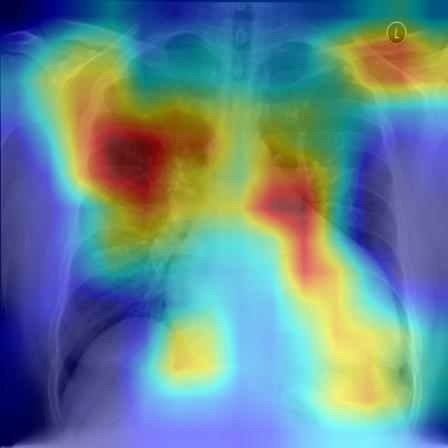}
& \includegraphics[width=\linewidth]{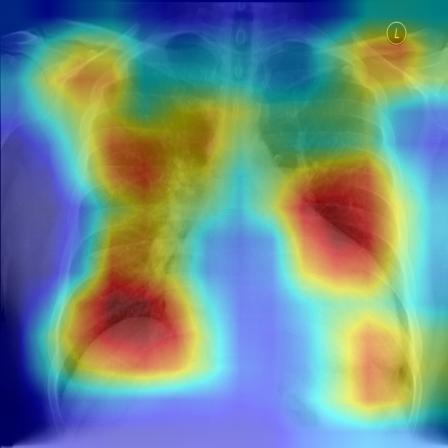}
& \includegraphics[width=\linewidth]{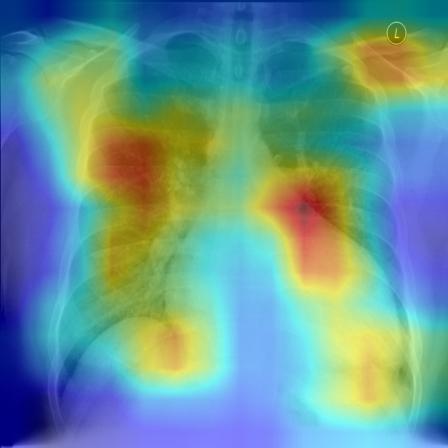}
& \includegraphics[width=\linewidth]{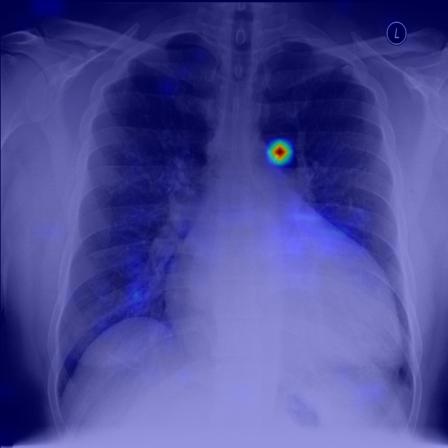}
& \includegraphics[width=\linewidth]{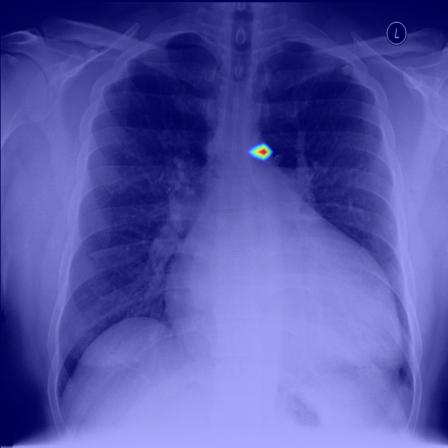} \\

CXR - Normal
& \includegraphics[width=\linewidth]{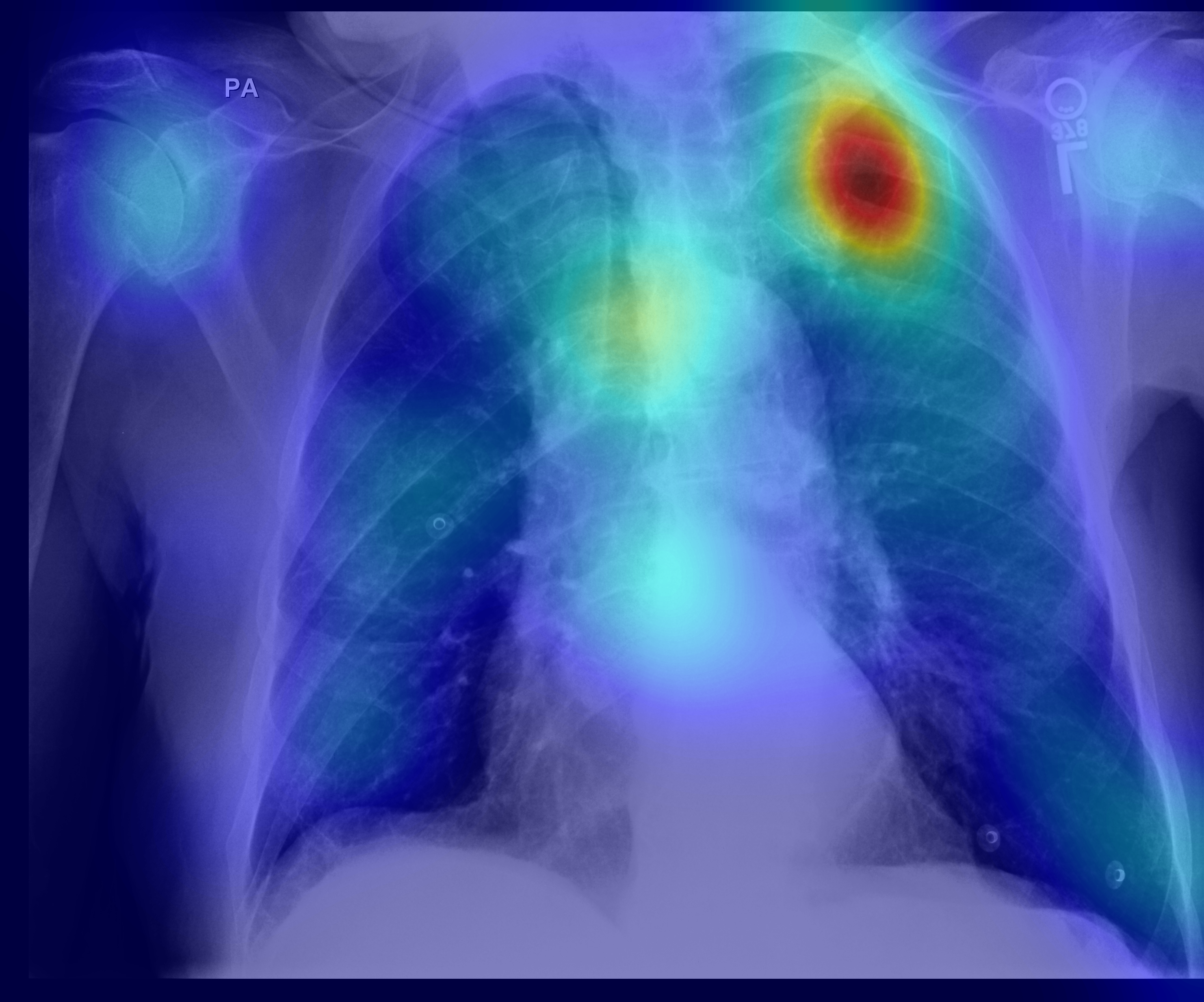}
& \includegraphics[width=\linewidth]{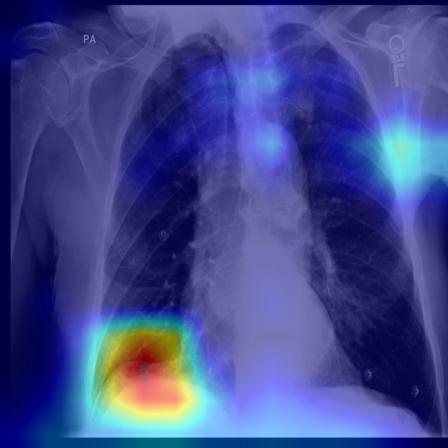}
& \includegraphics[width=\linewidth]{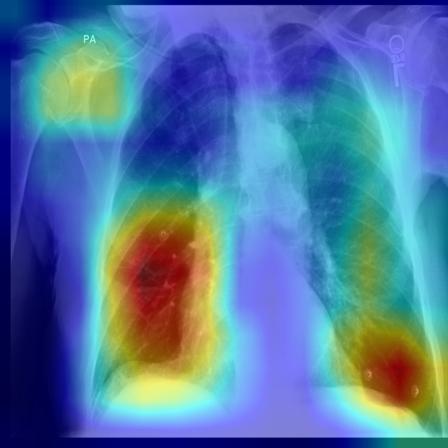}
& \includegraphics[width=\linewidth]{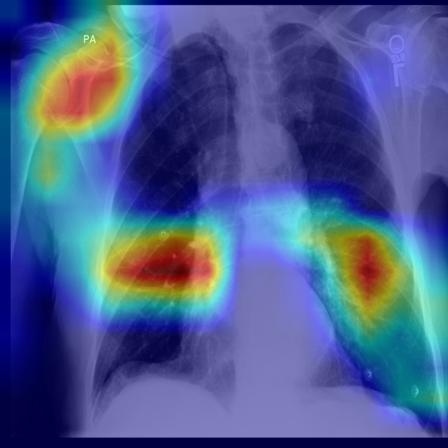}
& \includegraphics[width=\linewidth]{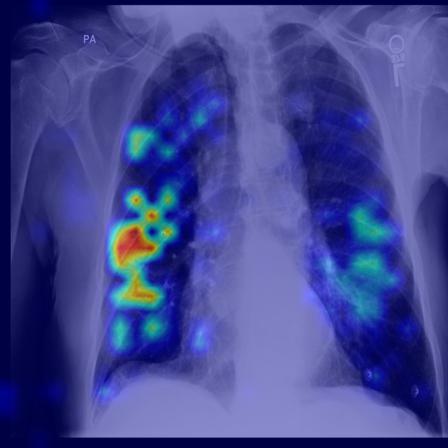}
& \includegraphics[width=\linewidth]{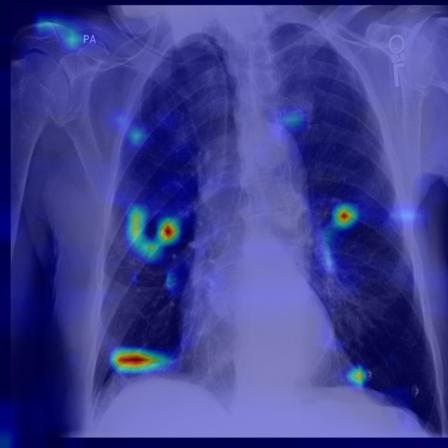} \\

KDEF-FGD - Happy
& \includegraphics[width=\linewidth]{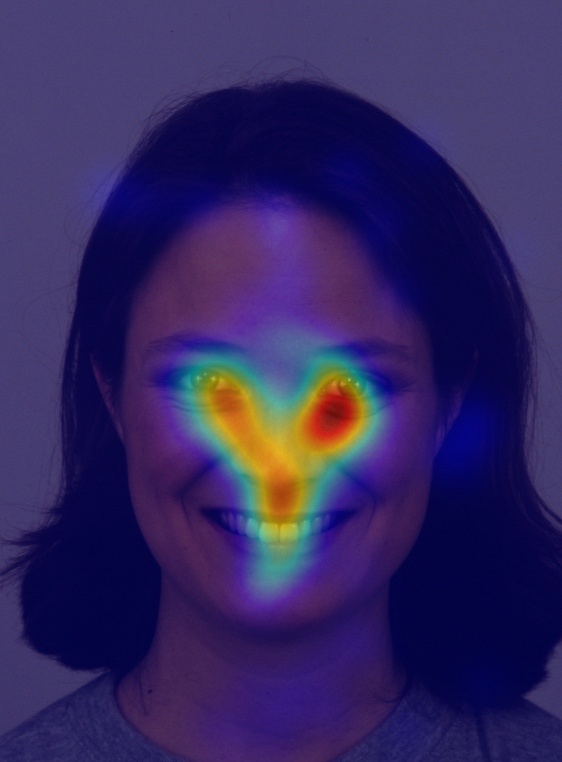}
& \includegraphics[width=\linewidth]{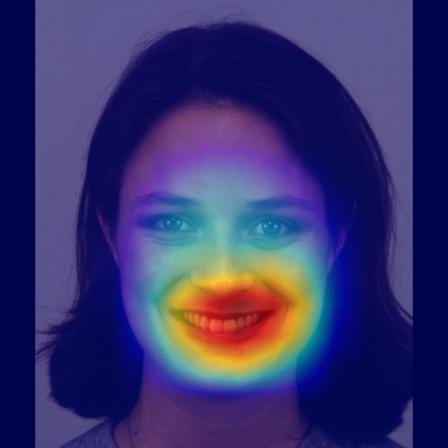}
& \includegraphics[width=\linewidth]{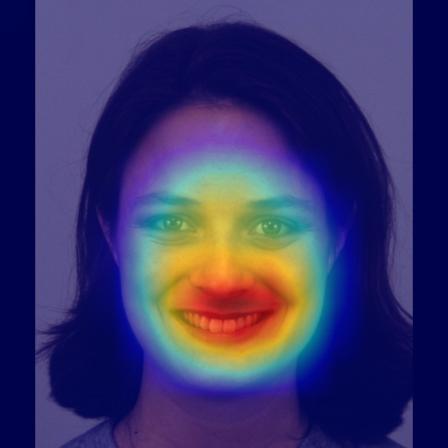}
& \includegraphics[width=\linewidth]{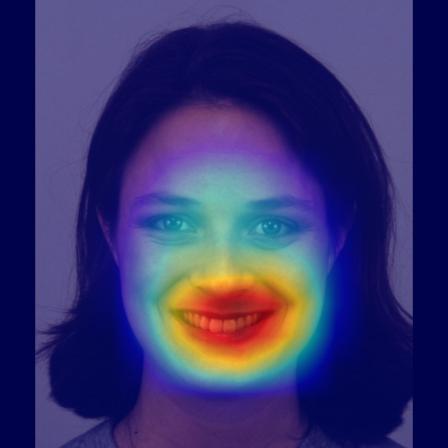}
& \includegraphics[width=\linewidth]{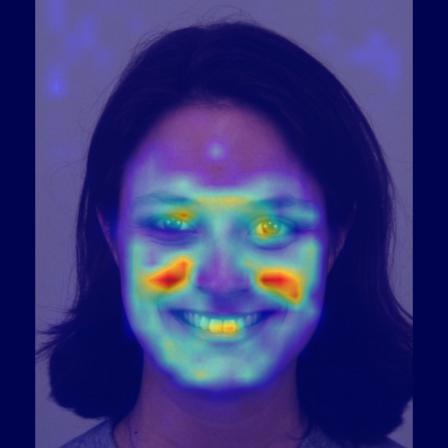}
& \includegraphics[width=\linewidth]{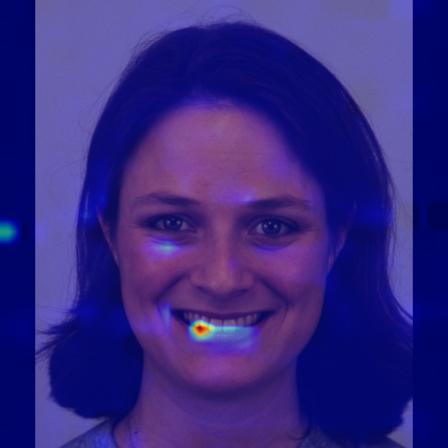} \\

KDEF-FGD - Sad
& \includegraphics[width=\linewidth]{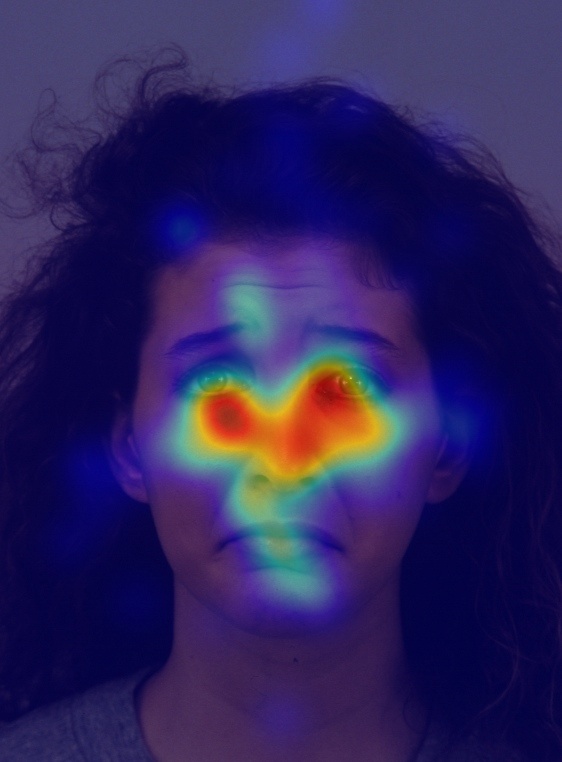}
& \includegraphics[width=\linewidth]{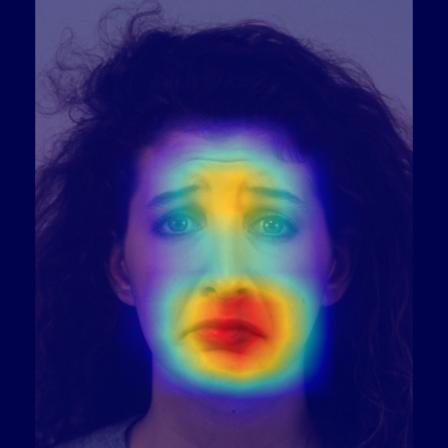}
& \includegraphics[width=\linewidth]{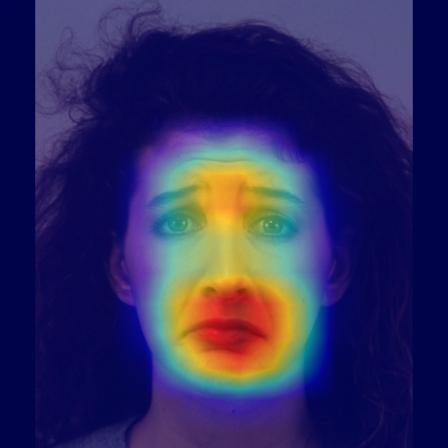}
& \includegraphics[width=\linewidth]{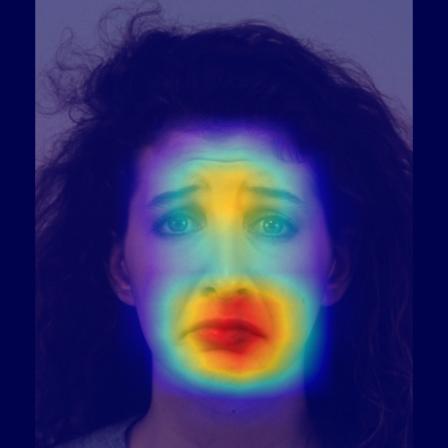}
& \includegraphics[width=\linewidth]{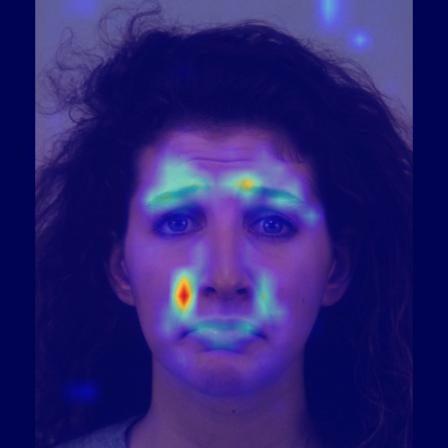}
& \includegraphics[width=\linewidth]{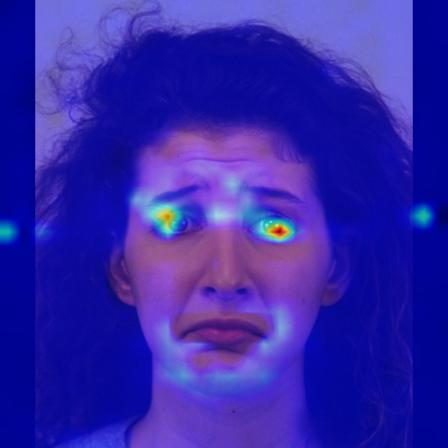} \\

KDEF-FGD - Neutral
& \includegraphics[width=\linewidth]{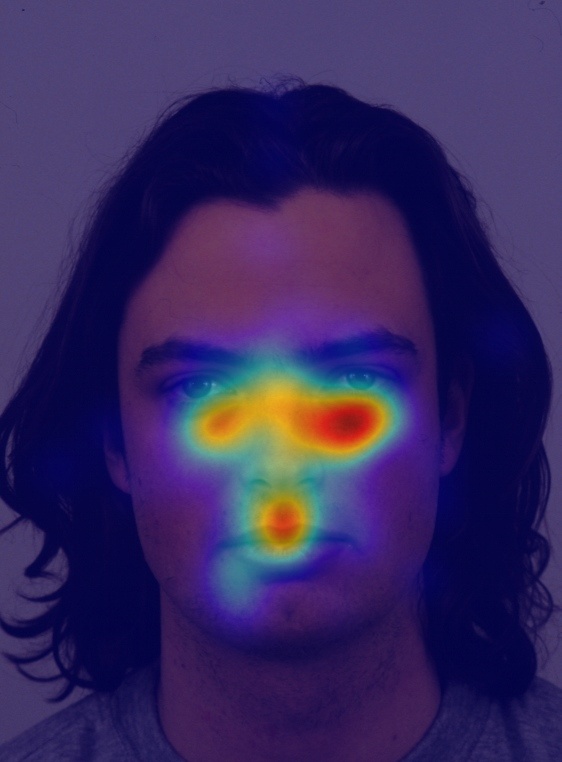}
& \includegraphics[width=\linewidth]{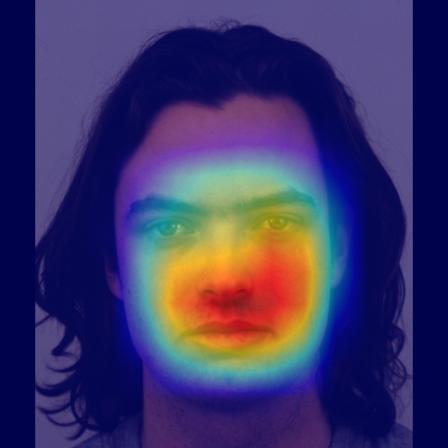}
& \includegraphics[width=\linewidth]{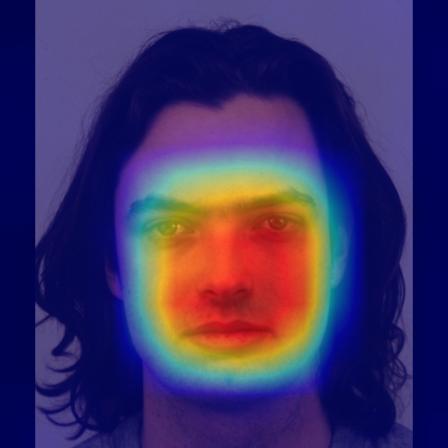}
& \includegraphics[width=\linewidth]{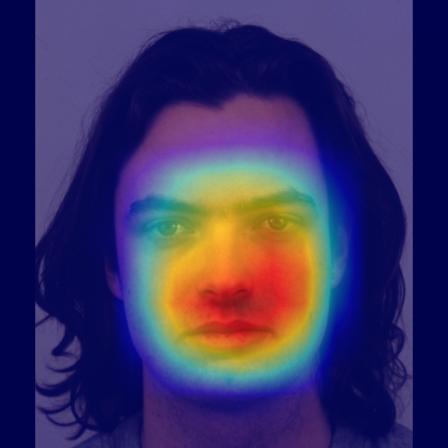}
& \includegraphics[width=\linewidth]{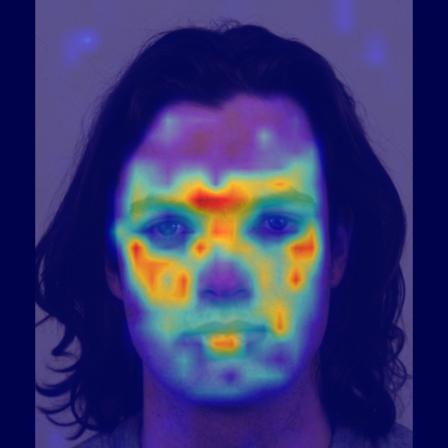}
& \includegraphics[width=\linewidth]{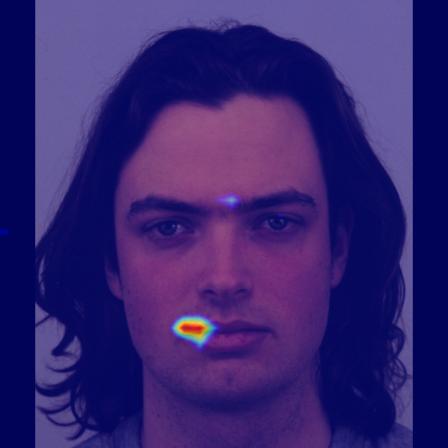} \\

\bottomrule
\end{tabular*}
\end{table*}

\FloatBarrier
\clearpage

\printcredits

\bibliographystyle{elsarticle-num}

\bibliography{references}

\newpage

\section*{Author Biographies}
\bio{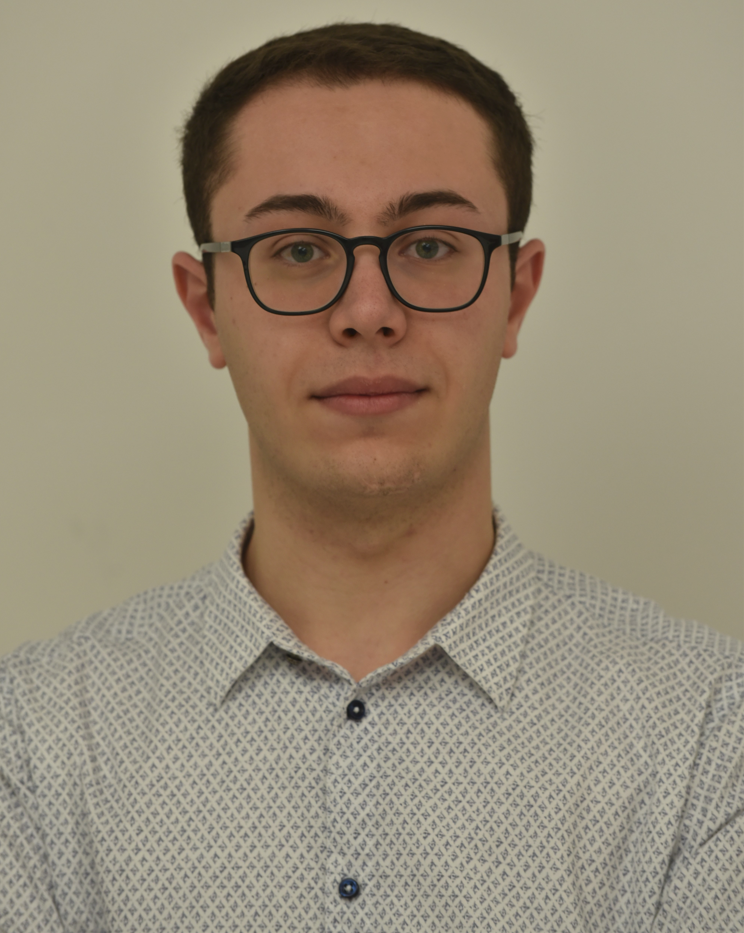}
\textbf{Marco Stefani} received the B.S. degree in Computer Engineering from the University of Padua in 2024. He is currently pursuing the M.S. degree in Artificial Intelligence and Robotics at the same university. His research interests include optimization problems in industrial and logistics domains, such as scheduling, Radio-Frequency Identification (RFID) systems management, and the analysis of explainable Artificial Intelligence (XAI) strategies.
\endbio

\bio{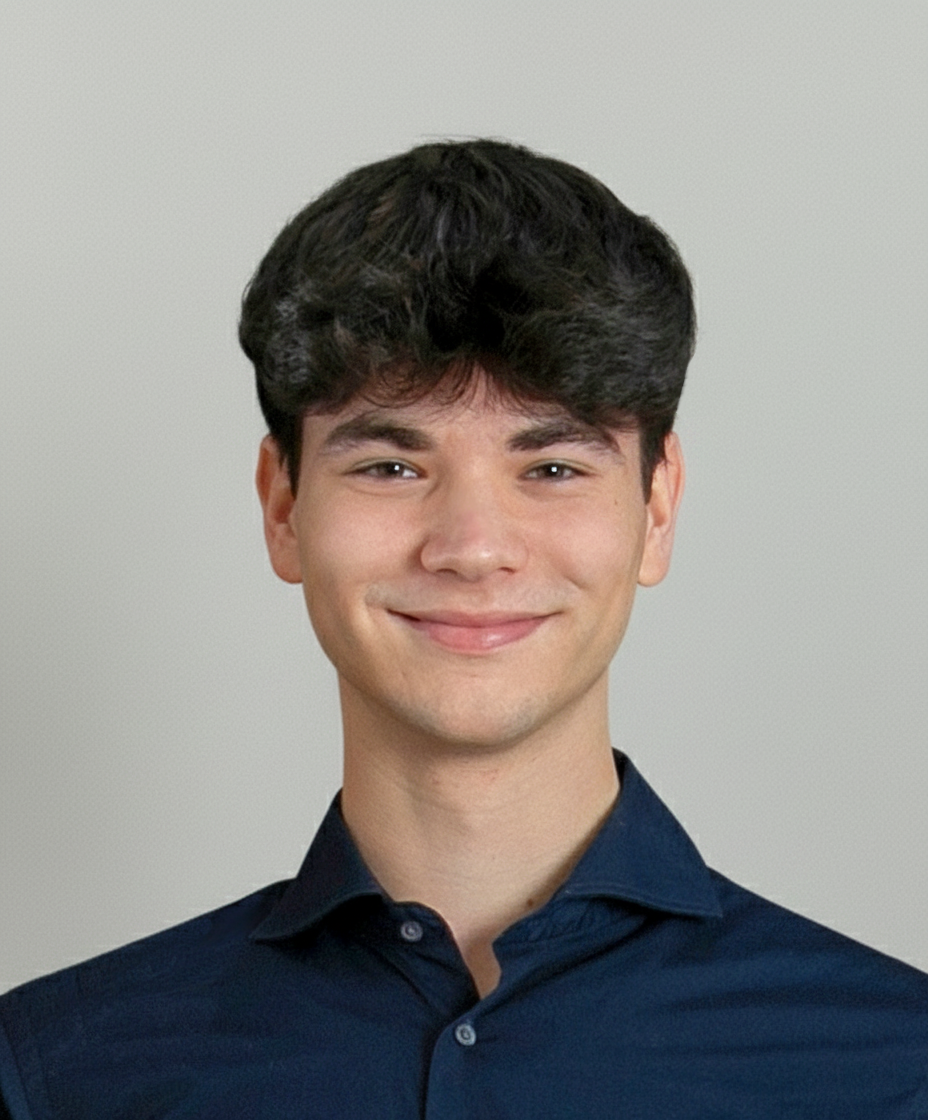}
\textbf{Edoardo Parpaiola} received the B.S. degree in Computer Engineering from the University of Padua, Italy, in 2024. He is currently pursuing the M.S. degree in Artificial Intelligence and Robotics at the same university. His research interests include industrial robotics and computer vision, with a particular focus on image classification, object detection, 3D data reconstruction, and explainable artificial intelligence (XAI).
\endbio

\bio{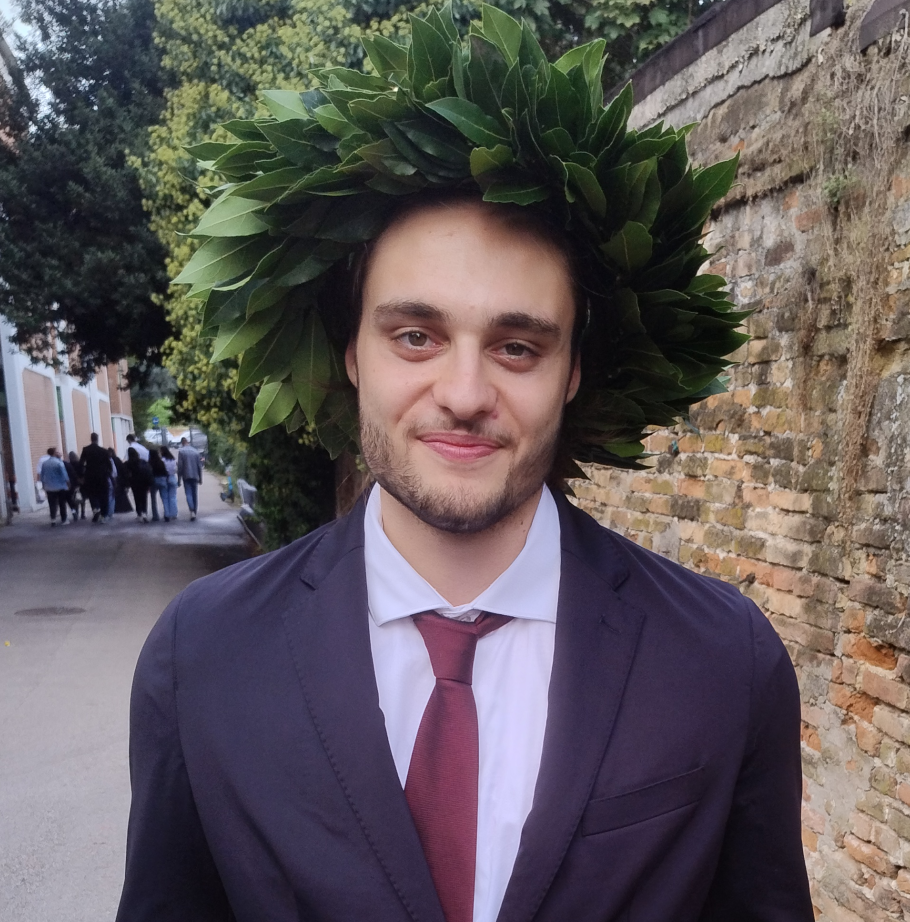}
\textbf{Matteo Bino} received the B.S. degree in Computer Engineering from the University of Padua, Italy, in 2024. He is currently pursuing the M.S. degree in Artificial Intelligence and Robotics at the same university. His research interests include computer vision, explainable AI, and mobile robotics, with a particular focus on the use of reinforcement learning for complex task execution and autonomous navigation.
\endbio

\bio{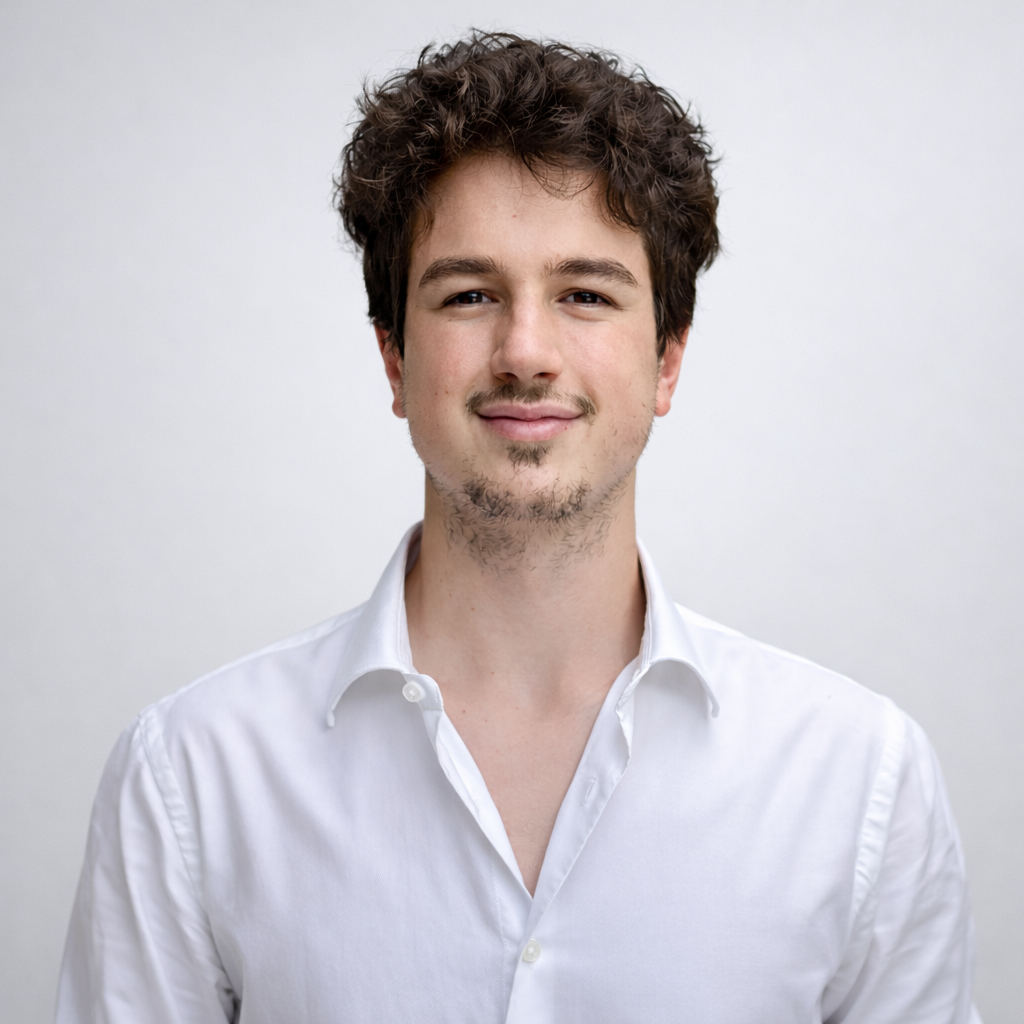}
\textbf{Gianluca Caregnato} received the B.S. degree in Computer Engineering from the University of Padua, Italy, in 2024. He is currently pursuing the M.S. degree in Artificial Intelligence and Robotics at the same university. His research lies at the intersection of computer vision and robotics, covering both autonomous and industrial platforms, with a particular emphasis on Explainable AI (XAI).
\endbio

\bio{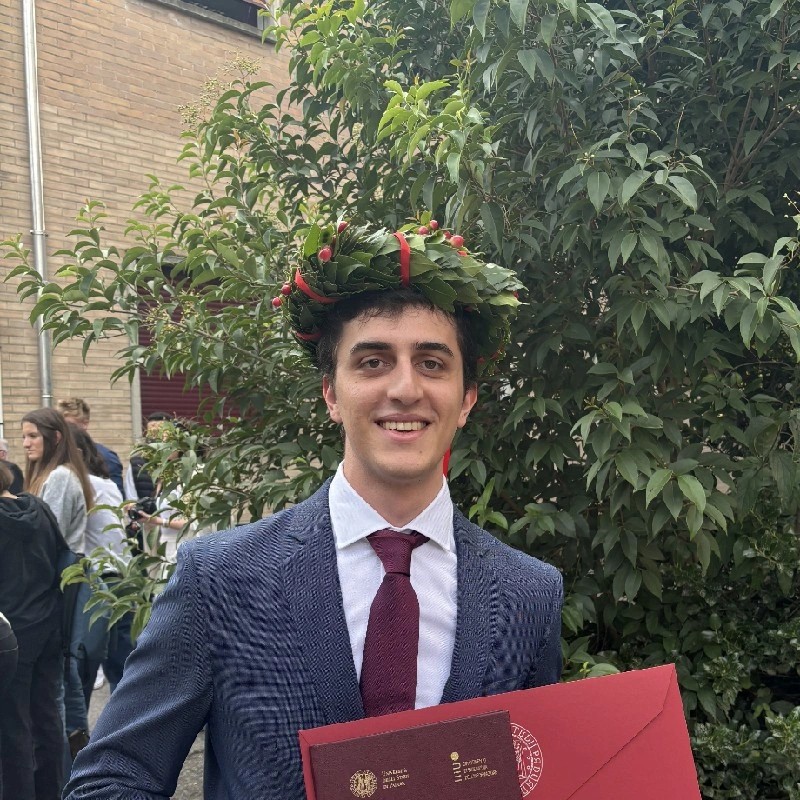}
\textbf{Francesco Roncolato} received the B.S. degree in Computer Engineering from the University of Padua, Italy, in 2024. He is currently pursuing the M.S. degree in Artificial Intelligence and Robotics at the same university. His research interests include Explainable AI (XAI), autonomous robotics, and computer vision, with a particular focus on applying optimization techniques to real-world control problems.
\endbio

\bio{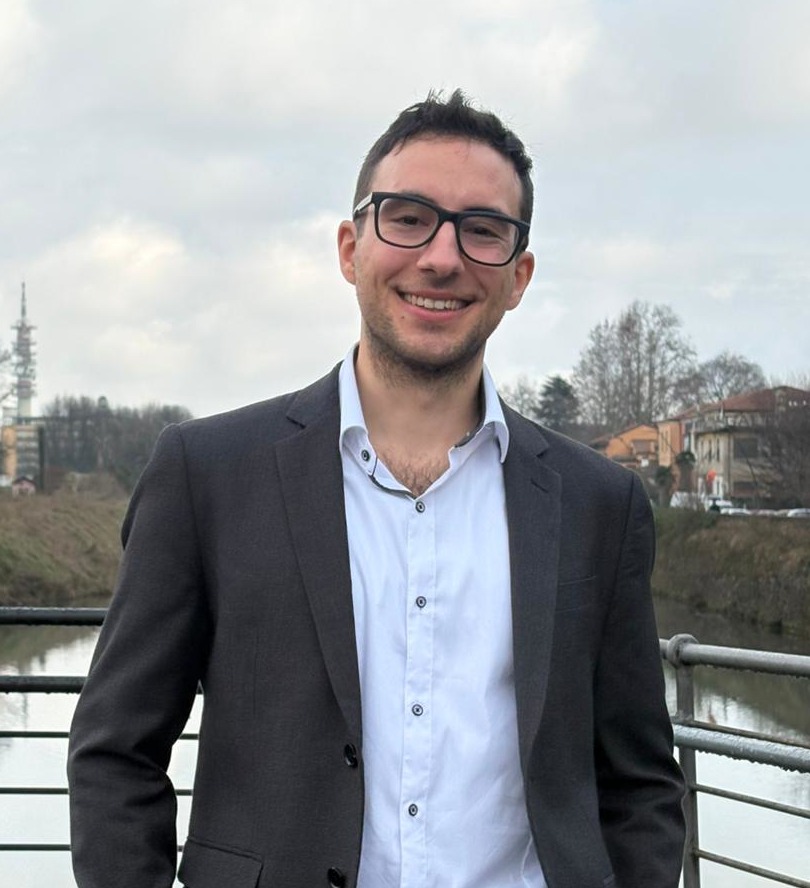}
\textbf{Daniel Fusaro} received the B.S., M.S., and Ph.D. degrees in Computer Engineering from the University of Padua, Italy, in 2019, 2021, and 2025, respectively. He is currently a Research Fellow at the University of Padua. His research interests include computer vision, robotics, and artificial intelligence, with a particular focus on 3D scene understanding, LiDAR point cloud processing, autonomous navigation, and self-supervised learning for multimodal perception.
\endbio

\bio{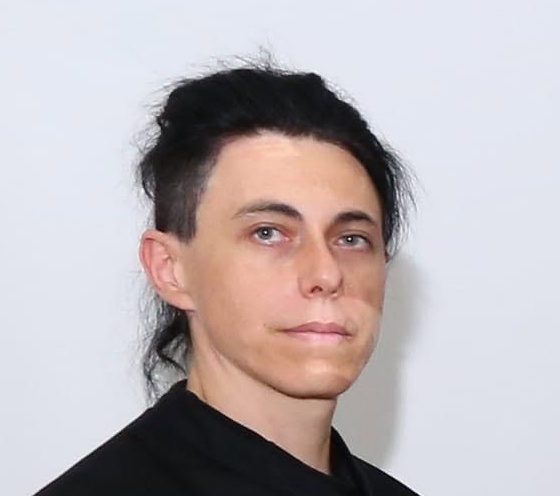}
\textbf{Loris Nanni} is a Professor with the Department of Information Engineering at the University of Padua, Italy. His research interests include artificial intelligence, machine learning, deep learning, computer vision, pattern recognition, bioinformatics, and medical image analysis. He has authored more than 350 peer-reviewed publications in leading international journals and conferences, with contributions spanning both methodological advances and real-world biomedical applications. According to Google Scholar, his research has received over 15,000 citations, with an h-index of 62, reflecting the broad impact of his work within the AI and pattern recognition communities.
\endbio

\bio{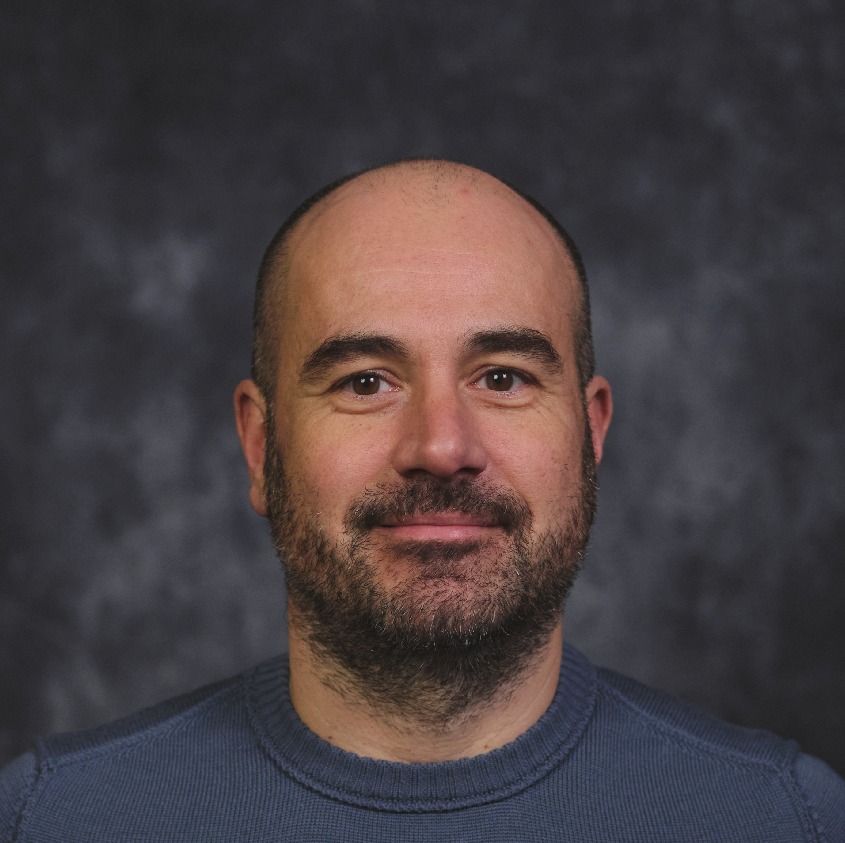}
\textbf{Alberto Pretto} 
Alberto Pretto is an Associate Professor with the Department of Information Engineering at the University of Padua, Italy, where he leads the Field Robotics Group within the Intelligent Autonomous Systems Laboratory (IAS Lab). His research interests include autonomous robotics, computer vision, robot perception, and artificial intelligence. His career combines academic research with industrial innovation, including the co-founding of technology companies in robotics, computer vision, and AI. He has also held visiting research positions at leading international research laboratories in Germany and the United States.
\endbio

\end{document}